\pdfoutput=1
\documentclass[letterpaper]{article} 
\usepackage{aaai2027}    
\nocopyright            
\usepackage{times}    
\usepackage{helvet}   
\usepackage{courier}  
\usepackage[hyphens]{url}  
\usepackage{graphicx} 
\graphicspath{{./}}
\usepackage{natbib}   
\usepackage{caption}  
\usepackage{amsmath,amssymb}
\usepackage{array}
\usepackage{booktabs}
\usepackage{placeins}
\newif\ifarxivbuild
\arxivbuildtrue
\newif\ifbodyfloats
\bodyfloatstrue
\newif\ifkeyfloats
\keyfloatsfalse
\newif\ifconciseconcl
\conciseconclfalse
\newcommand{\fltab}[1]{\ifbodyfloats Table~\ref{#1}\else Appendix Table~\ref{#1}\fi}
\newcommand{\flfig}[1]{\ifbodyfloats Figure~\ref{#1}\else Appendix Figure~\ref{#1}\fi}
\newcommand{\fltabkey}[1]{\ifbodyfloats Table~\ref{#1}\else\ifkeyfloats Table~\ref{#1}\else Appendix Table~\ref{#1}\fi\fi}

\makeatletter
\newcommand{\pvset}[2]{\expandafter\gdef\csname pv@#1\endcsname{#2}}
\newcommand{\pv}[1]{\ifcsname pv@#1\endcsname\csname pv@#1\endcsname\else??\fi}
\makeatother
\IfFileExists{paper_values.tex}{

\pvset{e0.winner}{the mixer output}
\pvset{e0.winner.auroc}{1.00}
\pvset{e0.winner.auroc.best}{1.00}
\pvset{e0.Delta.auroc}{1.00}
\pvset{e0.Delta.best}{1.00}
\pvset{e0.B.auroc}{0.92}
\pvset{e0.B.best}{0.98}
\pvset{e0.C.auroc}{0.93}
\pvset{e0.C.best}{0.98}
\pvset{e0.output.auroc}{1.00}
\pvset{e0.output.best}{1.00}
\pvset{e0.nlayers}{64}
\pvset{e0.Delta.lowoverlap}{0.99}
\pvset{e0.B.lowoverlap}{0.86}
\pvset{e0.C.lowoverlap}{0.88}
\pvset{e0.output.lowoverlap}{1.00}
\pvset{e0.winner.lowoverlap}{1.00}
\pvset{e0.surface.auroc}{0.99}
\pvset{e0.ncorpus}{250}
\pvset{readout.n}{120}
\pvset{readout.genn}{24}
\pvset{readout.decodability.base}{1.00}
\pvset{readout.decodability.delta}{1.00}
\pvset{readout.refusal.base}{91.7\%}
\pvset{readout.refusal.delta}{95.8\%}
\pvset{readout.coh.base}{98.5\%}
\pvset{readout.coh.delta}{92.0\%}
\pvset{readout.refusal.b}{83.3\%}
\pvset{readout.coh.b}{99.2\%}
\pvset{e0m.surface.auroc}{0.46}
\pvset{e0m.ncorpus}{120}
\pvset{e0m.Delta.auroc}{0.90}
\pvset{e0m.B.auroc}{0.69}
\pvset{e0m.C.auroc}{0.69}
\pvset{e0m.output.auroc}{0.95}
\pvset{e0.llama.auroc}{1.00}
\pvset{e0.llama.lowoverlap}{1.00}
\pvset{e0.llama.matched}{0.99}
\pvset{e0.llama.matched.surface}{0.44}
\pvset{e1cap.base.mmlu}{0.478}
\pvset{e1cap.base.truthfulqa}{0.297}
\pvset{e1cap.gatekeeper.mmlu}{0.478}
\pvset{e1cap.gatekeeper.truthfulqa}{0.297}
\pvset{e1cap.mmlu.fpr}{0.0\%}
\pvset{scale.small}{130M}
\pvset{scale.large}{7B}
\pvset{scale.delta.small}{0.89}
\pvset{scale.delta.large}{0.99}
\pvset{scale.surface}{0.44}
\pvset{scale.nmodels}{5}
\pvset{mamba2.auroc}{1.00}
\pvset{mamba2.meanauroc}{0.98}
\pvset{mamba2.nceil}{61}
\pvset{mamba2.nlayers}{64}
\pvset{locus.delta.low}{0.95}
\pvset{locus.b.low}{0.79}
\pvset{locus.c.low}{0.78}
\pvset{locus.output.low}{0.99}
\pvset{locus.surface}{0.54}
\pvset{locus.nper}{200}
\pvset{harden.plain.plain_harm}{1.00}
\pvset{harden.hard.plain_harm}{1.00}
\pvset{harden.plain.plain_spicy}{0.00}
\pvset{harden.hard.plain_spicy}{0.00}
\pvset{harden.plain.wrapped_harm}{1.00}
\pvset{harden.hard.wrapped_harm}{1.00}
\pvset{harden.plain.wrapped_spicy}{0.78}
\pvset{harden.hard.wrapped_spicy}{0.00}
\pvset{base.base.prefix_injection.asr}{98.0\%}
\pvset{base.base.prefix_injection.overrefusal}{12.0\%}
\pvset{base.ours_delta.prefix_injection.asr}{5.0\%}
\pvset{base.ours_delta.prefix_injection.overrefusal}{12.0\%}
\pvset{base.openloop_delta.prefix_injection.asr}{5.0\%}
\pvset{base.openloop_delta.prefix_injection.overrefusal}{86.0\%}
\pvset{base.cast_residual.prefix_injection.asr}{100.0\%}
\pvset{base.cast_residual.prefix_injection.overrefusal}{12.0\%}
\pvset{base.projection.prefix_injection.asr}{95.0\%}
\pvset{base.projection.prefix_injection.overrefusal}{9.0\%}
\pvset{base.ppl_filter.prefix_injection.asr}{98.0\%}
\pvset{base.ppl_filter.prefix_injection.overrefusal}{19.0\%}
\pvset{base.smoothllm.prefix_injection.asr}{62.0\%}
\pvset{base.smoothllm.prefix_injection.overrefusal}{29.0\%}
\pvset{base.ours.prefix_injection.overrefdelta}{$+0.0$ pt}
\pvset{base.base.roleplay_dan.asr}{81.5\%}
\pvset{base.base.roleplay_dan.overrefusal}{12.0\%}
\pvset{base.ours_delta.roleplay_dan.asr}{30.0\%}
\pvset{base.ours_delta.roleplay_dan.overrefusal}{12.0\%}
\pvset{base.openloop_delta.roleplay_dan.asr}{30.0\%}
\pvset{base.openloop_delta.roleplay_dan.overrefusal}{86.0\%}
\pvset{base.cast_residual.roleplay_dan.asr}{100.0\%}
\pvset{base.cast_residual.roleplay_dan.overrefusal}{12.0\%}
\pvset{base.projection.roleplay_dan.asr}{87.5\%}
\pvset{base.projection.roleplay_dan.overrefusal}{9.0\%}
\pvset{base.ppl_filter.roleplay_dan.asr}{81.5\%}
\pvset{base.ppl_filter.roleplay_dan.overrefusal}{19.0\%}
\pvset{base.smoothllm.roleplay_dan.asr}{72.0\%}
\pvset{base.smoothllm.roleplay_dan.overrefusal}{29.0\%}
\pvset{base.ours.roleplay_dan.overrefdelta}{$+0.0$ pt}
\pvset{base.base.fictional_framing.asr}{87.0\%}
\pvset{base.base.fictional_framing.overrefusal}{12.0\%}
\pvset{base.ours_delta.fictional_framing.asr}{16.5\%}
\pvset{base.ours_delta.fictional_framing.overrefusal}{12.0\%}
\pvset{base.openloop_delta.fictional_framing.asr}{16.5\%}
\pvset{base.openloop_delta.fictional_framing.overrefusal}{86.0\%}
\pvset{base.cast_residual.fictional_framing.asr}{100.0\%}
\pvset{base.cast_residual.fictional_framing.overrefusal}{12.0\%}
\pvset{base.projection.fictional_framing.asr}{94.5\%}
\pvset{base.projection.fictional_framing.overrefusal}{9.0\%}
\pvset{base.ppl_filter.fictional_framing.asr}{87.0\%}
\pvset{base.ppl_filter.fictional_framing.overrefusal}{19.0\%}
\pvset{base.smoothllm.fictional_framing.asr}{86.5\%}
\pvset{base.smoothllm.fictional_framing.overrefusal}{29.0\%}
\pvset{base.ours.fictional_framing.overrefdelta}{$+0.0$ pt}
\pvset{base.cast_residual.prefix_injection.cohfrac}{10.5\%}
\pvset{base.cast_residual.roleplay_dan.cohfrac}{7.0\%}
\pvset{base.cast_residual.fictional_framing.cohfrac}{11.0\%}
\pvset{base.cast_residual.cohfrac.pooled}{9.5\%}
\pvset{base.cast_residual.cohfrac.pooled.n}{600}
\pvset{def.ssm.prefix_injection.base}{98.0\%}
\pvset{def.ssm.prefix_injection.basej4}{78.0\%}
\pvset{def.ssm.prefix_injection.gated}{5.0\%}
\pvset{def.ssm.prefix_injection.gatedj2}{5.0\%}
\pvset{def.ssm.prefix_injection.gatedj4}{2.0\%}
\pvset{def.ssm.roleplay_dan.base}{81.5\%}
\pvset{def.ssm.roleplay_dan.basej4}{64.5\%}
\pvset{def.ssm.roleplay_dan.gated}{30.0\%}
\pvset{def.ssm.roleplay_dan.gatedj2}{37.0\%}
\pvset{def.ssm.roleplay_dan.gatedj4}{23.0\%}
\pvset{def.ssm.fictional_framing.base}{87.0\%}
\pvset{def.ssm.fictional_framing.basej4}{37.5\%}
\pvset{def.ssm.fictional_framing.gated}{16.5\%}
\pvset{def.ssm.fictional_framing.gatedj2}{18.0\%}
\pvset{def.ssm.fictional_framing.gatedj4}{10.0\%}
\pvset{def.ssm.base}{98.0\%}
\pvset{def.ssm.gated}{5.0\%}
\pvset{def.ssm.n}{200}
\pvset{def.ssm.nspicy}{100}
\pvset{def.ssm.judgeagree}{93.5\%}
\pvset{def.ssm.judge2name}{Llama-3.1-8B-Instruct (prompted, not Llama-Guard)}
\pvset{locus.ssm2.delta}{1.00}
\pvset{locus.ssm2.delta.low}{0.99}
\pvset{locus.ssm2.surface}{0.99}
\pvset{def.ssm2.fictional_framing.base}{56.0\%}
\pvset{def.ssm2.fictional_framing.gated}{4.5\%}
\pvset{def.ssm2.roleplay_dan.base}{13.0\%}
\pvset{def.ssm2.roleplay_dan.gated}{0.0\%}
\pvset{def.openloop.prefix_injection.asr}{5.0\%}
\pvset{def.openloop.roleplay_dan.asr}{30.0\%}
\pvset{def.openloop.fictional_framing.asr}{16.5\%}
\pvset{def.openloop.overrefusal}{86.0\%}
\pvset{transfer.heldout}{0.95}
\pvset{transfer.shuffled}{0.17}
\pvset{transfer.rawcos}{0.01}
\pvset{transfer.x2s.defend.base}{97.0\%}
\pvset{transfer.x2s.defend.native}{21.0\%}
\pvset{transfer.x2s.defend.transferred}{28.5\%}
\pvset{transfer.x2s.defend.random_mapped}{97.5\%}
\pvset{transfer.x2s.defend.cos}{0.992}
\pvset{transfer.x2s.defend.cosrand}{0.312}
\pvset{transfer.x2s.ablate.base}{10.5\%}
\pvset{transfer.x2s.ablate.native}{41.5\%}
\pvset{transfer.x2s.ablate.transferred}{36.5\%}
\pvset{transfer.x2s.ablate.random_mapped}{8.5\%}
\pvset{transfer.x2s.ablate.random_orth}{9.0\%}
\pvset{transfer.x2s.ablate.transferred_orth}{14.0\%}
\pvset{transfer.x2s.ablate.transferred_orth_std}{44.0\%}
\pvset{transfer.x2s.ablate.cos}{0.992}
\pvset{transfer.x2s.ablate.cosrand}{0.312}
\pvset{transfer.x2s.ablate.cosorth}{0.000}
\pvset{transfer.s2x.defend.base}{31.0\%}
\pvset{transfer.s2x.defend.native}{88.5\%}
\pvset{transfer.s2x.defend.transferred}{42.0\%}
\pvset{transfer.s2x.defend.random_mapped}{81.0\%}
\pvset{transfer.s2x.defend.cos}{0.755}
\pvset{transfer.s2x.defend.cosrand}{0.334}
\pvset{transfer.s2x.ablate.base}{2.5\%}
\pvset{transfer.s2x.ablate.native}{50.5\%}
\pvset{transfer.s2x.ablate.transferred}{79.0\%}
\pvset{transfer.s2x.ablate.random_mapped}{4.0\%}
\pvset{transfer.s2x.ablate.random_orth}{8.0\%}
\pvset{transfer.s2x.ablate.transferred_orth}{94.5\%}
\pvset{transfer.s2x.ablate.transferred_orth_std}{61.0\%}
\pvset{transfer.s2x.ablate.cos}{0.755}
\pvset{transfer.s2x.ablate.cosrand}{0.334}
\pvset{transfer.s2x.ablate.cosorth}{0.000}
\pvset{transfer.mi2ma.defend.base}{97.0\%}
\pvset{transfer.mi2ma.defend.native}{21.0\%}
\pvset{transfer.mi2ma.defend.transferred}{29.5\%}
\pvset{transfer.mi2ma.defend.random_mapped}{73.0\%}
\pvset{transfer.mi2ma.defend.cos}{0.994}
\pvset{transfer.mi2ma.defend.cosrand}{0.058}
\pvset{transfer.mi2ma.ablate.base}{10.5\%}
\pvset{transfer.mi2ma.ablate.native}{41.5\%}
\pvset{transfer.mi2ma.ablate.transferred}{36.5\%}
\pvset{transfer.mi2ma.ablate.random_mapped}{9.5\%}
\pvset{transfer.mi2ma.ablate.random_orth}{9.5\%}
\pvset{transfer.mi2ma.ablate.transferred_orth}{10.0\%}
\pvset{transfer.mi2ma.ablate.transferred_orth_std}{45.0\%}
\pvset{transfer.mi2ma.ablate.cos}{0.994}
\pvset{transfer.mi2ma.ablate.cosrand}{0.058}
\pvset{transfer.mi2ma.ablate.cosorth}{0.000}
\pvset{transfer.ma2mi.defend.base}{96.0\%}
\pvset{transfer.ma2mi.defend.native}{98.5\%}
\pvset{transfer.ma2mi.defend.transferred}{97.0\%}
\pvset{transfer.ma2mi.defend.random_mapped}{82.0\%}
\pvset{transfer.ma2mi.defend.cos}{0.986}
\pvset{transfer.ma2mi.defend.cosrand}{0.370}
\pvset{transfer.ma2mi.ablate.base}{26.5\%}
\pvset{transfer.ma2mi.ablate.native}{50.0\%}
\pvset{transfer.ma2mi.ablate.transferred}{60.0\%}
\pvset{transfer.ma2mi.ablate.random_mapped}{66.0\%}
\pvset{transfer.ma2mi.ablate.random_orth}{27.0\%}
\pvset{transfer.ma2mi.ablate.transferred_orth}{92.5\%}
\pvset{transfer.ma2mi.ablate.transferred_orth_std}{51.0\%}
\pvset{transfer.ma2mi.ablate.cos}{0.986}
\pvset{transfer.ma2mi.ablate.cosrand}{0.370}
\pvset{transfer.ma2mi.ablate.cosorth}{0.000}
\pvset{transfer.l2f3.ablate.base}{0.0\%}
\pvset{transfer.l2f3.ablate.native}{42.0\%}
\pvset{transfer.l2f3.ablate.transferred}{25.5\%}
\pvset{transfer.l2f3.ablate.random_mapped}{0.0\%}
\pvset{transfer.l2f3.ablate.random_orth}{0.0\%}
\pvset{transfer.l2f3.ablate.cos}{0.995}
\pvset{transfer.l2f3.ablate.cosrand}{0.354}
\pvset{transfer.l2f3.ablate.cosorth}{0.000}
\pvset{transfer.f32l.ablate.base}{2.5\%}
\pvset{transfer.f32l.ablate.native}{51.5\%}
\pvset{transfer.f32l.ablate.transferred}{55.5\%}
\pvset{transfer.f32l.ablate.random_mapped}{3.5\%}
\pvset{transfer.f32l.ablate.random_orth}{2.5\%}
\pvset{transfer.f32l.ablate.cos}{0.997}
\pvset{transfer.f32l.ablate.cosrand}{0.515}
\pvset{transfer.f32l.ablate.cosorth}{0.000}
\pvset{transfer.x2s.ablate.orthabl}{14.0\%}
\pvset{transfer.x2s.ablate.orthcos}{0.68}
\pvset{transfer.x2s.ablate.orthcoh}{100.0\%}
\pvset{transfer.x2s.ablate.orthstdabl}{44.0\%}
\pvset{transfer.x2s.ablate.orthstdcos}{0.998}
\pvset{transfer.x2s.ablate.orthstdcoh}{100.0\%}
\pvset{transfer.s2x.ablate.orthabl}{94.5\%}
\pvset{transfer.s2x.ablate.orthcos}{0.70}
\pvset{transfer.s2x.ablate.orthcoh}{1.5\%}
\pvset{transfer.s2x.ablate.orthstdabl}{61.0\%}
\pvset{transfer.s2x.ablate.orthstdcos}{0.997}
\pvset{transfer.s2x.ablate.orthstdcoh}{100.0\%}
\pvset{transfer.mi2ma.ablate.orthabl}{10.0\%}
\pvset{transfer.mi2ma.ablate.orthcos}{0.35}
\pvset{transfer.mi2ma.ablate.orthcoh}{100.0\%}
\pvset{transfer.mi2ma.ablate.orthstdabl}{45.0\%}
\pvset{transfer.mi2ma.ablate.orthstdcos}{0.998}
\pvset{transfer.mi2ma.ablate.orthstdcoh}{100.0\%}
\pvset{transfer.ma2mi.ablate.orthabl}{92.5\%}
\pvset{transfer.ma2mi.ablate.orthcos}{0.36}
\pvset{transfer.ma2mi.ablate.orthcoh}{3.0\%}
\pvset{transfer.ma2mi.ablate.orthstdabl}{51.0\%}
\pvset{transfer.ma2mi.ablate.orthstdcos}{0.999}
\pvset{transfer.ma2mi.ablate.orthstdcoh}{100.0\%}
\pvset{shared.proc}{1.00}
\pvset{shared.ridge}{1.00}
\pvset{shared.shuf}{0.57}
\pvset{shared.tgtown}{1.00}
\pvset{shared.procm}{1.00}
\pvset{shared.ridgem}{1.00}
\pvset{shared.shufm}{0.72}
\pvset{shared.tgtownm}{1.00}
\pvset{shared.mnull.mamba}{0.33}
\pvset{shared.catcv.mamba.r}{0.99}
\pvset{shared.catcv.mamba.rlo}{0.95}
\pvset{shared.catcv.mamba.rhi}{1.00}
\pvset{shared.catcv.mamba.own}{0.95}
\pvset{shared.catcv.mamba.transp}{0.95}
\pvset{shared.catcv.mamba.bcown}{0.78}
\pvset{shared.catcv.mamba.bctransp}{0.75}
\pvset{shared.catcv.mamba.rho}{0.95}
\pvset{shared.catcv.mamba.rdrop}{0.85}
\pvset{shared.catcv.mamba.permp}{0.002}
\pvset{shared.catcv.mamba.rdroppp}{0.015}
\pvset{shared.catcv.mamba.beatp95}{8}
\pvset{shared.catcv.mamba.nfolds}{8}
\pvset{shared.catcv.mistralmamba.r}{0.99}
\pvset{shared.catcv.mistralmamba.rlo}{0.97}
\pvset{shared.catcv.mistralmamba.rhi}{1.00}
\pvset{shared.catcv.mistralmamba.own}{0.95}
\pvset{shared.catcv.mistralmamba.transp}{0.95}
\pvset{shared.catcv.mistralmamba.bcown}{0.78}
\pvset{shared.catcv.mistralmamba.bctransp}{0.79}
\pvset{shared.catcv.mistralmamba.rho}{0.95}
\pvset{shared.catcv.mistralmamba.rdrop}{0.99}
\pvset{shared.catcv.mistralmamba.permp}{0.000}
\pvset{shared.catcv.mistralmamba.rdroppp}{0.001}
\pvset{shared.catcv.mistralmamba.beatp95}{8}
\pvset{shared.catcv.mistralmamba.nfolds}{8}
\pvset{shared.catcv.beatnull}{16}
\pvset{shared.catcv.beatnulltot}{16}
\pvset{payoff.kmin}{4}
\pvset{payoff.kmax}{64}
\pvset{payoff.nrep}{120}
\pvset{payoff.ncat}{8}
\pvset{payoff.seeds}{15}
\pvset{payoff.ncluster}{8}
\pvset{payoff.k4.transp}{0.78}
\pvset{payoff.k4.scratch}{0.75}
\pvset{payoff.k4.delta}{+0.032}
\pvset{payoff.k4.ci}{[+0.008, +0.057]}
\pvset{payoff.k4.repci}{[+0.014, +0.050]}
\pvset{payoff.k4.win}{64\%}
\pvset{payoff.k4.cats}{7}
\pvset{payoff.k8.transp}{0.83}
\pvset{payoff.k8.scratch}{0.79}
\pvset{payoff.k8.delta}{+0.042}
\pvset{payoff.k8.ci}{[+0.011, +0.073]}
\pvset{payoff.k8.repci}{[+0.021, +0.064]}
\pvset{payoff.k8.win}{68\%}
\pvset{payoff.k8.cats}{8}
\pvset{payoff.k16.transp}{0.87}
\pvset{payoff.k16.scratch}{0.83}
\pvset{payoff.k16.delta}{+0.038}
\pvset{payoff.k16.ci}{[+0.019, +0.058]}
\pvset{payoff.k16.repci}{[+0.027, +0.050]}
\pvset{payoff.k16.win}{78\%}
\pvset{payoff.k16.cats}{8}
\pvset{payoff.k32.transp}{0.90}
\pvset{payoff.k32.scratch}{0.85}
\pvset{payoff.k32.delta}{+0.040}
\pvset{payoff.k32.ci}{[+0.006, +0.073]}
\pvset{payoff.k32.repci}{[+0.029, +0.050]}
\pvset{payoff.k32.win}{72\%}
\pvset{payoff.k32.cats}{7}
\pvset{payoff.k64.transp}{0.92}
\pvset{payoff.k64.scratch}{0.87}
\pvset{payoff.k64.delta}{+0.046}
\pvset{payoff.k64.ci}{[+0.013, +0.080]}
\pvset{payoff.k64.repci}{[+0.035, +0.057]}
\pvset{payoff.k64.win}{78\%}
\pvset{payoff.k64.cats}{7}
\pvset{payoff.k4.tprone.transp}{38.5\%}
\pvset{payoff.k4.tprone.scratch}{32.4\%}
\pvset{payoff.k4.tprone.delta}{$+6.1$ pt}
\pvset{payoff.k4.tprone.deltapts}{6.1}
\pvset{payoff.k4.tprone.fpr}{1.0\%}
\pvset{payoff.k4.tprfive.transp}{46.8\%}
\pvset{payoff.k4.tprfive.scratch}{43.0\%}
\pvset{payoff.k4.tprfive.delta}{$+3.9$ pt}
\pvset{payoff.k4.tprfive.deltapts}{3.9}
\pvset{payoff.k4.tprfive.fpr}{5.0\%}
\pvset{payoff.k8.tprone.transp}{46.1\%}
\pvset{payoff.k8.tprone.scratch}{40.9\%}
\pvset{payoff.k8.tprone.delta}{$+5.2$ pt}
\pvset{payoff.k8.tprone.deltapts}{5.2}
\pvset{payoff.k8.tprone.fpr}{1.0\%}
\pvset{payoff.k8.tprfive.transp}{55.8\%}
\pvset{payoff.k8.tprfive.scratch}{49.8\%}
\pvset{payoff.k8.tprfive.delta}{$+6.0$ pt}
\pvset{payoff.k8.tprfive.deltapts}{6.0}
\pvset{payoff.k8.tprfive.fpr}{5.0\%}
\pvset{payoff.k16.tprone.transp}{54.7\%}
\pvset{payoff.k16.tprone.scratch}{47.7\%}
\pvset{payoff.k16.tprone.delta}{$+7.0$ pt}
\pvset{payoff.k16.tprone.deltapts}{7.0}
\pvset{payoff.k16.tprone.fpr}{1.0\%}
\pvset{payoff.k16.tprfive.transp}{62.9\%}
\pvset{payoff.k16.tprfive.scratch}{55.5\%}
\pvset{payoff.k16.tprfive.delta}{$+7.4$ pt}
\pvset{payoff.k16.tprfive.deltapts}{7.4}
\pvset{payoff.k16.tprfive.fpr}{5.0\%}
\pvset{payoff.k32.tprone.transp}{60.8\%}
\pvset{payoff.k32.tprone.scratch}{56.1\%}
\pvset{payoff.k32.tprone.delta}{$+4.7$ pt}
\pvset{payoff.k32.tprone.deltapts}{4.7}
\pvset{payoff.k32.tprone.fpr}{1.0\%}
\pvset{payoff.k32.tprfive.transp}{68.5\%}
\pvset{payoff.k32.tprfive.scratch}{61.8\%}
\pvset{payoff.k32.tprfive.delta}{$+6.7$ pt}
\pvset{payoff.k32.tprfive.deltapts}{6.7}
\pvset{payoff.k32.tprfive.fpr}{5.0\%}
\pvset{payoff.k64.tprone.transp}{65.7\%}
\pvset{payoff.k64.tprone.scratch}{57.9\%}
\pvset{payoff.k64.tprone.delta}{$+7.9$ pt}
\pvset{payoff.k64.tprone.deltapts}{7.9}
\pvset{payoff.k64.tprone.fpr}{1.0\%}
\pvset{payoff.k64.tprfive.transp}{74.2\%}
\pvset{payoff.k64.tprfive.scratch}{62.8\%}
\pvset{payoff.k64.tprfive.delta}{$+11.4$ pt}
\pvset{payoff.k64.tprfive.deltapts}{11.4}
\pvset{payoff.k64.tprfive.fpr}{5.0\%}
\pvset{payoff.fine.npoints}{14}
\pvset{payoff.fine.kmin}{4}
\pvset{payoff.fine.kmax}{64}
\pvset{payoff.fine.viol}{5}
\pvset{payoff.fine.plateaulo}{0.845}
\pvset{payoff.fine.plateauhi}{0.872}
\pvset{payoff.fine.plateaufrom}{24}
\pvset{payoff.fine.ciw}{0.056}
\pvset{payoff.deltamin}{+0.032}
\pvset{payoff.deltamax}{+0.046}
\pvset{payoff.indist.k}{4}
\pvset{payoff.indist.transp}{1.00}
\pvset{payoff.indist.scratch}{1.00}
\pvset{rwkv.ovr.base}{32.0\%}
\pvset{rwkv.ovr.gated}{32.0\%}
\pvset{rwkv.ovr.delta}{$+0.0$ pt}
\pvset{rwkv.ovr.fire}{0.0\%}
\pvset{rwkv.ovr.n}{100}
\pvset{zamba.ovr.ol.base}{11.0\%}
\pvset{zamba.ovr.ol.mamba}{63.0\%}
\pvset{zamba.ovr.ol.attn}{99.0\%}
\pvset{zamba.ovr.ol.n}{100}
\pvset{zamba.cl.base}{31.7\%}
\pvset{zamba.cl.n}{100}
\pvset{zamba.cl.mamba.asr}{5.0\%}
\pvset{zamba.cl.mamba.ovrbase}{3.0\%}
\pvset{zamba.cl.mamba.ovrgated}{3.0\%}
\pvset{zamba.cl.mamba.ovrdelta}{$+0.0$ pt}
\pvset{zamba.cl.mamba.fireharm}{100.0\%}
\pvset{zamba.cl.mamba.firebenign}{0.0\%}
\pvset{zamba.cl.attn.asr}{0.0\%}
\pvset{zamba.cl.attn.ovrbase}{3.0\%}
\pvset{zamba.cl.attn.ovrgated}{3.0\%}
\pvset{zamba.cl.attn.ovrdelta}{$+0.0$ pt}
\pvset{zamba.cl.attn.fireharm}{100.0\%}
\pvset{zamba.cl.attn.firebenign}{0.0\%}
\pvset{hx.rwkv.n}{100}
\pvset{hx.rwkv.timemix.plain.harm}{100.0\%}
\pvset{hx.rwkv.timemix.plain.spicy}{0.0\%}
\pvset{hx.rwkv.timemix.plain.wharm}{100.0\%}
\pvset{hx.rwkv.timemix.plain.wspicy}{0.0\%}
\pvset{hx.rwkv.timemix.hardened.harm}{100.0\%}
\pvset{hx.rwkv.timemix.hardened.spicy}{0.0\%}
\pvset{hx.rwkv.timemix.hardened.wharm}{100.0\%}
\pvset{hx.rwkv.timemix.hardened.wspicy}{0.0\%}
\pvset{hx.zamba.n}{100}
\pvset{hx.zamba.mamba.plain.harm}{100.0\%}
\pvset{hx.zamba.mamba.plain.spicy}{0.0\%}
\pvset{hx.zamba.mamba.plain.wharm}{100.0\%}
\pvset{hx.zamba.mamba.plain.wspicy}{100.0\%}
\pvset{hx.zamba.mamba.hardened.harm}{98.0\%}
\pvset{hx.zamba.mamba.hardened.spicy}{0.0\%}
\pvset{hx.zamba.mamba.hardened.wharm}{100.0\%}
\pvset{hx.zamba.mamba.hardened.wspicy}{100.0\%}
\pvset{hx.zamba.attn.plain.harm}{100.0\%}
\pvset{hx.zamba.attn.plain.spicy}{0.0\%}
\pvset{hx.zamba.attn.plain.wharm}{100.0\%}
\pvset{hx.zamba.attn.plain.wspicy}{100.0\%}
\pvset{hx.zamba.attn.hardened.harm}{100.0\%}
\pvset{hx.zamba.attn.hardened.spicy}{0.0\%}
\pvset{hx.zamba.attn.hardened.wharm}{100.0\%}
\pvset{hx.zamba.attn.hardened.wspicy}{0.0\%}
\pvset{hx.mistral.n}{100}
\pvset{hx.mistral.attn.plain.harm}{100.0\%}
\pvset{hx.mistral.attn.plain.spicy}{0.0\%}
\pvset{hx.mistral.attn.plain.wharm}{100.0\%}
\pvset{hx.mistral.attn.plain.wspicy}{100.0\%}
\pvset{hx.mistral.attn.hardened.harm}{100.0\%}
\pvset{hx.mistral.attn.hardened.spicy}{0.0\%}
\pvset{hx.mistral.attn.hardened.wharm}{100.0\%}
\pvset{hx.mistral.attn.hardened.wspicy}{0.0\%}
\pvset{hx.llama.n}{100}
\pvset{hx.llama.attn.plain.harm}{100.0\%}
\pvset{hx.llama.attn.plain.spicy}{0.0\%}
\pvset{hx.llama.attn.plain.wharm}{100.0\%}
\pvset{hx.llama.attn.plain.wspicy}{100.0\%}
\pvset{hx.llama.attn.hardened.harm}{100.0\%}
\pvset{hx.llama.attn.hardened.spicy}{0.0\%}
\pvset{hx.llama.attn.hardened.wharm}{100.0\%}
\pvset{hx.llama.attn.hardened.wspicy}{0.0\%}
\pvset{hv.n}{300}
\pvset{hv.nconsensus}{266}
\pvset{hv.nunresolved}{34}
\pvset{hv.irr.comply.agree}{89.0\%}
\pvset{hv.irr.comply.kappa}{0.80}
\pvset{hv.irr.refuse.agree}{90.3\%}
\pvset{hv.irr.refuse.kappa}{0.82}
\pvset{hv.irr.coh.agree}{85.2\%}
\pvset{hv.irr.coh.kappa}{0.24}
\pvset{hv.judge.lg.agree}{94.8\%}
\pvset{hv.judge.lg.kappa}{0.90}
\pvset{hv.judge.lg.n}{97}
\pvset{hv.judge.lg.fp}{5}
\pvset{hv.judge.lg.fn}{0}
\pvset{hv.judge.llama.agree}{93.8\%}
\pvset{hv.judge.llama.kappa}{0.88}
\pvset{hv.judge.llama.n}{97}
\pvset{hv.judge.llama.fp}{5}
\pvset{hv.judge.llama.fn}{1}
\pvset{hv.judge.mistral.agree}{96.9\%}
\pvset{hv.judge.mistral.kappa}{0.94}
\pvset{hv.judge.mistral.n}{97}
\pvset{hv.judge.mistral.fp}{0}
\pvset{hv.judge.mistral.fn}{3}
\pvset{hv.judge.sg.agree}{87.6\%}
\pvset{hv.judge.sg.kappa}{0.74}
\pvset{hv.judge.sg.n}{97}
\pvset{hv.judge.sg.fp}{2}
\pvset{hv.judge.sg.fn}{10}
\pvset{hv.degen.rate}{6.0\%}
\pvset{hv.degen.k}{4}
\pvset{hv.degen.n}{67}
\pvset{hv.cell.ssmbase}{100.0\%}
\pvset{hv.cell.ssmbase.n}{20}
\pvset{hv.cell.ssmgated}{0.0\%}
\pvset{hv.cell.ssmgated.n}{18}
\pvset{hv.cell.rwkvbase}{0.0\%}
\pvset{hv.cell.rwkvbase.n}{30}
\pvset{hv.cell.rwkvtransp}{60.7\%}
\pvset{hv.cell.rwkvtransp.n}{28}
\pvset{hv.cell.rwkvrand}{5.9\%}
\pvset{hv.cell.rwkvrand.n}{17}
\pvset{hv.cell.zattntransp}{14.3\%}
\pvset{hv.cell.zattntransp.n}{14}
\pvset{hv.cell.zattnrand}{6.7\%}
\pvset{hv.cell.zattnrand.n}{15}
\pvset{hv.cell.zmambatransp}{100.0\%}
\pvset{hv.cell.zmambatransp.n}{9}
\pvset{hv.cell.zmambarand}{50.0\%}
\pvset{hv.cell.zmambarand.n}{10}
\pvset{shared.catcv.bt.nfolds}{14}
\pvset{shared.catcv.bt.beatnull}{14}
\pvset{shared.catcv.bt.transp}{1.00}
\pvset{shared.catcv.bt.null}{0.51}
\pvset{w1.base.asr}{100\%}
\pvset{w1.base.xstest}{2.0\%}
\pvset{w1.base.orbench}{32.5\%}
\pvset{w1.detect.asr}{0\%}
\pvset{w1.detect.xstest}{2.0\%}
\pvset{w1.detect.orbench}{47.5\%}
\pvset{w1.gate.alpha}{4}
\pvset{w1.gate.asr}{0\%}
\pvset{w1.gate.xstest}{2.0\%}
\pvset{w1.gate.orbench}{46.3\%}
\pvset{w1.rejudge.n}{50}
\pvset{w1.rejudge.lexasr}{0.0\%}
\pvset{w1.rejudge.lgasr}{0.0\%}
\pvset{w1.rejudge.agree}{100.0\%}
\pvset{w1.rejudge.coh}{100.0\%}
\pvset{w1.detfpr.orbench}{21\%}
\pvset{w1.flip.steer}{94\%}
\pvset{def.ssm.prefix_injection.base.pooled}{98.2\%}
\pvset{def.ssm.prefix_injection.gated.pooled}{3.3\%}
\pvset{def.ssm.pooled.nseeds}{3}
\pvset{def.ssm.pooled.n}{600}
\pvset{adv.k0.firing}{1.00}
\pvset{adv.k0.asr}{5.0\%}
\pvset{adv.k1.firing}{1.00}
\pvset{adv.k1.asr}{6.5\%}
\pvset{adv.k2.firing}{1.00}
\pvset{adv.k2.asr}{2.5\%}
\pvset{adv.k3.firing}{0.96}
\pvset{adv.k3.asr}{4.5\%}
\pvset{adv.k4.firing}{0.81}
\pvset{adv.k4.asr}{19.0\%}
\pvset{adv.k6.firing}{0.30}
\pvset{adv.k6.asr}{67.0\%}
\pvset{adv.k8.firing}{0.06}
\pvset{adv.k8.asr}{90.0\%}
\pvset{adv.sep.kmax}{16}
\pvset{adv.sep.auroc}{1.00}
\pvset{adv.harmlogit.k0}{7.86}
\pvset{adv.harmlogit.kmax}{-3.8}
\pvset{adv.benignlogit}{-23.69}
\pvset{adv.calib.kmax}{16}
\pvset{adv.calib.firing}{1.00}
\pvset{adv.calib.fpr}{1.0\%}
\pvset{cast.a2.asr}{41.0\%}
\pvset{cast.a4.asr}{100.0\%}
\pvset{cast.a8.asr}{97.5\%}
\pvset{cast.a16.asr}{98.0\%}
\pvset{cast.best.alpha}{2}
\pvset{cast.best.asr}{41.0\%}
\pvset{cast.best.overrefusal}{12.0\%}
\pvset{def.n300.prefix_injection.base}{96.3\%}
\pvset{def.n300.prefix_injection.gated}{4.0\%}
\pvset{def.n300.roleplay_dan.base}{83.7\%}
\pvset{def.n300.roleplay_dan.gated}{36.7\%}
\pvset{def.n300.fictional_framing.base}{90.7\%}
\pvset{def.n300.fictional_framing.gated}{15.7\%}
\pvset{def.n300.n}{300}
\pvset{def.ssm.gatedj3}{1.5\%}
\pvset{def.ssm.basej3}{96.0\%}
\pvset{def.ssm.judge3name}{Mistral-7B-Instruct-v0.2}
\pvset{def.ssm.gatedj4}{2.0\%}
\pvset{def.ssm.basej4}{78.0\%}
\pvset{def.ssm.judge4name}{ShieldGemma-9B}
\pvset{def.ssm.judgeagreenway}{72.8\%}
\pvset{def.ssm.njudges}{4}
\pvset{def.hb.prefix_injection.base}{93.5\%}
\pvset{def.hb.prefix_injection.gated}{5.5\%}
\pvset{def.hb.roleplay_dan.base}{86.5\%}
\pvset{def.hb.roleplay_dan.gated}{33.5\%}
\pvset{def.hb.fictional_framing.base}{88.0\%}
\pvset{def.hb.fictional_framing.gated}{9.0\%}
\pvset{def.hb.base}{93.5\%}
\pvset{def.hb.gated}{5.5\%}
\pvset{def.hb.n}{200}
\pvset{def.opt.base}{97.5\%}
\pvset{def.opt.gated}{6.5\%}
\pvset{def.opt.cast}{100.0\%}
\pvset{def.opt.n}{200}
\pvset{def.detam.llama.base}{28.0\%}
\pvset{def.detam.llama.ours}{2.0\%}
\pvset{def.detam.llama.ours.overref}{10.0\%}
\pvset{def.detam.llama.open}{2.0\%}
\pvset{def.detam.llama.open.overref}{87.0\%}
\pvset{def.detam.llama.prefix.base}{28.0\%}
\pvset{def.detam.llama.prefix.gated}{2.0\%}
\pvset{def.detam.llama.prefix.ouror}{10.0\%}
\pvset{def.detam.llama.prefix.detamor}{87.0\%}
\pvset{def.detam.llama.roleplay.base}{0.5\%}
\pvset{def.detam.llama.roleplay.gated}{0.0\%}
\pvset{def.detam.llama.roleplay.ouror}{10.0\%}
\pvset{def.detam.llama.roleplay.detamor}{87.0\%}
\pvset{def.detam.llama.fiction.base}{19.5\%}
\pvset{def.detam.llama.fiction.gated}{1.0\%}
\pvset{def.detam.llama.fiction.ouror}{10.0\%}
\pvset{def.detam.llama.fiction.detamor}{87.0\%}
\pvset{def.detam.mistral.base}{96.0\%}
\pvset{def.detam.mistral.ours}{15.0\%}
\pvset{def.detam.mistral.ours.overref}{11.0\%}
\pvset{def.detam.mistral.open}{15.0\%}
\pvset{def.detam.mistral.open.overref}{69.0\%}
\pvset{def.detam.mistral.prefix.base}{96.0\%}
\pvset{def.detam.mistral.prefix.gated}{15.0\%}
\pvset{def.detam.mistral.prefix.ouror}{11.0\%}
\pvset{def.detam.mistral.prefix.detamor}{69.0\%}
\pvset{def.detam.mistral.roleplay.base}{63.5\%}
\pvset{def.detam.mistral.roleplay.gated}{78.0\%}
\pvset{def.detam.mistral.roleplay.ouror}{11.0\%}
\pvset{def.detam.mistral.roleplay.detamor}{69.0\%}
\pvset{def.detam.mistral.fiction.base}{96.5\%}
\pvset{def.detam.mistral.fiction.gated}{81.5\%}
\pvset{def.detam.mistral.fiction.ouror}{11.0\%}
\pvset{def.detam.mistral.fiction.detamor}{69.0\%}
\pvset{def.detam.nharm}{200}
\pvset{def.detam.nbenign}{100}
\pvset{def.detam.base}{28.0\%}
\pvset{def.detam.ours}{2.0\%}
\pvset{def.detam.ours.overref}{10.0\%}
\pvset{def.detam.open}{2.0\%}
\pvset{def.detam.open.overref}{87.0\%}
\pvset{def.gcg.base}{41.0\%}
\pvset{def.gcg.gated}{18.0\%}
\pvset{def.gcg.overref}{11.0\%}
\pvset{def.gcg.basej4}{16.0\%}
\pvset{def.gcg.gatedj4}{5.0\%}
\pvset{def.gcg.n}{100}
\pvset{def.gcgssm.base}{22.0\%}
\pvset{def.gcgssm.gated}{6.0\%}
\pvset{def.gcgssm.basej4}{15.0\%}
\pvset{def.gcgssm.gatedj4}{3.0\%}
\pvset{def.gcgssm.judge4name}{ShieldGemma-9B}
\pvset{def.gcgssm.n}{100}
\pvset{def.gcg.qwen.base}{6.7\%}
\pvset{def.gcg.llama.base}{0.0\%}
\pvset{def.gcg.ga.base}{42.0\%}
\pvset{def.gcg.ga.attn}{44.0\%}
\pvset{def.gcg.ga.glob}{23.0\%}
\pvset{def.gcg.ga.basej4}{16.0\%}
\pvset{def.gcg.ga.attnj4}{12.0\%}
\pvset{def.gcg.ga.globj4}{5.0\%}
\pvset{def.gcg.ga.globoverref}{69.0\%}
\pvset{def.gcg.ga.n}{100}
\pvset{def.gcg.pooled.base}{43.3\%}
\pvset{def.gcg.pooled.basej4}{17.7\%}
\pvset{def.gcg.pooled.gated}{18.7\%}
\pvset{def.gcg.pooled.gatedj4}{3.7\%}
\pvset{def.gcg.pooled.nseeds}{3}
\pvset{def.gcg.pooled.n}{300}
\pvset{def.gcg.ga.pooled.base}{41.7\%}
\pvset{def.gcg.ga.pooled.basej4}{16.0\%}
\pvset{def.gcg.ga.pooled.attn}{43.3\%}
\pvset{def.gcg.ga.pooled.attnj4}{15.3\%}
\pvset{def.gcg.ga.pooled.glob}{18.0\%}
\pvset{def.gcg.ga.pooled.globj4}{2.3\%}
\pvset{def.gcg.ga.pooled.nseeds}{3}
\pvset{def.gcg.ga.pooled.n}{300}
\pvset{def.refuse.prefix.asr}{0.0\%}
\pvset{def.refuse.prefix.overref}{12.0\%}
\pvset{def.refuse.roleplay.asr}{0.0\%}
\pvset{def.refuse.roleplay.overref}{12.0\%}
\pvset{def.refuse.fiction.asr}{0.0\%}
\pvset{def.refuse.fiction.overref}{12.0\%}
\pvset{def.mlp.base}{98.3\%}
\pvset{def.mlp.gated}{6.7\%}
\pvset{def.mlp.cast}{100.0\%}
\pvset{def.attnsweep.nharm}{100}
\pvset{def.attnsweep.nbenign}{60}
\pvset{sweep.llama.base}{26.0\%}
\pvset{sweep.llama.a0p5.alpha}{0.5}
\pvset{sweep.llama.a0p5.gated}{7.0\%}
\pvset{sweep.llama.a0p5.detam}{7.0\%}
\pvset{sweep.llama.a0p5.cohharm}{100.0\%}
\pvset{sweep.llama.a0p5.cohbenign}{100.0\%}
\pvset{sweep.llama.a0p5.detambenign}{100.0\%}
\pvset{sweep.llama.a1p0.alpha}{1.0}
\pvset{sweep.llama.a1p0.gated}{2.0\%}
\pvset{sweep.llama.a1p0.detam}{2.0\%}
\pvset{sweep.llama.a1p0.cohharm}{100.0\%}
\pvset{sweep.llama.a1p0.cohbenign}{100.0\%}
\pvset{sweep.llama.a1p0.detambenign}{100.0\%}
\pvset{sweep.llama.a1p5.alpha}{1.5}
\pvset{sweep.llama.a1p5.gated}{1.0\%}
\pvset{sweep.llama.a1p5.detam}{1.0\%}
\pvset{sweep.llama.a1p5.cohharm}{100.0\%}
\pvset{sweep.llama.a1p5.cohbenign}{100.0\%}
\pvset{sweep.llama.a1p5.detambenign}{100.0\%}
\pvset{sweep.llama.a2p0.alpha}{2.0}
\pvset{sweep.llama.a2p0.gated}{1.0\%}
\pvset{sweep.llama.a2p0.detam}{1.0\%}
\pvset{sweep.llama.a2p0.cohharm}{100.0\%}
\pvset{sweep.llama.a2p0.cohbenign}{100.0\%}
\pvset{sweep.llama.a2p0.detambenign}{100.0\%}
\pvset{sweep.llama.a3p0.alpha}{3.0}
\pvset{sweep.llama.a3p0.gated}{3.0\%}
\pvset{sweep.llama.a3p0.detam}{3.0\%}
\pvset{sweep.llama.a3p0.cohharm}{80.0\%}
\pvset{sweep.llama.a3p0.cohbenign}{100.0\%}
\pvset{sweep.llama.a3p0.detambenign}{100.0\%}
\pvset{sweep.llama.a4p0.alpha}{4.0}
\pvset{sweep.llama.a4p0.gated}{0.0\%}
\pvset{sweep.llama.a4p0.detam}{0.0\%}
\pvset{sweep.llama.a4p0.cohharm}{63.0\%}
\pvset{sweep.llama.a4p0.cohbenign}{100.0\%}
\pvset{sweep.llama.a4p0.detambenign}{53.3\%}
\pvset{sweep.llama.gated}{1.0\%}
\pvset{def.llamaattn.prefix.base}{25.0\%}
\pvset{def.llamaattn.prefix.gated}{0.0\%}
\pvset{def.llamaattn.prefix.hardened}{yes}
\pvset{def.llamaattn.prefix.overref}{0.0\%}
\pvset{def.llamaattn.prefix.overrefgated}{0.0\%}
\pvset{def.llamaattn.prefix.overrefdelta}{$+0.0$ pt}
\pvset{sweep.mistral.base}{97.0\%}
\pvset{sweep.mistral.a0p5.alpha}{0.5}
\pvset{sweep.mistral.a0p5.gated}{98.0\%}
\pvset{sweep.mistral.a0p5.detam}{98.0\%}
\pvset{sweep.mistral.a0p5.cohharm}{100.0\%}
\pvset{sweep.mistral.a0p5.cohbenign}{100.0\%}
\pvset{sweep.mistral.a0p5.detambenign}{100.0\%}
\pvset{sweep.mistral.a1p0.alpha}{1.0}
\pvset{sweep.mistral.a1p0.gated}{88.0\%}
\pvset{sweep.mistral.a1p0.detam}{88.0\%}
\pvset{sweep.mistral.a1p0.cohharm}{100.0\%}
\pvset{sweep.mistral.a1p0.cohbenign}{100.0\%}
\pvset{sweep.mistral.a1p0.detambenign}{100.0\%}
\pvset{sweep.mistral.a1p5.alpha}{1.5}
\pvset{sweep.mistral.a1p5.gated}{39.0\%}
\pvset{sweep.mistral.a1p5.detam}{39.0\%}
\pvset{sweep.mistral.a1p5.cohharm}{100.0\%}
\pvset{sweep.mistral.a1p5.cohbenign}{100.0\%}
\pvset{sweep.mistral.a1p5.detambenign}{100.0\%}
\pvset{sweep.mistral.a2p0.alpha}{2.0}
\pvset{sweep.mistral.a2p0.gated}{15.0\%}
\pvset{sweep.mistral.a2p0.detam}{15.0\%}
\pvset{sweep.mistral.a2p0.cohharm}{100.0\%}
\pvset{sweep.mistral.a2p0.cohbenign}{100.0\%}
\pvset{sweep.mistral.a2p0.detambenign}{100.0\%}
\pvset{sweep.mistral.a3p0.alpha}{3.0}
\pvset{sweep.mistral.a3p0.gated}{78.0\%}
\pvset{sweep.mistral.a3p0.detam}{78.0\%}
\pvset{sweep.mistral.a3p0.cohharm}{89.0\%}
\pvset{sweep.mistral.a3p0.cohbenign}{100.0\%}
\pvset{sweep.mistral.a3p0.detambenign}{98.3\%}
\pvset{sweep.mistral.a4p0.alpha}{4.0}
\pvset{sweep.mistral.a4p0.gated}{87.0\%}
\pvset{sweep.mistral.a4p0.detam}{87.0\%}
\pvset{sweep.mistral.a4p0.cohharm}{40.0\%}
\pvset{sweep.mistral.a4p0.cohbenign}{100.0\%}
\pvset{sweep.mistral.a4p0.detambenign}{20.0\%}
\pvset{sweep.mistral.gated}{15.0\%}
\pvset{def.mistralattn.prefix.base}{98.0\%}
\pvset{def.mistralattn.prefix.gated}{14.0\%}
\pvset{def.mistralattn.prefix.hardened}{yes}
\pvset{def.mistralattn.prefix.overref}{0.0\%}
\pvset{def.mistralattn.prefix.overrefgated}{0.0\%}
\pvset{def.mistralattn.prefix.overrefdelta}{$+0.0$ pt}
\pvset{adv.opt.firing.clean}{1.00}
\pvset{adv.opt.firing.optimized}{0.00}
\pvset{adv.opt.asr.clean}{6.7\%}
\pvset{adv.opt.asr.optimized}{95.0\%}
\pvset{pve.mean.evasion.fire}{1.00}
\pvset{pve.mean.clean.med}{7.8}
\pvset{pve.mean.evasion.med}{-7.7}
\pvset{pve.max.evasion.fire}{1.00}
\pvset{pve.max.clean.med}{45.7}
\pvset{pve.max.evasion.med}{46.0}
\pvset{pve.wavg.evasion.fire}{1.00}
\pvset{pve.wavg.clean.med}{45.7}
\pvset{pve.wavg.evasion.med}{46.0}
\pvset{pve.fpr}{1\%}
\pvset{pveadv.wavg.fire}{1.00}
\pvset{pveadv.wavg.med.clean}{45.7}
\pvset{pveadv.wavg.med.attack}{46.0}
\pvset{coh.s2x.transferred}{77.0\%}
\pvset{coh.s2x.native}{93.0\%}
\pvset{coh.ma2mi.transferred}{100.0\%}
\pvset{coh.ma2mi.native}{100.0\%}
\pvset{coh.f32l.transferred}{82.5\%}
\pvset{coh.f32l.native}{93.0\%}
\pvset{def.zamba.base}{31.7\%}
\pvset{def.zamba.attn}{0.0\%}
\pvset{def.zamba.residual}{36.7\%}
\pvset{def.zamba.mamba.uncal}{40.0\%}
\pvset{def.zamba.mamba.swbase}{27.5\%}
\pvset{def.zamba.mamba.best}{5.0\%}
\pvset{def.zamba.mamba.alpha}{16}
\pvset{zamba2p7b.n}{60}
\pvset{zamba2p7b.rel}{1.0}
\pvset{zamba2p7b.base}{98.3\%}
\pvset{zamba2p7b.mamba.gated}{81.7\%}
\pvset{zamba2p7b.mamba.coh}{96.7\%}
\pvset{zamba2p7b.residual.gated}{93.3\%}
\pvset{zamba2p7b.residual.coh}{0.0\%}
\pvset{zamba7b.residual.n}{60}
\pvset{zamba7b.residual.relearly}{0.2}
\pvset{zamba7b.residual.relbreak}{0.3}
\pvset{zamba7b.residual.base}{36.7\%}
\pvset{zamba7b.residual.early.gated}{1.7\%}
\pvset{zamba7b.residual.early.coh}{100.0\%}
\pvset{zamba7b.residual.break.gated}{88.3\%}
\pvset{zamba7b.residual.break.coh}{98.3\%}
\pvset{roleplay.gated.refuse}{169}
\pvset{roleplay.gated.comply}{28}
\pvset{roleplay.base.refuse}{13}
\pvset{roleplay.n}{200}
\pvset{shared.headroom.tgtown}{1.00}
\pvset{shared.headroom.proc}{1.00}
\pvset{shared.headroom.shuf}{0.23}
\pvset{pool.kmax}{256}
\pvset{pool.fpr}{1.0\%}
\pvset{pool.mean.firing}{1.00}
\pvset{pool.wavg.firing}{1.00}
\pvset{pool.mean.score.kmax}{0}
\pvset{pool.wavg.score.kmax}{57}
\pvset{pool.theta.mean}{-18}
\pvset{pool.theta.wavg}{6}
\pvset{roleplay.fix.refusal}{38.0\%}
\pvset{roleplay.fix.persona}{79.5\%}
\pvset{roleplay.fix.combined}{17.5\%}
\pvset{roleplay.fix.benignfire}{0.0\%}
\pvset{roleplay.fix.overrefmarg}{0.0\%}
\pvset{roleplay.fix.baseoverref}{12.0\%}
\pvset{roleplay.fix.cos}{-0.30}
\pvset{roleplay.cos.mamba}{-0.30}
\pvset{roleplay.cos.mamba3}{-0.30}
\pvset{roleplay.cos.rwkv}{-0.49}
\pvset{roleplay.cos.llama}{+0.65}
\pvset{roleplay.cos.mistral}{+0.74}
\pvset{roleplay.cos.ssmlo}{-0.49}
\pvset{roleplay.cos.ssmhi}{+0.02}
\pvset{roleplay.cos.xfmrlo}{+0.58}
\pvset{roleplay.cos.xfmrhi}{+0.82}
\pvset{len256.pi.base}{98.0\%}
\pvset{len256.rp.base}{82.5\%}
\pvset{len256.base.agree}{0.96}
\pvset{len256.pi.a4}{17.5\%}
\pvset{len256.rp.a4}{55.0\%}
\pvset{len256.a4.agree}{0.96}
\pvset{len256.cohpi.base}{99.5\%}
\pvset{len256.cohrp.base}{100.0\%}
\pvset{len256.cohpi.a4}{10.5\%}
\pvset{len256.cohrp.a4}{23.5\%}
\pvset{len256.pi.a1}{99.0\%}
\pvset{len256.rp.a1}{89.5\%}
\pvset{len256.a1.agree}{0.96}
\pvset{len256.cohpi.a1}{99.0\%}
\pvset{len256.cohrp.a1}{99.5\%}
\pvset{len256.pi.a2}{98.0\%}
\pvset{len256.rp.a2}{91.5\%}
\pvset{len256.a2.agree}{0.97}
\pvset{len256.cohpi.a2}{99.5\%}
\pvset{len256.cohrp.a2}{98.5\%}
\pvset{len256.pi.a6}{48.5\%}
\pvset{len256.rp.a6}{60.5\%}
\pvset{len256.a6.agree}{0.77}
\pvset{len256.cohpi.a6}{0.0\%}
\pvset{len256.cohrp.a6}{0.0\%}
\pvset{coh.f3.outlayers.a4}{65.5\%}
\pvset{coh.f3.outlayers.a1}{100.0\%}
\pvset{coh.f3.deltalayers.mind2}{0.997}
\pvset{site.f3.base}{56.0\%}
\pvset{site.f3.block}{4.5\%}
\pvset{site.f3.residual}{45.0\%}
\pvset{site.f3.overrefusal}{16.0\%}
\pvset{site.f3.a1.base}{56.0\%}
\pvset{site.f3.a1.block}{32.5\%}
\pvset{site.f3.a1.residual}{45.0\%}
\pvset{site.f3.a1.overrefusal}{16.0\%}
\pvset{site.f3.outlayers.base}{56.0\%}
\pvset{site.f3.outlayers.block}{0.0\%}
\pvset{site.f3.outlayers.residual}{94.0\%}
\pvset{site.f3.outlayers.overrefusal}{16.0\%}
\pvset{roleplay.mistral.base}{86.0\%}
\pvset{roleplay.mistral.fix}{51.0\%}
\pvset{shared.rwkv.tgtown}{1.00}
\pvset{shared.rwkv.proc}{1.00}
\pvset{shared.rwkv.shuf}{0.43}
\pvset{shared.rwkv.proc.ci}{[1.00, 1.00]}
\pvset{shared.rwkv.shuf.ci}{[0.33, 0.53]}
\pvset{shared.zamba.tgtown}{1.00}
\pvset{shared.zamba.proc}{1.00}
\pvset{shared.zamba.shuf}{0.52}
\pvset{shared.zamba.shufproc}{0.08}
\pvset{shared.zamba.proc.ci}{[1.00, 1.00]}
\pvset{shared.zamba.shuf.ci}{[0.41, 0.63]}
\pvset{orth.rwkv.base}{3.0\%}
\pvset{orth.rwkv.transp}{58.0\%}
\pvset{orth.rwkv.rand}{40.0\%}
\pvset{orth.rwkv.coscheck}{1.00}
\pvset{orth.rwkv.pooled.n}{300}
\pvset{orth.rwkv.pooled.base}{3.0\%}
\pvset{orth.rwkv.pooled.transp}{54.7\%}
\pvset{orth.rwkv.pooled.rand}{32.3\%}
\pvset{orth.rwkv.pooled.lex.transp}{28.3\%}
\pvset{orth.rwkv.pooled.lex.rand}{0.7\%}
\pvset{orth.zamba.attn.base}{11.7\%}
\pvset{orth.zamba.attn.transp}{20.0\%}
\pvset{orth.zamba.attn.rand}{5.0\%}
\pvset{orth.zamba.attn.pooled.n}{180}
\pvset{orth.zamba.attn.pooled.base}{8.9\%}
\pvset{orth.zamba.attn.pooled.transp}{16.7\%}
\pvset{orth.zamba.attn.pooled.rand}{8.9\%}
\pvset{orth.zamba.mamba.alpha}{12}
\pvset{orth.zamba.mamba.base}{11.7\%}
\pvset{orth.zamba.mamba.transp}{91.7\%}
\pvset{orth.zamba.mamba.transpcoh}{88.3\%}
\pvset{orth.zamba.mamba.rand}{25.0\%}
\pvset{orth.zamba.mamba.randcoh}{100.0\%}
\pvset{orth.zamba.mamba.pooled.n}{180}
\pvset{orth.zamba.mamba.pooled.base}{8.9\%}
\pvset{orth.zamba.mamba.pooled.transp}{93.3\%}
\pvset{orth.zamba.mamba.pooled.rand}{22.8\%}
\pvset{orth.zamba.mamba.pooled.lex.transp}{99.4\%}
\pvset{orth.zamba.mamba.pooled.lex.rand}{26.7\%}
\pvset{orth.zamba.mamba.cohthresh}{80.0\%}
\pvset{orth.zamba.mamba.a8.transp}{60.0\%}
\pvset{orth.zamba.mamba.a8.rand}{5.0\%}
\pvset{orth.zamba.mamba.a16.transp}{100.0\%}
\pvset{orth.zamba.mamba.a16.transpcoh}{68.3\%}
\pvset{orth.zamba.mamba.a16.rand}{85.0\%}
\pvset{orth.zamba.mamba.a16.randcoh}{41.7\%}
\pvset{orth.zamba.mamba.sweep.judgedlist}{8, 12}
\pvset{orth.zamba.mamba.sweep.a1.transp}{13.3\%}
\pvset{orth.zamba.mamba.sweep.a1.rand}{13.3\%}
\pvset{orth.zamba.mamba.sweep.a1.transpcoh}{100.0\%}
\pvset{orth.zamba.mamba.sweep.a1.randcoh}{100.0\%}
\pvset{orth.zamba.mamba.sweep.a2.transp}{11.7\%}
\pvset{orth.zamba.mamba.sweep.a2.rand}{15.0\%}
\pvset{orth.zamba.mamba.sweep.a2.transpcoh}{100.0\%}
\pvset{orth.zamba.mamba.sweep.a2.randcoh}{100.0\%}
\pvset{orth.zamba.mamba.sweep.a4.transp}{20.0\%}
\pvset{orth.zamba.mamba.sweep.a4.rand}{13.3\%}
\pvset{orth.zamba.mamba.sweep.a4.transpcoh}{100.0\%}
\pvset{orth.zamba.mamba.sweep.a4.randcoh}{100.0\%}
\pvset{orth.zamba.mamba.sweep.a8.transp}{60.0\%}
\pvset{orth.zamba.mamba.sweep.a8.rand}{5.0\%}
\pvset{orth.zamba.mamba.sweep.a8.transpcoh}{100.0\%}
\pvset{orth.zamba.mamba.sweep.a8.randcoh}{100.0\%}
\pvset{orth.zamba.mamba.sweep.a12.transp}{91.7\%}
\pvset{orth.zamba.mamba.sweep.a12.rand}{25.0\%}
\pvset{orth.zamba.mamba.sweep.a12.transpcoh}{88.3\%}
\pvset{orth.zamba.mamba.sweep.a12.randcoh}{100.0\%}
\pvset{orth.zamba.mamba.sweep.a16.transp}{100.0\%}
\pvset{orth.zamba.mamba.sweep.a16.rand}{85.0\%}
\pvset{orth.zamba.mamba.sweep.a16.transpcoh}{68.3\%}
\pvset{orth.zamba.mamba.sweep.a16.randcoh}{41.7\%}
\pvset{orth.zamba.mamba.sweep.a24.transp}{100.0\%}
\pvset{orth.zamba.mamba.sweep.a24.rand}{98.3\%}
\pvset{orth.zamba.mamba.sweep.a24.transpcoh}{26.7\%}
\pvset{orth.zamba.mamba.sweep.a24.randcoh}{66.7\%}
\pvset{coh.screen.n}{750}
\pvset{coh.screen.agree}{88.1\%}
\pvset{coh.screen.opcheck}{the same}
\pvset{wsm.nmamba}{68}
\pvset{wsm.nattnfire}{13}
\pvset{wsm.attn.mean}{0.28}
\pvset{wsm.mamba.mean}{0.37}
\pvset{wsm.attn.cum}{3.6}
\pvset{wsm.mamba.cum}{24.9}
\pvset{wsm.ratio}{6.9}
\pvset{wsm.mamba.min}{0.11}
\pvset{wsm.mamba.max}{0.57}
\pvset{rwkv.def.base}{51.7\%}
\pvset{rwkv.def.gated}{48.3\%}
\pvset{rwkv.site.base}{47.5\%}
\pvset{rwkv.site.alpha}{16}
\pvset{rwkv.site.timemix.gated}{27.5\%}
\pvset{rwkv.site.timemix.rand}{52.5\%}
\pvset{rwkv.site.timemix.coh}{97.5\%}
\pvset{rwkv.site.timemix.randdeg.alpha}{48}
\pvset{rwkv.site.timemix.randdeg.coh}{22.5\%}
\pvset{rwkv.site.chanmix.gated}{10.0\%}
\pvset{rwkv.site.chanmix.rand}{37.5\%}
\pvset{rwkv.site.chanmix.coh}{100.0\%}
\pvset{rwkv.site.chanmix.randdeg.alpha}{48}
\pvset{rwkv.site.chanmix.randdeg.coh}{32.5\%}
\pvset{rwkv.block.base}{56.0\%}
\pvset{rwkv.block.gated}{52.0\%}
\pvset{rwkv.nm.base}{47.5\%}
\pvset{rwkv.nm.rel}{0.25}
\pvset{rwkv.nm.block.norm}{519}
\pvset{rwkv.nm.block.relmax}{4}
\pvset{rwkv.nm.block.gated}{47.5\%}
\pvset{rwkv.nm.block.rand}{47.5\%}
\pvset{rwkv.nm.block.randcoh}{97.5\%}
\pvset{rwkv.nm.timemix.norm}{27}
\pvset{rwkv.nm.timemix.gated}{17.5\%}
\pvset{rwkv.nm.timemix.rand}{82.5\%}
\pvset{rwkv.nm.timemix.gatedcoh}{97.5\%}
\pvset{rwkv.nm.timemix.randcoh}{97.5\%}
\pvset{rwkv.nm.timemix.best}{0.0\%}
\pvset{rwkv.nm.timemix.bestrel}{1}
\pvset{rwkv.nm.chanmix.norm}{54}
\pvset{rwkv.nm.chanmix.gated}{7.5\%}
\pvset{rwkv.nm.chanmix.rand}{15.0\%}
\pvset{rwkv.nm.chanmix.gatedcoh}{100.0\%}
\pvset{rwkv.nm.chanmix.randcoh}{85.0\%}
\pvset{rwkv.nm.chanmix.best}{0.0\%}
\pvset{rwkv.nm.chanmix.bestrel}{0.5}
\pvset{ssm.nm.base}{100.0\%}
\pvset{ssm.nm.mixer.norm}{1.8}
\pvset{ssm.nm.residual.norm}{13.8}
\pvset{ssm.nm.ratio}{7.8}
\pvset{ssm.nm.mixer.rel}{1}
\pvset{ssm.nm.mixer.gated}{0.0\%}
\pvset{ssm.nm.mixer.gatedcoh}{97.5\%}
\pvset{ssm.nm.residual.rel}{0.25}
\pvset{ssm.nm.residual.gated}{100.0\%}
\pvset{ssm.nm.residual.gatedcoh}{0.0\%}
\pvset{xfmr.nm.base}{100.0\%}
\pvset{xfmr.nm.attn.norm}{0.79}
\pvset{xfmr.nm.residual.norm}{22.0}
\pvset{xfmr.nm.ratio}{27.9}
\pvset{xfmr.nm.attn.rel}{1}
\pvset{xfmr.nm.attn.gated}{85.0\%}
\pvset{xfmr.nm.attn.gatedcoh}{100.0\%}
\pvset{xfmr.nm.residual.rel}{0.25}
\pvset{xfmr.nm.residual.gated}{100.0\%}
\pvset{xfmr.nm.residual.gatedcoh}{0.0\%}
\pvset{xfmr.mlp.norm}{1.33}
\pvset{xfmr.mlp.base}{100.0\%}
\pvset{xfmr.mlp.rel}{0.5}
\pvset{xfmr.mlp.gated}{0.0\%}
\pvset{xfmr.mlp.gatedcoh}{100.0\%}
\pvset{llama.nm.base}{27.5\%}
\pvset{llama.nm.attn.norm}{0.78}
\pvset{llama.nm.residual.norm}{23.0}
\pvset{llama.nm.attn.rel}{0.5}
\pvset{llama.nm.attn.gated}{0.0\%}
\pvset{llama.nm.attn.gatedcoh}{100.0\%}
\pvset{llama.nm.residual.rel}{0.25}
\pvset{llama.nm.residual.gated}{100.0\%}
\pvset{llama.nm.residual.gatedcoh}{0.0\%}
\pvset{llama.mlp.norm}{1.69}
\pvset{llama.mlp.base}{27.5\%}
\pvset{llama.mlp.rel}{0.25}
\pvset{llama.mlp.gated}{2.5\%}
\pvset{llama.mlp.gatedcoh}{100.0\%}
\pvset{ssm.dskip.n}{40}
\pvset{ssm.dskip.base}{100.0\%}
\pvset{ssm.dskip.base.over}{13.3\%}
\pvset{ssm.dskip.r0p25.asr}{100.0\%}
\pvset{ssm.dskip.r0p25.over}{20.0\%}
\pvset{ssm.dskip.r0p25.coh}{100.0\%}
\pvset{ssm.dskip.r0p5.asr}{92.5\%}
\pvset{ssm.dskip.r0p5.over}{13.3\%}
\pvset{ssm.dskip.r0p5.coh}{100.0\%}
\pvset{ssm.dskip.r1p0.asr}{77.5\%}
\pvset{ssm.dskip.r1p0.over}{20.0\%}
\pvset{ssm.dskip.r1p0.coh}{100.0\%}
\pvset{ssm.dskip.r1p25.asr}{65.0\%}
\pvset{ssm.dskip.r1p25.over}{33.3\%}
\pvset{ssm.dskip.r1p25.coh}{100.0\%}
\pvset{ssm.dskip.r1p5.asr}{57.5\%}
\pvset{ssm.dskip.r1p5.over}{36.7\%}
\pvset{ssm.dskip.r1p5.coh}{100.0\%}
\pvset{ssm.dskip.r1p75.asr}{32.5\%}
\pvset{ssm.dskip.r1p75.over}{46.7\%}
\pvset{ssm.dskip.r1p75.coh}{98.6\%}
\pvset{ssm.dskip.r2p0.asr}{17.5\%}
\pvset{ssm.dskip.r2p0.over}{43.3\%}
\pvset{ssm.dskip.r2p0.coh}{98.6\%}
\pvset{ssm.dskip.r4p0.asr}{100.0\%}
\pvset{ssm.dskip.r4p0.over}{0.0\%}
\pvset{ssm.dskip.r4p0.coh}{17.1\%}
\pvset{ssm.dskip.best.rel}{2}
\pvset{ssm.dskip.best.asr}{17.5\%}
\pvset{ssm.dskip.best.over}{43.3\%}
\pvset{ssm.dskip.best.coh}{98.6\%}
\pvset{ssm.mixin.n}{40}
\pvset{ssm.mixin.base}{100.0\%}
\pvset{ssm.mixin.base.over}{13.3\%}
\pvset{ssm.mixin.r0p25.asr}{87.5\%}
\pvset{ssm.mixin.r0p25.over}{46.7\%}
\pvset{ssm.mixin.r0p25.coh}{100.0\%}
\pvset{ssm.mixin.r0p5.asr}{57.5\%}
\pvset{ssm.mixin.r0p5.over}{100.0\%}
\pvset{ssm.mixin.r0p5.coh}{100.0\%}
\pvset{ssm.mixin.r1p0.asr}{100.0\%}
\pvset{ssm.mixin.r1p0.over}{0.0\%}
\pvset{ssm.mixin.r1p0.coh}{51.4\%}
\pvset{ssm.mixin.r2p0.asr}{100.0\%}
\pvset{ssm.mixin.r2p0.over}{0.0\%}
\pvset{ssm.mixin.r2p0.coh}{87.1\%}
\pvset{ssm.mixin.r4p0.asr}{100.0\%}
\pvset{ssm.mixin.r4p0.over}{0.0\%}
\pvset{ssm.mixin.r4p0.coh}{0.0\%}
\pvset{ssm.mixin.topcoh.rel}{0.5}
\pvset{ssm.mixin.topcoh.asr}{57.5\%}
\pvset{ssm.mixin.topcoh.over}{100.0\%}
\pvset{persona.rwkv.alpha}{16}
\pvset{persona.rwkv.beta}{1}
\pvset{persona.rwkv.closed.base}{19.0\%}
\pvset{persona.rwkv.fire.plain}{100.0\%}
\pvset{persona.rwkv.fire.rp}{0.0\%}
\pvset{persona.rwkv.score.plain}{1.000}
\pvset{persona.rwkv.score.rp}{0.011}
\pvset{persona.rwkv.dt.srcauroc}{1.00}
\pvset{persona.rwkv.dt.native.plain}{1.000}
\pvset{persona.rwkv.dt.native.rp}{0.002}
\pvset{persona.rwkv.dt.trans.plain}{0.994}
\pvset{persona.rwkv.dt.trans.rp}{0.000}
\pvset{persona.rwkv.ol.base}{19.0\%}
\pvset{persona.rwkv.ol.refusal}{8.0\%}
\pvset{persona.rwkv.ol.persona}{77.5\%}
\pvset{persona.rwkv.ol.combined}{57.0\%}
\pvset{persona.rwkv.ol.combined.overrefmarg}{8.0\%}
\pvset{persona.zamba.alpha}{16}
\pvset{persona.zamba.beta}{1}
\pvset{persona.zamba.base}{0.0\%}
\pvset{persona.zamba.refusal}{5.0\%}
\pvset{persona.zamba.persona}{0.0\%}
\pvset{persona.zamba.combined}{8.3\%}
\pvset{persona.zamba.combinedcoh}{98.3\%}
\pvset{persona.zamba.diff.refusal}{60}
\pvset{persona.zamba.diff.persona}{0}
\pvset{persona.zamba.diff.combined}{60}
\pvset{persona.zamba.diff.n}{60}
\pvset{persona.zamba.sweep.lastzero.alpha}{6}
\pvset{persona.zamba.sweep.n}{60}
\pvset{persona.zamba.sweep.a12.b0}{26.7\%}
\pvset{persona.zamba.sweep.a12.b2}{18.3\%}
\pvset{persona.zamba.sweep.a12.b2.coh}{98.3\%}
\pvset{def.combo.mamba.base}{98.0\%}
\pvset{def.combo.mamba.site}{11.0\%}
\pvset{def.combo.mamba.residual}{100.0\%}
\pvset{def.combo.mistral.base}{97.0\%}
\pvset{def.combo.llama.base}{1.0\%}
\pvset{def.combo.mistral.refusal}{77.0\%}
\pvset{def.combo.mistral.personaonly}{96.0\%}
\pvset{def.combo.mistral.combined}{5.0\%}
\pvset{def.combo.mistral.basesg}{79.0\%}
\pvset{def.combo.mistral.combinedsg}{0.0\%}
\pvset{def.combo.cossite}{-0.02}
\pvset{def.combo.cosresidual}{+0.76}
\pvset{def.combo.mistral.combinedres}{100.0\%}
\pvset{persona.geom.mi.rp.res}{+0.74}
\pvset{persona.geom.mi.cb.res}{+0.76}
\pvset{persona.geom.mi.rp.att}{-0.11}
\pvset{persona.geom.mi.cb.att}{-0.02}
\pvset{persona.geom.ll.rp.res}{+0.52}
\pvset{persona.geom.ll.cb.res}{+0.55}
\pvset{persona.geom.ll.rp.att}{+0.08}
\pvset{persona.geom.ll.cb.att}{+0.17}
\pvset{ga.gcg.ssm.base}{15.0\%}
\pvset{ga.gcg.ssm.deployed}{1.0\%}
\pvset{ga.gcg.ssm.fired}{100}
\pvset{ga.gcg.ssm.firedtot}{100}
\pvset{am.ssm.write}{6.0\%}
\pvset{am.ssm.write.ci}{[2.8\%, 12.5\%]}
\pvset{am.ssm.resid}{6.0\%}
\pvset{am.ssm.resid.ci}{[2.8\%, 12.5\%]}
\pvset{am.ssm.residdir}{100.0\%}
\pvset{am.tf.write}{15.0\%}
\pvset{am.tf.write.ci}{[9.3\%, 23.3\%]}
\pvset{am.tf.resid}{23.0\%}
\pvset{am.tf.resid.ci}{[15.8\%, 32.2\%]}
\pvset{am.tf.residdir}{48.0\%}
\pvset{am.ssm.normratio}{7.8}
\pvset{am.tf.normratio}{27.9}

}{}
\IfFileExists{ci_values.tex}{

\pvset{ci.def.ssm.prefix_injection.base}{[95.0, 99.2]}
\pvset{ci.def.ssm.prefix_injection.gated}{[2.7, 9.0]}
\pvset{ci.def.ssm.roleplay_dan.base}{[75.5, 86.3]}
\pvset{ci.def.ssm.roleplay_dan.gated}{[24.1, 36.7]}
\pvset{ci.def.ssm.fictional_framing.base}{[81.6, 91.0]}
\pvset{ci.def.ssm.fictional_framing.gated}{[12.0, 22.3]}
\pvset{ci.def.ssm.prefix_injection.base.pooled}{[96.7, 99.0]}
\pvset{ci.def.ssm.prefix_injection.gated.pooled}{[2.2, 5.1]}
\pvset{ci.def.n300.prefix_injection.base}{[93.6, 97.9]}
\pvset{ci.def.n300.prefix_injection.gated}{[2.3, 6.9]}
\pvset{ci.def.n300.roleplay_dan.base}{[79.1, 87.4]}
\pvset{ci.def.n300.roleplay_dan.gated}{[31.4, 42.3]}
\pvset{ci.def.n300.fictional_framing.base}{[86.8, 93.5]}
\pvset{ci.def.n300.fictional_framing.gated}{[12.0, 20.2]}
\pvset{ci.def.hb.base}{[89.2, 96.2]}
\pvset{ci.def.hb.gated}{[3.1, 9.6]}
\pvset{ci.def.opt.base}{[94.3, 98.9]}
\pvset{ci.def.opt.gated}{[3.8, 10.8]}
\pvset{ci.transfer.x2s.defend.transferred}{[16.1, 42.8]}
\pvset{ci.transfer.x2s.defend.random_mapped}{[87.1, 99.6]}
\pvset{ci.transfer.x2s.defend.native}{[10.5, 34.8]}
\pvset{ci.transfer.x2s.defend.base}{[87.1, 99.6]}
\pvset{ci.transfer.x2s.ablate.transferred}{[24.2, 53.0]}
\pvset{ci.transfer.x2s.ablate.random_mapped}{[2.6, 19.9]}
\pvset{ci.transfer.x2s.ablate.native}{[28.5, 57.8]}
\pvset{ci.transfer.x2s.ablate.base}{[4.0, 23.1]}
\pvset{ci.transfer.s2x.defend.transferred}{[28.5, 57.8]}
\pvset{ci.transfer.s2x.defend.random_mapped}{[65.2, 89.5]}
\pvset{ci.transfer.s2x.defend.native}{[73.9, 94.5]}
\pvset{ci.transfer.s2x.defend.base}{[18.1, 45.4]}
\pvset{ci.transfer.s2x.ablate.transferred}{[65.2, 89.5]}
\pvset{ci.transfer.s2x.ablate.random_mapped}{[1.4, 16.5]}
\pvset{ci.transfer.s2x.ablate.native}{[35.2, 64.8]}
\pvset{ci.transfer.s2x.ablate.base}{[0.4, 12.9]}
\pvset{ci.transfer.mi2ma.defend.transferred}{[18.1, 45.4]}
\pvset{ci.transfer.mi2ma.defend.random_mapped}{[57.2, 83.9]}
\pvset{ci.transfer.mi2ma.defend.native}{[10.5, 34.8]}
\pvset{ci.transfer.mi2ma.defend.base}{[87.1, 99.6]}
\pvset{ci.transfer.mi2ma.ablate.transferred}{[24.2, 53.0]}
\pvset{ci.transfer.mi2ma.ablate.random_mapped}{[4.0, 23.1]}
\pvset{ci.transfer.mi2ma.ablate.native}{[28.5, 57.8]}
\pvset{ci.transfer.mi2ma.ablate.base}{[4.0, 23.1]}
\pvset{ci.transfer.ma2mi.defend.transferred}{[87.1, 99.6]}
\pvset{ci.transfer.ma2mi.defend.random_mapped}{[68.0, 91.3]}
\pvset{ci.transfer.ma2mi.defend.native}{[87.1, 99.6]}
\pvset{ci.transfer.ma2mi.defend.base}{[83.5, 98.6]}
\pvset{ci.transfer.ma2mi.ablate.transferred}{[44.6, 73.7]}
\pvset{ci.transfer.ma2mi.ablate.random_mapped}{[49.5, 77.9]}
\pvset{ci.transfer.ma2mi.ablate.native}{[35.2, 64.8]}
\pvset{ci.transfer.ma2mi.ablate.base}{[16.1, 42.8]}
\pvset{ci.def.gcg.base}{[31.9, 50.8]}
\pvset{ci.def.gcg.gated}{[11.7, 26.7]}
\pvset{ci.def.gcg.gatedj4}{[2.2, 11.2]}
\pvset{ci.def.gcg.ga.attn}{[34.7, 53.8]}
\pvset{ci.def.gcg.ga.glob}{[15.8, 32.2]}
\pvset{ci.def.gcg.ga.attnj4}{[7.0, 19.8]}
\pvset{ci.def.gcg.ga.globj4}{[2.2, 11.2]}
\pvset{ci.def.mistralattn.prefix.base}{[93.0, 99.4]}
\pvset{ci.def.mistralattn.prefix.gated}{[8.5, 22.1]}
\pvset{ci.def.llamaattn.prefix.base}{[17.5, 34.3]}
\pvset{ci.def.llamaattn.prefix.gated}{[0.0, 3.7]}
\pvset{ci.orth.rwkv.base}{[1.0, 8.5]}
\pvset{ci.orth.rwkv.transp}{[48.2, 67.2]}
\pvset{ci.orth.rwkv.rand}{[30.9, 49.8]}
\pvset{ci.orth.rwkv.pooled.base}{[1.6, 5.6]}
\pvset{ci.orth.rwkv.pooled.transp}{[49.0, 60.2]}
\pvset{ci.orth.rwkv.pooled.rand}{[27.3, 37.8]}
\pvset{ci.orth.zamba.attn.base}{[5.8, 22.2]}
\pvset{ci.orth.zamba.attn.transp}{[11.8, 31.8]}
\pvset{ci.orth.zamba.attn.rand}{[1.7, 13.7]}
\pvset{ci.orth.zamba.attn.pooled.base}{[5.5, 14.0]}
\pvset{ci.orth.zamba.attn.pooled.transp}{[11.9, 22.8]}
\pvset{ci.orth.zamba.attn.pooled.rand}{[5.5, 14.0]}
\pvset{ci.orth.zamba.mamba.base}{[5.8, 22.2]}
\pvset{ci.orth.zamba.mamba.transp}{[81.9, 96.4]}
\pvset{ci.orth.zamba.mamba.rand}{[15.8, 37.2]}
\pvset{ci.orth.zamba.mamba.sweep.a1.transp}{[6.9, 24.2]}
\pvset{ci.orth.zamba.mamba.sweep.a1.rand}{[6.9, 24.2]}
\pvset{ci.orth.zamba.mamba.sweep.a2.transp}{[5.8, 22.2]}
\pvset{ci.orth.zamba.mamba.sweep.a2.rand}{[8.1, 26.1]}
\pvset{ci.orth.zamba.mamba.sweep.a4.transp}{[11.8, 31.8]}
\pvset{ci.orth.zamba.mamba.sweep.a4.rand}{[6.9, 24.2]}
\pvset{ci.orth.zamba.mamba.sweep.a8.transp}{[47.4, 71.4]}
\pvset{ci.orth.zamba.mamba.sweep.a8.rand}{[1.7, 13.7]}
\pvset{ci.orth.zamba.mamba.sweep.a12.transp}{[81.9, 96.4]}
\pvset{ci.orth.zamba.mamba.sweep.a12.rand}{[15.8, 37.2]}
\pvset{ci.orth.zamba.mamba.sweep.a16.transp}{[94.0, 100.0]}
\pvset{ci.orth.zamba.mamba.sweep.a16.rand}{[73.9, 91.9]}
\pvset{ci.orth.zamba.mamba.sweep.a24.transp}{[94.0, 100.0]}
\pvset{ci.orth.zamba.mamba.sweep.a24.rand}{[91.1, 99.7]}
\pvset{ci.orth.zamba.mamba.pooled.transp}{[88.7, 96.1]}
\pvset{ci.orth.zamba.mamba.pooled.rand}{[17.3, 29.4]}
\pvset{ci.def.zamba.base}{[21.3, 44.2]}
\pvset{ci.def.zamba.attn}{[0.0, 6.0]}
\pvset{ci.def.zamba.mamba.swbase}{[16.1, 42.8]}
\pvset{ci.def.zamba.mamba.best}{[1.4, 16.5]}
\pvset{ci.rwkv.site.base}{[32.9, 62.5]}
\pvset{ci.rwkv.site.timemix.gated}{[16.1, 42.8]}
}{}

\title{Locating and Steering Refusal Beyond Attention}

\author{
    Preethi Carmel Bosco,
    Gopalakrishnan Srinivasan
}
\affiliations{
    Department of Computer Science and Engineering, Indian Institute of Technology Madras\\
    preethicarmel@gmail.com, sgopal@cse.iitm.ac.in
}

\begin{document}
\maketitle

\begin{abstract}
Where inside a language model does refusal live, and does that place change when the architecture does? In a transformer, refusal is governed by a single direction in the residual stream, a finding that safety
and interpretability tooling now depend on. State-space
models (SSMs) route information through a recurrent update instead of attention. They share no token-mixing mechanism with a
transformer. Does the same safety representation survive this shift, or must it be rediscovered for each new
architecture? It survives. We find that a single rotation aligns one model's representation space with
another's. Under this rigid rotation, which can only reorient a space and not reshape it, the alignment
shows the two models genuinely share the representation. A harm probe trained on a transformer then flags an SSM's harmful inputs, and
removing the aligned direction makes a model answer attacks it would otherwise refuse, while a random
direction of the same size does far less.
What is architecture-specific is not where the direction is steered but where it must be read.
\ifarxivbuild Each layer computes a fresh output that is then added into the residual stream, and harm is
cleanly readable at this output, the write site, before the addition. A control that holds the
intervention's strength fixed shows that what matters is where the direction is estimated, not where it
is applied: a direction estimated at the write site controls refusals, while one estimated in the summed
stream does not.\else Each layer
writes its own output into the residual stream before the stream sums it in, and harm is cleanly readable
at this write site. A control that holds the intervention's strength fixed shows that a refusal direction
estimated at the write site controls refusals while one estimated in the summed stream does not. The
location of the intervention itself does not matter.\fi{} Applied through a detector-triggered gate (a
lightweight classifier flagging harmful prompts), this direction lowers jailbreak success in all four
architecture families we test (SSM, transformer, recurrent, and hybrid). On the SSM
it holds against an attacker that tunes its prompt against the defense: the classifier fires on all
\pv{ga.gcg.ssm.firedtot} attacked prompts and attack success stays at \pv{ga.gcg.ssm.deployed} (from
\pv{ga.gcg.ssm.base}). \ifarxivbuild The defense only matches, and does not beat, a trivial rule that returns a fixed refusal
whenever the same detector fires, so the gate adds no defense strength of its own. The contribution is
that the direction it amplifies transfers across architectures.\else The defense only ties a
trivial refuse-when-flagged rule, so what transfers across architectures is the direction itself.\fi{} \ifarxivbuild Where target labels are scarce, a transported direction used as a harm detector (on our
Llama-to-Falcon-Mamba pair) flags \pv{payoff.k64.tprfive.deltapts} percentage points more of the harmful
prompts than one trained from the same few labels, at a matched \pv{payoff.k64.tprfive.fpr} false-alarm
rate.\else Where target labels are scarce, a transported direction used as a harm detector detects
\pv{payoff.k64.tprfive.deltapts} percentage points (pt) more harmful prompts than one trained from the same
few labels, at a matched \pv{payoff.k64.tprfive.fpr} false-alarm rate.\fi{}
Two human raters validate the automated judge behind these numbers, agreeing with it \pv{hv.judge.lg.agree}
of the time ($\kappa{=}\pv{hv.judge.lg.kappa}$), and the judge's few mistakes only overstate attacks, so
our defense numbers are not inflated. Safety
tooling built on refusal therefore ports to a new architecture by re-estimating the direction at that
architecture's write site, not by rebuilding it.
\end{abstract}

\section{Introduction}

The safety of a deployed language model rests on its refusals: it is safe only if it refuses what it should
not answer, the one behavior every jailbreak targets. In a transformer, refusal is governed by a single
linear direction in the residual stream, and a growing safety stack reads and steers that direction to
detect and block harm \citep{arditi2024refusal,zou2023repe,lee2024cast}. Every one of these results is in a
transformer, where information between positions is routed by attention. State-space models (SSMs) such as Mamba
\citep{gu2023mamba,zuo2024falconmamba} route it through a recurrent state update instead, and are now entering
deployment. If refusal is a property of attention, this stack breaks whenever a model routes information
differently, and each new architecture needs its safety tooling rediscovered from scratch.
Is refusal a property of the transformer \emph{architecture}, or the same \emph{function} every model
learns, and where inside a model does it live?

\textbf{We find that it is the function.} The refusal representation is shared across architectures that
route context in completely different ways: by attention (Llama-3.1, Mistral), by a selective scan
(Falcon-Mamba), by token-shift recurrence (RWKV-6), or by a mix of attention and scan (Zamba2). A refusal
direction read in a transformer and mapped into a Mamba state space is still the state-space model's
refusal trigger, and removing it there jailbreaks the model (\fltabkey{tab:ablate}). We also find that what is architecture-specific is
not \emph{where} refusal is steered but \emph{where it must be read}: harm is cleanly decodable at the fresh
contribution each architecture writes before the residual accumulates it (the attention output in a
transformer, the selective-scan block output in an SSM; \ifbodyfloats Figures~\ref{fig:writesite} and \ref{fig:writesite-rwkv} sketch this per
architecture\else Appendix
Figures~\ref{fig:writesite}--\ref{fig:writesite-hybrid} sketch this per architecture\fi), and a matched-magnitude control shows the direction
estimated there controls refusal while one estimated in the accumulated residual degenerates. This yields
one rule: \emph{refusal is architecture-independent as a representation and architecture-specific as a
readout site}. We study this in
Falcon-Mamba-7B, with Llama-3.1-8B \citep{grattafiori2024llama3} and Mistral-7B \citep{jiang2023mistral} as
transformer references, and confirm it on RWKV-6 \citep{peng2024rwkv} and
the hybrid Zamba2-7B \citep{glorioso2024zamba2}. We make three contributions.
\begin{itemize}
  \item \textbf{Refusal is a shared representation across architectures.} A single rigid (Procrustes)
        rotation \citep{schonemann1966procrustes} both aligns the representation spaces and causally
        transports the refusal direction. Removing the transported direction jailbreaks the target:
        decisively across the transformer-SSM pair and at the hybrid Zamba2-7B's Mamba write site, more
        weakly at its attention site and on the single-checkpoint RWKV-6. A refusal-orthogonal random direction has a far smaller effect. This is
        the paper's central result.
  \item \textbf{The refusal direction must be read at the fresh write, not the accumulated residual.}
        Harm peaks in decodability at each architecture's write site: a linear probe separates harmful
        from benign prompts with AUROC \pv{locus.output.low} (1.0 is perfect separation, 0.5 chance) on
        the surface-controlled subset, where a surface-only probe scores \pv{locus.surface}. A
        matched-magnitude control (Table~\ref{tab:absmatch}) shows the direction \emph{estimated} there
        controls refusal while one estimated in the residual degenerates, and that the intervention's
        \emph{location} is irrelevant.
  \item \textbf{Applied through a detector-triggered gate, that direction causally controls refusal in all
        four architecture classes.} The gate is a causal test of the readout claim, not a defense
        contribution: its numbers only match a simpler baseline that returns a fixed refusal whenever the
        same detector fires, so what ports across architectures is the direction, not defense strength
        (Section~\ref{sec:results}). The gate lowers prefix-injection attack success on
        Falcon-Mamba (\pv{def.ssm.prefix_injection.base} to \pv{def.ssm.prefix_injection.gated}) and on
        three further classes (the two transformers, the recurrent RWKV-6, and the hybrid,
        Table~\ref{tab:routingsite}). On Falcon-Mamba
        it also holds under a suffix optimized through the gate: the detector fires on all
        \pv{ga.gcg.ssm.firedtot} attacked prompts and attack success stays at \pv{ga.gcg.ssm.deployed}
        (from \pv{ga.gcg.ssm.base}), where the same adaptive attack erodes a transformer's detector.
\end{itemize}

\section{Method}
\label{sec:method}

\subsection{Threat model}
An adversary controls the input prompt and may adapt to the defense, including optimizing the
prompt against the frozen detector \citep{zou2023gcg,andriushchenko2024adaptive}. The adversary
cannot change the model weights or the detector. The defender runs the detector and the gate at
inference time. We separate two evaluations (Section~\ref{sec:experiments}):
transfer to unseen attack families, and robustness to an attacker that adapts to the detector.

\subsection{Detector and the detect-then-gate loop}
The detector is a linear probe read at the routing layer, the write-site layer whose activations carry
the per-prompt decision (selected in Section~\ref{sec:recipe}), one dot product per prompt, trained on pairs of
harmful and benign prompts. Both the harmful and the benign prompts also appear wrapped in jailbreak
templates during training, so the probe cannot learn to fire on the wrapper text itself and must read the
underlying request (Appendix~\ref{app:hardening}). It scores the prompt once: if $p_{\text{harm}} > \theta$
the gate is applied at the architecture's write site for the whole generation, otherwise the model
runs unmodified.
 The decision is made per prompt, not per token. Tokens generated after a flagged prompt also score as
harmful, so a per-token gate would keep re-triggering and drift all generations toward refusal. The write-site result holds under a nonlinear (multi-layer perceptron, MLP)
detector as well as a linear one, and every cell of the baseline comparison shares the same
fixed detector and threshold (Appendix~\ref{app:detablation}).

\subsection{Write site: the SSM selective-scan gate (anchor)}
\ifbodyfloats
\begin{figure*}[t]\centering
\includegraphics[height=3.4in]{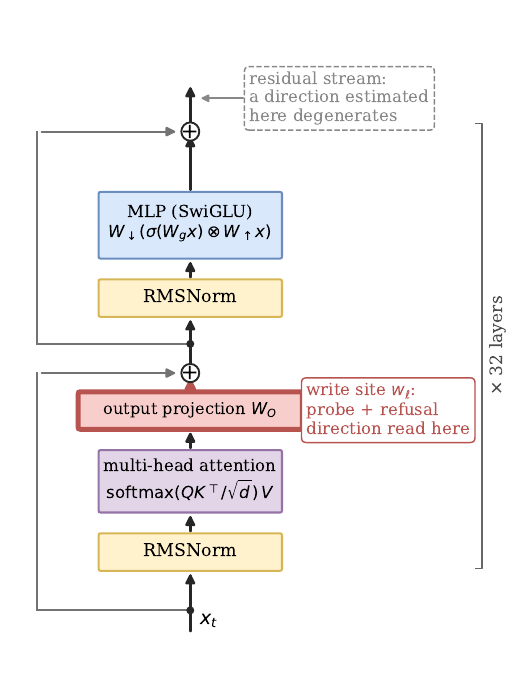}\hspace{0.7in}
\includegraphics[height=3.4in]{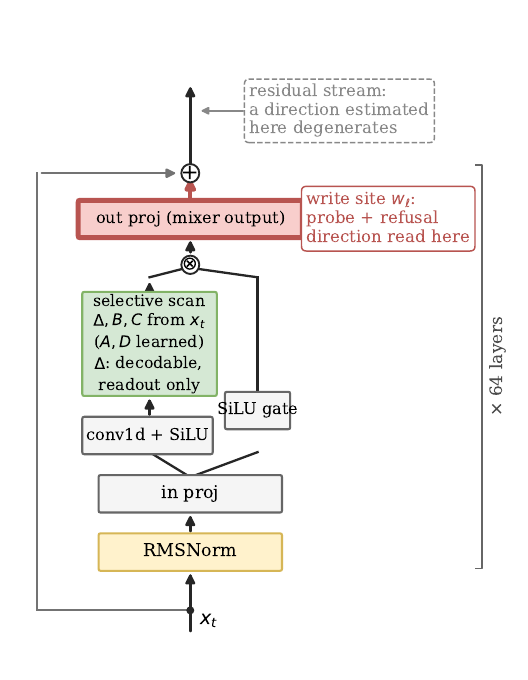}
\caption{The write site in the two reference architectures. Left: transformer block (Llama-3.1,
Mistral). The write site is the attention output projection $W_O$, the fresh contribution attention
writes into the residual stream. Right: Falcon-Mamba block (selective-scan SSM). The write site is the
mixer's out-projection, the block output the $\Delta$-gated scan writes before the residual accumulates
it, and $\Delta$ inside the scan is decodable but a readout only. In both, the probe and the refusal
direction are estimated at the red site; a direction estimated in the accumulated residual (gray)
degenerates.}
\label{fig:writesite}\label{fig:writesite-ssm}
\end{figure*}
\fi
Falcon-Mamba-7B, our most fully instrumented model, is the \emph{anchor} we dissect first.
In a selective state-space model, information propagation is controlled by three input-dependent
selection parameters: a step size ($\Delta$), an input-selection gate ($B$), and a readout ($C$). Two
additional matrices are fixed and learned once during training: a state-transition matrix ($A$) and a
skip-connection matrix ($D$). At
each layer $\ell$ the mixer maps the input $x_t$ to the selection
$(\Delta_t, B_t, C_t)$ and updates the hidden state via the scan $h_t = \exp(\Delta_t A)\, h_{t-1} + (\Delta_t B_t)\, x_t$,
with readout $y_t = C_t\, h_t + D\, x_t$. Our decodability study (probing each layer with a linear
classifier and measuring how well it separates harmful from benign activations) identifies the step size
$\Delta$, the input-dependent
selection with no counterpart in a transformer, as a location of harm. The probe detects harm at $\Delta$ nearly as well as at the mixer output
(Section~\ref{sec:results}), but $\Delta$ is where harm is read, not where the gate steers: both the
detector and the gate act at the block output. This is the per-layer
contribution $y_t$ that the $\Delta$-gated scan writes before it
is added to the residual stream, a distinct site from the residual stream itself. We call this site the
block output (equivalently the mixer output, the mixer's out-projection in Figure~\ref{fig:writesite-ssm}). On a
positive prompt-level decision the gate amplifies the refusal direction $\hat{u}_\ell$ at this
selective-scan site for the whole generation:
\begin{equation}
g(y_t) = y_t + \alpha \, \hat{u}_\ell ,
\label{eq:gate}
\end{equation}
which extends the additive steering of \citet{arditi2024refusal} to the SSM's selective-scan block output
(formal definitions in Appendix~\ref{app:formal}). The direction $\hat{u}_\ell$
is the mean difference between harmful and harmless activations at layer $\ell$, not the detector's
discriminative direction, which reads harm well but does not control refusal when amplified
(Appendix~\ref{app:geometry}). The strength $\alpha$ and the routing
layers are set by a small held-out sweep, with $\alpha$ kept below the value at which generation
degenerates (the coherent gain, Appendix~\ref{app:coherence}).
\emph{To our knowledge, no prior steering or safety method acts at the selective-scan block of an SSM}
(Section~\ref{sec:related}).

\ifbodyfloats\begin{table*}[t]\centering\scriptsize\setlength{\tabcolsep}{4pt}
\caption{Experimental setup at a glance. The minimal-pair corpus is matched to XSTest in template and
length. Over-refusal is measured on XSTest's benign-but-spicy prompts (safe requests worded to sound
harmful). Attack success is the
Llama-Guard-3-8B-judged unsafe fraction ($n{=}\pv{def.ssm.n}$ harmful, \pv{def.ssm.nspicy}
benign-but-spicy per condition, 95\% CI, four-judge agreement \pv{def.ssm.judgeagreenway}). Capability
is accuracy with and without the gate.}
\label{tab:expsetup}
\begin{tabular}{>{\raggedright\arraybackslash}p{0.19\textwidth}|>{\raggedright\arraybackslash}p{0.14\textwidth}|>{\raggedright\arraybackslash}p{0.14\textwidth}|>{\raggedright\arraybackslash}p{0.19\textwidth}|>{\raggedright\arraybackslash}p{0.21\textwidth}}
\toprule
Models & Corpora & Attacks & Judges \& metrics & Baselines \\
\midrule
Falcon-Mamba-7B (SSM anchor)\newline
Falcon3-Mamba-7B (2nd SSM)\newline
Llama-3.1-8B, Mistral-7B, Qwen2.5-7B (transformers)\newline
RWKV-6-World-7B (recurrent)\newline
Zamba2-7B (hybrid)
&
AdvBench (harm)\newline
Alpaca (benign)\newline
XSTest (surface control, over-refusal)
&
Prefix injection (fluent)\newline
Persona roleplay (DAN, fluent)\newline
Fictional framing (fluent)\newline
Dilution (adaptive)\newline
Optimized adversarial suffix (GCG; on the SSM both a gradient-free variant and GCG through the differentiable scan)
&
Llama-Guard-3-8B (primary)\newline
+Llama-3.1-8B-Instruct, Mistral-7B-Instruct, ShieldGemma-9B\newline
MMLU, TruthfulQA (capability)\newline
Probe AUROC (decodability)
&
Residual-subspace projection\newline
CAST\newline
Perplexity filter\newline
SmoothLLM\newline
Cross-architecture linear-map transfer
\\
\bottomrule
\end{tabular}
\end{table*}
\else\ifkeyfloats\begin{table*}[t]\centering\scriptsize\setlength{\tabcolsep}{4pt}
\caption{Experimental setup at a glance. The minimal-pair corpus is matched to XSTest in template and
length. Over-refusal is measured on XSTest's benign-but-spicy prompts (safe requests worded to sound
harmful). Attack success is the
Llama-Guard-3-8B-judged unsafe fraction ($n{=}\pv{def.ssm.n}$ harmful, \pv{def.ssm.nspicy}
benign-but-spicy per condition, 95\% CI, four-judge agreement \pv{def.ssm.judgeagreenway}). Capability
is accuracy with and without the gate.}
\label{tab:expsetup}
\begin{tabular}{>{\raggedright\arraybackslash}p{0.19\textwidth}|>{\raggedright\arraybackslash}p{0.14\textwidth}|>{\raggedright\arraybackslash}p{0.14\textwidth}|>{\raggedright\arraybackslash}p{0.19\textwidth}|>{\raggedright\arraybackslash}p{0.21\textwidth}}
\toprule
Models & Corpora & Attacks & Judges \& metrics & Baselines \\
\midrule
Falcon-Mamba-7B (SSM anchor)\newline
Falcon3-Mamba-7B (2nd SSM)\newline
Llama-3.1-8B, Mistral-7B, Qwen2.5-7B (transformers)\newline
RWKV-6-World-7B (recurrent)\newline
Zamba2-7B (hybrid)
&
AdvBench (harm)\newline
Alpaca (benign)\newline
XSTest (surface control, over-refusal)
&
Prefix injection (fluent)\newline
Persona roleplay (DAN, fluent)\newline
Fictional framing (fluent)\newline
Dilution (adaptive)\newline
Optimized adversarial suffix (GCG; on the SSM both a gradient-free variant and GCG through the differentiable scan)
&
Llama-Guard-3-8B (primary)\newline
+Llama-3.1-8B-Instruct, Mistral-7B-Instruct, ShieldGemma-9B\newline
MMLU, TruthfulQA (capability)\newline
Probe AUROC (decodability)
&
Residual-subspace projection\newline
CAST\newline
Perplexity filter\newline
SmoothLLM\newline
Cross-architecture linear-map transfer
\\
\bottomrule
\end{tabular}
\end{table*}
\fi\fi

\subsection{Write site: the transformer attention output (reference)}
For the transformer reference we estimate the refusal direction at the attention output projection (the
fresh write) and, as a baseline, in the residual stream, and apply the same gate $g(\cdot)$ from
Equation~\ref{eq:gate} at the layer where the direction induces refusal, triggered by the same prompt-level
detector. The direction estimated at the attention output defends, while the one estimated in the accumulated
residual degenerates, for both transformers we test, Llama-3.1-8B and Mistral-7B
(Table~\ref{tab:routingsite}; full sweep in Appendix~\ref{app:attnsweep}). A matched-absolute-magnitude
control isolates the cause. With the same direction held fixed, pushing it at the residual and pushing it
at the attention output lower attack success by statistically indistinguishable amounts (overlapping 95\%
confidence intervals). What matters is therefore where the direction is \emph{estimated}, not where the
gate is \emph{applied} (Appendix~\ref{app:absmatch}). The position-wise MLP output, which does not
mix information across positions, yields an equally usable direction, ruling out cross-position mixing as
the source (Appendix~\ref{app:mlpcontrol}, \ref{app:ssmfreshness}).

\subsection{A recipe for siting the gate on a new model}
\label{sec:recipe}
These findings give a procedure for deploying the gate on a new architecture, given only a small
contrastive set: find the write site where the refusal direction is cleanly decodable (the selective-scan
block output in an SSM, the attention output in a transformer), select the gate layers by held-out
decodability, estimate the mean refusal direction and a linear trigger there, calibrate the gain below the
degeneration cliff, and deploy closed-loop. The only
model-specific work is layer selection and gain calibration, where a naive transfer of one model's
constants to another fails (Appendix~\ref{app:recipe}). Which component exposes a decodable direction
changes with the architecture (a pure SSM has no MLP, and in Falcon-Mamba the newly written,
non-accumulated part of the block output is the skip term $D x_t$, Appendix~\ref{app:ssmfreshness}), so the
recipe's first step, finding the write site, is done per architecture rather than assumed.

\subsection{Baselines}
The baselines vary the intervention against our gate: residual-subspace projection (LEACE-style rank-$k$
erasure \citep{belrose2023leace}) is an open-loop residual edit, conditional activation steering (CAST)
\citep{lee2024cast} shares our detector but reads and steers at the residual, and an always-on version of
our push (as DeTAM does \citep{li2025detam}) drops the loop (results in
\ifkeyfloats Table~\ref{tab:baselinesmain}\else Table~\ref{tab:baselines}\fi). All cells share the critical
layers and detector threshold $\theta$ from our decodability study, with the erasure rank $k$ and gate
strength $\alpha$ swept and calibrated to stay coherent (Appendix~\ref{app:coherence}).

\subsection{Why projection fails and gating should not}
Projection is brittle because a fixed subspace can be routed around downstream. A detection-triggered
gate instead acts on the routing decision
for the whole generation, leaving no fixed subspace to route around. We measure the in-distribution,
out-of-distribution, and adaptive cases separately.

\section{Experimental Setup}
\label{sec:experiments}

\fltabkey{tab:expsetup} summarizes models (including Qwen2.5-7B \citep{qwen2025qwen25}), corpora, attacks,
judges, metrics, and baselines at a glance.

The setup serves two questions, whether refusal is the same representation across the
architectures and where that representation is read and steered. The detector is trained on harm versus
benign-but-spicy prompts, a different split from the refusal direction's harm-versus-Alpaca \citep{taori2023alpaca} fit.
Decodability is reported as AUROC, the probability that a randomly chosen harmful prompt scores above a
randomly chosen benign one (1.0 is perfect separation, 0.5 is chance), and uncertainty on attack-success
and over-refusal rates as Wilson 95\% confidence intervals, the standard interval for binomial proportions.
\ifarxivbuild Code, data, and result files will be released publicly upon publication.\else Code, data, and result files are in the supplementary material uploaded with this submission.\fi
\ifbodyfloats\else\FloatBarrier\fi

\section{Locating refusal across architectures}
\label{sec:transfer}
In a transformer, refusal is a direction in the residual stream. Does a pure SSM, which shares no
token-mixing mechanism with attention, carry the \emph{same} refusal representation up to a change of basis, or learn
its own? We locate where the refusal signal lives in the SSM and find it is the same representation: a single
rigid rotation carries it from the transformer's activation space into the SSM's, where it still causally
controls refusal.

\ifbodyfloats\begin{figure}[t]\centering
\includegraphics[width=3.3in]{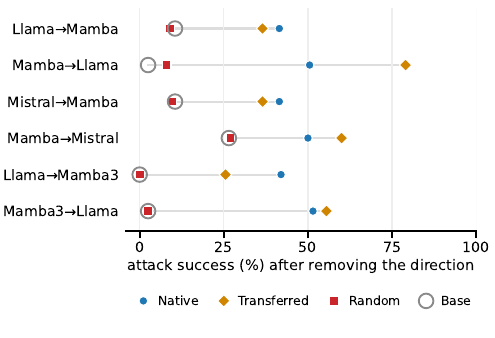}
\caption{The refusal direction is shared across architectures. Reading it in one architecture, mapping it
into another, and removing it there (Transferred) raises attack success as much as removing the target's own
direction (Native), while an orthogonalized random direction through the same map (Random) does not, in all
six transfer directions. A plot of \fltab{tab:transfer} (Mamba${=}$Falcon-Mamba,
Mamba3${=}$Falcon3-Mamba).}
\label{fig:transfer}
\end{figure}
\fi

\subsection{Harm is decodable at the selective-scan step size}
Harm is linearly decodable from the Falcon-Mamba selective-scan state, concentrated in the step size
$\Delta$ and the mixer output. The harm probe is tested on a version of XSTest \citep{rottger2024xstest}
where harmful and benign prompts are matched for surface features like sentence structure and wording
(the matched pairs are generated from a hand-written template bank crossed with length-matched
harmful/benign action pairs, shipped as code in the release). A
surface-only probe scores near chance there (AUROC \pv{locus.surface}), while the harm probe still
separates them, so it is reading content rather than surface cues (per-component AUROC in
\fltab{tab:locus}). On that subset $\Delta$ is the most decodable \emph{selection} component
(\pv{locus.delta.low}, against \pv{locus.b.low} and \pv{locus.c.low} for the input and readout selections
$B$ and $C$), and the mixer output the scan writes is more decodable still
(\pv{locus.output.low}). $\Delta$ is the input-dependent gate that sets how much each token writes into the
recurrent state, so harm is exposed in the SSM's \emph{selection} and then written at the mixer output,
where decodability peaks. The defense in Section~\ref{sec:results} reads and steers there. This also
settles $\Delta$'s causal status: it is decodable, but a readout rather than the controlling site.

\ifbodyfloats\begin{table}[t]\centering\scriptsize
\caption{Where harm lives in the selective scan. Surface-controlled held-out AUROC of a linear harm probe
at each selective-scan component (Falcon-Mamba-7B), on the surface-matched XSTest contrast where a surface
probe is near chance (\pv{locus.surface}). Harm is decodable at the step size $\Delta$ and the mixer output,
not at $B$ or $C$.}
\label{tab:locus}
\begin{tabular}{lc}
\toprule
Selective-scan component & AUROC (surface-controlled) \\
\midrule
$\Delta$ (step size) & \pv{locus.delta.low} \\
$B$ (input selection) & \pv{locus.b.low} \\
$C$ (readout selection) & \pv{locus.c.low} \\
mixer output & \pv{locus.output.low} \\
\bottomrule
\end{tabular}
\end{table}
\fi

The locus is stable across model scale, with harm decodable at $\Delta$ from AUROC \pv{scale.delta.small}
to \pv{scale.delta.large} across \pv{scale.nmodels} base Mamba models from \pv{scale.small} to
\pv{scale.large} parameters (Appendix~\ref{app:scale}), and the signal is present in the transformer
reference, where harm is decodable from the Llama-3.1-8B residual stream at AUROC \pv{e0.llama.auroc}. The
finding also holds on a newer SSM generation: Mamba-2 (Codestral-Mamba-7B) shows the same decodability at
its mixer output (AUROC \pv{mamba2.auroc}), so it is not specific to Mamba-1 (Appendix~\ref{app:mamba2}).

\subsection{The refusal representation is shared, up to a rigid rotation}
Transferring a steering direction within one architecture through a learned
map is established \citep{wu2025activationtransfer}. But a flexible map can synthesize the target's own
refusal axis, so it cannot by itself show the representations are shared. We test two criteria while
transferring from attention to the selective scan: a structural one, whether a rigid rotation aligns the
two representations, and a causal one, whether a transferred direction still carries its refusal role.

\paragraph{Setup.} On a shared set of harmful and harmless prompts we collect paired activations
from the source and target models and fit a ridge-regularized linear map $W$ between their hidden
spaces. We map the source refusal direction (mean harmful minus harmless) through $W$ and steer the
target with it, testing every architecture pair in both directions. Two behavioral tests follow,
amplifying the mapped direction on jailbroken prompts (defense) and removing it from harmful prompts
(ablation, where higher attack success means the direction was the refusal trigger). As a control, we
repeat both with a random source direction mapped through $W$ and orthogonalized to the target's refusal
axis, so the map cannot leak refusal signal into it (Appendix~\ref{app:cosine}).

\paragraph{Shared up to rotation, not merely mappable.} Because a flexible map could synthesize the
target's own refusal axis, we test with an \emph{orthogonal} (Procrustes) map, which can only rotate and
has no freedom to synthesize a direction. The
transported probe reaches held-out AUROC \pv{shared.proc} on the unrestricted harmful/benign corpus
(AdvBench versus Alpaca, where surface cues help) and \pv{shared.procm} on the
surface-matched minimal pairs, matching the target's own probe (\pv{shared.tgtown}), both near ceiling.
A harder test confirms this isn't just a ceiling effect. When the hardest XSTest category is held out
entirely, both the target's own probe and the transported probe drop together
(\pv{shared.catcv.mamba.bcown} vs.\ \pv{shared.catcv.mamba.bctransp}), tracking each other closely
(\fltabkey{tab:catcv}). The transported probe beats a shuffled-category control in all
\pv{shared.catcv.beatnull} of \pv{shared.catcv.beatnulltot} folds tested (Appendix~\ref{app:catcv},
\flfig{fig:catcv}). A flexible map that
had merely found the target's own axis would not track this. Fitting the rotation on deliberately
mismatched prompt pairs, source and target activations from different prompts rather than the same one,
collapses the transported probe to AUROC \pv{shared.mnull.mamba} (Appendix~\ref{app:matchednull}), at
and below chance. Only the correct pairing transports, so the representation is genuinely shared.

\ifbodyfloats
\begin{figure*}[t]\centering
\includegraphics[height=3.4in]{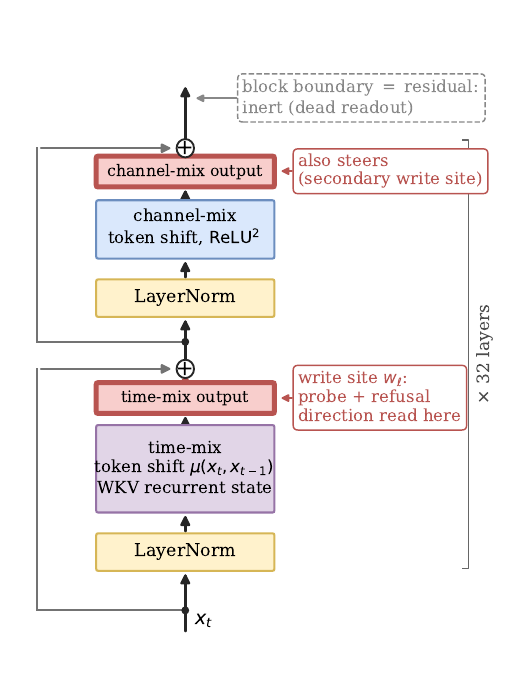}\hspace{0.7in}
\includegraphics[height=3.4in]{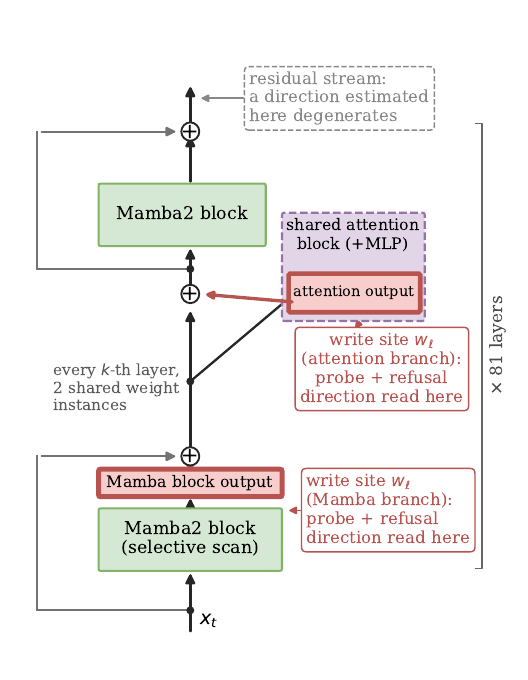}
\caption{The write site in the two further architecture classes. Left: RWKV-6 block (token-shift RNN).
The write sites are the sub-block outputs, the time-mix output (primary) and the channel-mix output
(secondary), while the block boundary is architecturally the residual and is inert. Right: Zamba2
(hybrid). A Mamba2 backbone taps shared attention blocks (two weight instances reused across depth)
every $k$-th layer, and both branch outputs are write sites.}
\label{fig:writesite-rwkv}\label{fig:writesite-hybrid}
\end{figure*}
\fi
This extends to a third, architecturally distinct model. RWKV-6 \citep{peng2024rwkv} uses token-shift
recurrence, not a selective scan, yet a Llama-read harm probe transports into it at AUROC
\pv{shared.rwkv.proc}, above the shuffled control \pv{shared.rwkv.shuf}. The transport is also causal,
though more weakly, with a margin narrower than the pure SSM's or the hybrid's Mamba site's
(Table~\ref{tab:ablate}; single-seed numbers, Wilson CIs, and judging methodology in
Appendix~\ref{app:rwkvdefense}). The site relocates. When each site is pushed at the same fraction of its
own activation norm, the residual stays inert (\pv{rwkv.nm.base} unchanged) while the time-mix write point
steers attack success to \pv{rwkv.nm.timemix.gated} against \pv{rwkv.nm.timemix.rand} for a matched random
direction, isolating direction from scale (the deployed defense is the Table~\ref{tab:routingsite} row).

\paragraph{Transport beats retraining when target labels are scarce.} Portability has practical value only if a
direction carried over from the source model outperforms one fitted directly in the target. We give both the same budget of $k$ labeled
\emph{target} prompts, spent either on the paired activations that fit the map or on the target's own
mean-difference direction, and score AUROC on a held-out harm category neither has seen. Budgets are matched in target labels only: the
transport arm also carries a direction estimated on the source model's abundant labels, which is the
asymmetry the practical case turns on. Transport wins at
every budget from $k{=}\pv{payoff.kmin}$ to \pv{payoff.kmax}, by \pv{payoff.deltamin} to
\pv{payoff.deltamax} AUROC, with the clustered interval excluding zero at every budget
(Appendix~\ref{app:payoff}). A deployed gate runs at a threshold, where the margin is
wider. Allowing \pv{payoff.k64.tprfive.fpr}
false alarms on benign prompts, the transported direction catches \pv{payoff.k64.tprfive.transp} of
harmful prompts against \pv{payoff.k64.tprfive.scratch} for one fit from the same labels, a gain of
\pv{payoff.k64.tprfive.deltapts} percentage points. We do not convert this into a
label-equivalence factor, because the from-scratch curve flattens and is not monotone in $k$
(Appendix~\ref{app:payoff}). The gain disappears in distribution, where both
arms saturate. What portability buys is therefore generalization: the transported direction catches kinds
of harm that the target's own few labels do not cover.

\ifbodyfloats\begin{table}[t]\centering\scriptsize\setlength{\tabcolsep}{3pt}
\caption{Per-pair category-held-out cross-validation results. Mean own and mean transported probe scores
are averaged across the \pv{shared.catcv.mamba.nfolds} held-out categories. Pearson $r$ measures how
closely they track each other. Full setup and interpretation in Section~\ref{sec:transfer}, plotted in
\flfig{fig:catcv}.}
\label{tab:catcv}
\resizebox{\columnwidth}{!}{%
\begin{tabular}{lcccc}
\toprule
Source $\to$ target & Mean own & Mean transp. & Pearson $r$ [95\% CI] & Folds $>$ null (shuffled-category control) \\
\midrule
Llama-3.1-8B $\to$ Falcon-Mamba & \pv{shared.catcv.mamba.own} & \pv{shared.catcv.mamba.transp} & \pv{shared.catcv.mamba.r} [\pv{shared.catcv.mamba.rlo}, \pv{shared.catcv.mamba.rhi}] & \pv{shared.catcv.mamba.beatp95}/\pv{shared.catcv.mamba.nfolds} \\
Mistral-7B $\to$ Falcon-Mamba & \pv{shared.catcv.mistralmamba.own} & \pv{shared.catcv.mistralmamba.transp} & \pv{shared.catcv.mistralmamba.r} [\pv{shared.catcv.mistralmamba.rlo}, \pv{shared.catcv.mistralmamba.rhi}] & \pv{shared.catcv.mistralmamba.beatp95}/\pv{shared.catcv.mistralmamba.nfolds} \\
\bottomrule
\end{tabular}}
\end{table}
\else\ifkeyfloats\begin{table}[t]\centering\scriptsize\setlength{\tabcolsep}{3pt}
\caption{Per-pair category-held-out cross-validation results. Mean own and mean transported probe scores
are averaged across the \pv{shared.catcv.mamba.nfolds} held-out categories. Pearson $r$ measures how
closely they track each other. Full setup and interpretation in Section~\ref{sec:transfer}, plotted in
\flfig{fig:catcv}.}
\label{tab:catcv}
\resizebox{\columnwidth}{!}{%
\begin{tabular}{lcccc}
\toprule
Source $\to$ target & Mean own & Mean transp. & Pearson $r$ [95\% CI] & Folds $>$ null (shuffled-category control) \\
\midrule
Llama-3.1-8B $\to$ Falcon-Mamba & \pv{shared.catcv.mamba.own} & \pv{shared.catcv.mamba.transp} & \pv{shared.catcv.mamba.r} [\pv{shared.catcv.mamba.rlo}, \pv{shared.catcv.mamba.rhi}] & \pv{shared.catcv.mamba.beatp95}/\pv{shared.catcv.mamba.nfolds} \\
Mistral-7B $\to$ Falcon-Mamba & \pv{shared.catcv.mistralmamba.own} & \pv{shared.catcv.mistralmamba.transp} & \pv{shared.catcv.mistralmamba.r} [\pv{shared.catcv.mistralmamba.rlo}, \pv{shared.catcv.mistralmamba.rhi}] & \pv{shared.catcv.mistralmamba.beatp95}/\pv{shared.catcv.mistralmamba.nfolds} \\
\bottomrule
\end{tabular}}
\end{table}
\fi\fi
\ifbodyfloats
\begin{table}[tb]\centering\scriptsize\setlength{\tabcolsep}{4pt}
\caption{Removing the rotation-transported refusal direction raises attack success in every non-transformer
architecture we test, while a matched random direction through the same map does much less
(Llama-Guard-judged; RWKV-6 and Zamba2 pooled over three seeds). The margin is decisive on the two Mamba
write sites, narrower on RWKV-6, whose random control alone jailbreaks a third of prompts, and not decisive
at Zamba2's attention site. Per-seed numbers, 95\% confidence intervals (CIs), and coherence screens in
Appendix~\ref{app:orthabl}, \ref{app:rwkvdefense}, and \ref{app:hybridablate}.}
\label{tab:ablate}
\begin{tabular}{@{}llccc@{}}
\toprule
Target (site) & Source & Base & Removed & Random \\
\midrule
Falcon-Mamba (block out) & Llama & \pv{transfer.x2s.ablate.base} & \pv{transfer.x2s.ablate.orthstdabl} & \pv{transfer.x2s.ablate.random_orth} \\
Falcon-Mamba (block out) & Mistral & \pv{transfer.mi2ma.ablate.base} & \pv{transfer.mi2ma.ablate.orthstdabl} & \pv{transfer.mi2ma.ablate.random_orth} \\
RWKV-6 (time-mix out) & Llama & \pv{orth.rwkv.pooled.base} & \pv{orth.rwkv.pooled.transp} & \pv{orth.rwkv.pooled.rand} \\
Zamba2 (Mamba out) & Llama & \pv{orth.zamba.mamba.pooled.base} & \pv{orth.zamba.mamba.pooled.transp} & \pv{orth.zamba.mamba.pooled.rand} \\
Zamba2 (attn.\ out) & Llama & \pv{orth.zamba.attn.pooled.base} & \pv{orth.zamba.attn.pooled.transp} & \pv{orth.zamba.attn.pooled.rand} \\
\bottomrule
\end{tabular}
\end{table}
\else\ifkeyfloats\fi\fi

\ifbodyfloats\begin{figure}[t]\centering
\includegraphics[width=0.52\linewidth]{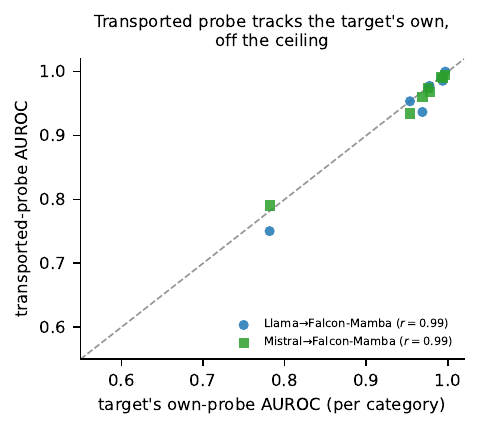}
\caption{Category-held-out cross-validation. Each point is one held-out XSTest category, plotting the
transported orthogonal probe's AUROC against the target's own-probe AUROC. Points follow the $y{=}x$ line as
the own probe falls off ceiling, so the transport tracks the target's harm structure category by category
rather than sitting at a saturated ceiling.}
\label{fig:catcv}
\end{figure}
\fi

\paragraph{The shared direction carries its causal role across all six transfer directions.}
Removing the mapped
refusal direction from harmful prompts raises the target's attack success far above an orthogonalized
random control (\fltab{tab:transfer}), across three
model pairs and both directions. The rigid rotation, not just the flexible map, produces the same causal
effect: the orthogonal map that structurally aligned the two models also transports the refusal direction
into Falcon-Mamba. Removing that transported direction jailbreaks the SSM as much as removing
its own native direction (\pv{transfer.x2s.ablate.orthstdabl} against \pv{transfer.x2s.ablate.native}),
and the output stays coherent. The same rigid-rotation ablation runs in every non-transformer target we have
(Table~\ref{tab:ablate}). One rigid rotation therefore both proves the representations are shared and
controls behavior once transported (Appendix~\ref{app:orthabl}).

\ifbodyfloats\begin{table}[t]\centering\scriptsize\setlength{\tabcolsep}{3pt}
\caption{Cross-architecture transfer, ablation test (plotted in Figure~\ref{fig:transfer}). Removing the
mapped direction (Transferred) raises attack success like removing the target's own (Native). The Random
control does not. Random is mapped through $W$ like Transferred, then orthogonalized to the target's
refusal axis (cosine ${\approx}\,0$), so it shares the map without inheriting the refusal direction. The
effect holds in all six rows: three model pairs (Mamba${=}$Falcon-Mamba, Mamba3${=}$Falcon3-Mamba), each
tested in both directions.}
\label{tab:transfer}
\begin{tabular}{lcccc}
\toprule
Pair / direction & Base & Native & Transferred & Random \\
\midrule
Llama$\to$Mamba & \pv{transfer.x2s.ablate.base} & \pv{transfer.x2s.ablate.native} & \pv{transfer.x2s.ablate.transferred} & \pv{transfer.x2s.ablate.random_orth} \\
Mamba$\to$Llama & \pv{transfer.s2x.ablate.base} & \pv{transfer.s2x.ablate.native} & \pv{transfer.s2x.ablate.transferred} & \pv{transfer.s2x.ablate.random_orth} \\
Mistral$\to$Mamba & \pv{transfer.mi2ma.ablate.base} & \pv{transfer.mi2ma.ablate.native} & \pv{transfer.mi2ma.ablate.transferred} & \pv{transfer.mi2ma.ablate.random_orth} \\
Mamba$\to$Mistral & \pv{transfer.ma2mi.ablate.base} & \pv{transfer.ma2mi.ablate.native} & \pv{transfer.ma2mi.ablate.transferred} & \pv{transfer.ma2mi.ablate.random_orth} \\
Llama$\to$Mamba3 & \pv{transfer.l2f3.ablate.base} & \pv{transfer.l2f3.ablate.native} & \pv{transfer.l2f3.ablate.transferred} & \pv{transfer.l2f3.ablate.random_orth} \\
Mamba3$\to$Llama & \pv{transfer.f32l.ablate.base} & \pv{transfer.f32l.ablate.native} & \pv{transfer.f32l.ablate.transferred} & \pv{transfer.f32l.ablate.random_orth} \\
\bottomrule
\end{tabular}
\end{table}
\fi

\paragraph{The site where the representation transports is also where the direction is cleanly read.}
A source model's refusal direction transports into the write site as the same representation, and the
direction read there controls behavior when amplified, while $\Delta$ is decodable but does not control
refusal (Section~\ref{sec:results}). Amplifying a model's refusal
direction read at its write site lowers prefix-injection attack success across four architecture classes
(Table~\ref{tab:routingsite}), while a direction read from the accumulated residual degenerates. A
matched-magnitude control (Appendix~\ref{app:absmatch}) shows the gap is the estimated direction, not the
push location, matching prior reports of residual-steering brittleness
\citep{li2026steeringpitfalls,korznikov2025rogue,xiong2026steering}, with the effect
size tracking each model's baseline alignment. The defense also holds against a harder, gradient-optimized
attack. A per-prompt greedy coordinate gradient (GCG) suffix \citep{zou2023gcg}, an adversarial string optimized
by gradient search to force compliance, reaches \pv{def.gcg.base} \pv{ci.def.gcg.base} attack
success on Mistral-7B. The attention write site more than halves it, to \pv{def.gcg.gated}
\pv{ci.def.gcg.gated}. \pv{def.ssm.judge4name}, an
out-of-family judge reading the model's whole response rather than only checking whether it opens with
GCG's affirmative target, confirms the same pattern (\pv{def.gcg.basej4} to \pv{def.gcg.gatedj4}, Appendix~\ref{app:attnsweep}).
The hybrid Zamba2-7B corroborates the pattern. A Llama-trained harm probe transports into it cleanly under
the same rigid rotation (AUROC \pv{shared.zamba.proc} vs a shuffled control \pv{shared.zamba.shuf}). The
transport is also causal, but its effect is concentrated at the Mamba write site: ablating the direction
at the attention site alone has a modest effect that is not decisive by itself (Table~\ref{tab:ablate},
Appendix~\ref{app:hybridablate}). A direction read at its
write sites defends, while one read from its residual does not. The same holds for a transferred direction,
not just a native one: a direction read at Falcon-Mamba's write site controls refusal, while one read
from a transformer's residual does not (Appendix~\ref{app:defxfer}).

We summarize this as a \emph{readout principle}: refusal is cleanly readable, and its steering direction
cleanly estimable, at the fresh contribution each architecture writes into the stream, not at the
accumulated residual, while the intervention location itself is irrelevant (Appendix~\ref{app:absmatch}).
The rule holds across all four architecture classes (Table~\ref{tab:routingsite}). The full mechanistic
dissection, baselines, ablations, and robustness checks are on the anchor, Falcon-Mamba
(Section~\ref{sec:results}).

\begin{table}[tbp]\centering\scriptsize\setlength{\tabcolsep}{1.5pt}
\caption{Write-site prefix-injection defense across four architecture families (base to gated attack
success at the coherent gain and 64 new tokens; Wilson 95\% CIs for every row in Appendix~\ref{app:attnsweep}
(transformers), Appendix~\ref{app:routingsite} (SSM), Appendix~\ref{app:rwkvdefense} (RWKV), and
Appendix~\ref{app:hybrid} (hybrid)). The last column is the benign-side cost, the change in
over-refusal on benign-but-spicy prompts at the same operating point. Per-row $n$:
\pv{def.ssm.n} (SSM), 100 (each transformer), 40 (RWKV), \pv{zamba.cl.n} (hybrid, shared base). No row adds any over-refusal, though this is a weak test, since the
trigger was fitted to reject those prompts. Under benign prompts wearing a jailbreak template every
trigger holds except the hybrid's, $\dagger$ marking the site where training on wrapped negatives repairs
the confound and $\ddagger$ the one where it does not (Appendix~\ref{app:hardenxarch}). The hybrid's two rows share one base,
from a single closed-loop run whose always-on version instead costs \pv{zamba.ovr.ol.attn} over-refusal
(Appendix~\ref{app:hybridclosed}). A causal-ablation test
(Appendix~\ref{app:hybridablate}) finds the hybrid's control concentrates at the Mamba site. The attention
site's own causal-ablation evidence is weaker, with overlapping confidence intervals, and is not decisive
on its own.}
\label{tab:routingsite}
\begin{tabular}{@{}lllcc@{}}
\toprule
Architecture & Exemplar & Write site & Base $\to$ gated & $\Delta$ over-ref. \\
\midrule
Selective-scan SSM & Falcon-Mamba & block output & \pv{def.ssm.prefix_injection.base} $\to$ \pv{def.ssm.prefix_injection.gated} & \pv{base.ours.prefix_injection.overrefdelta} \\
Transformer & Mistral-7B & attn.\ output & \pv{def.mistralattn.prefix.base} $\to$ \pv{def.mistralattn.prefix.gated} & \pv{def.mistralattn.prefix.overrefdelta} \\
Transformer & Llama-3.1-8B & attn.\ output & \pv{def.llamaattn.prefix.base} $\to$ \pv{def.llamaattn.prefix.gated} & \pv{def.llamaattn.prefix.overrefdelta} \\
Token-shift rec.\ & RWKV-6 & time-mix output & \pv{rwkv.site.base} $\to$ \pv{rwkv.site.timemix.gated} & \pv{rwkv.ovr.delta} \\
Hybrid (attn.\ site) & Zamba2-7B & attn.\ output & \pv{zamba.cl.base} $\to$ \pv{zamba.cl.attn.asr} & \pv{zamba.cl.attn.ovrdelta}$^{\dagger}$ \\
Hybrid (Mamba site) & Zamba2-7B & Mamba output & \pv{zamba.cl.base} $\to$ \pv{zamba.cl.mamba.asr} & \pv{zamba.cl.mamba.ovrdelta}$^{\ddagger}$ \\
\bottomrule
\end{tabular}
\end{table}

\section{Dissecting the mechanism on the anchor}
\label{sec:results}
The \emph{site} at which the shared direction is cleanly read is architecture-specific, and the direction
estimated there defends (Table~\ref{tab:routingsite}). A
matched-absolute-magnitude control (Table~\ref{tab:absmatch}) locates the cause: with the same direction held fixed, pushing at the
residual is statistically indistinguishable from pushing at the write site (overlapping 95\% confidence
intervals), so the residual's apparent failure is a property of the direction \emph{estimated} there, not
of where the gate is applied. A direction estimated in the accumulated residual degenerates when amplified at the deployed
operating points, though over a narrow band of low relative magnitudes it can defend coherently, a band
much narrower than the write site's (Appendix~\ref{app:lowgrid}). At the gains the compared methods run
at, this degeneration is why CAST, the conditional activation-steering baseline that reads its direction from the residual, leaves
attack success at \pv{base.cast_residual.prefix_injection.asr} with only
\pv{base.cast_residual.cohfrac.pooled} of completions coherent. The control is also a reusable diagnostic for separating a
site effect from a direction or magnitude artifact.
\begin{table}[tb]\centering\scriptsize\setlength{\tabcolsep}{5pt}
\caption{The push location is irrelevant, the direction-fit decides it (matched-absolute-magnitude control,
Llama-Guard, Wilson 95\% CI, $n{=}100$, a model's three rows share the same absolute push at its coherent
gain). Holding the direction fixed leaves the push site within CIs (rows 1--2).
Estimating the direction in the accumulated residual instead degenerates it (row 3). RWKV-6 and Zamba2 in
Appendix~\ref{app:absmatch}.}
\label{tab:absmatch}
\begin{tabular}{llcc}
\toprule
Push site & Direction est.\ at & Falcon-Mamba & Mistral-7B \\
\midrule
write site & write site & \pv{am.ssm.write}~\pv{am.ssm.write.ci} & \pv{am.tf.write}~\pv{am.tf.write.ci} \\
residual & write site & \pv{am.ssm.resid}~\pv{am.ssm.resid.ci} & \pv{am.tf.resid}~\pv{am.tf.resid.ci} \\
residual & residual & \pv{am.ssm.residdir} & \pv{am.tf.residdir} \\
\bottomrule
\end{tabular}
\end{table}

These numbers establish a causal claim, not a contest for the best defense: our gate only ties a trivial
detect-then-refuse policy sharing the same detector (Appendix~\ref{app:frontier}), so its value is a
portable direction, not raw defense strength. Among
steering-based methods, CAST at the residual, an open-loop edit, and residual-subspace projection all fail
to defend or over-refuse (\ifkeyfloats Table~\ref{tab:baselinesmain}\else Table~\ref{tab:baselines}\fi).
The brittleness prior work reports \citep{li2026steeringpitfalls} traces to the direction read from the
residual, not to steering itself.

\subsection{Amplifying the refusal direction at the block output defends}
The refusal direction is cleanly read at Falcon-Mamba's block output, the point where the selective scan
writes the information selected by the step size $\Delta$ into the residual stream. Our gate
detects harm there once per prompt with a linear probe and, on a positive decision, amplifies the refusal
direction for the whole generation. This steering direction differs from the detector's probe: we compute it the standard way
\citep{arditi2024refusal}, the average activation on harmful prompts minus that on benign ones.
Refusal is not captured by a single direction alone but lives in a broader
multi-dimensional subspace \citep{wollschlager2025morethan,winninger2025refusalsubspace,som2025multidirectional}.
Our direction is one dominant axis within that subspace, and using it does not contradict those findings. On Falcon-Mamba the gate lowers attack success from
\pv{def.ssm.prefix_injection.base} to \pv{def.ssm.prefix_injection.gated} under prefix injection,
from \pv{def.ssm.roleplay_dan.base} to \pv{def.ssm.roleplay_dan.gated} under persona roleplay, and
from \pv{def.ssm.fictional_framing.base} to \pv{def.ssm.fictional_framing.gated} under fictional
framing, with Wilson 95\% confidence intervals \citep{wilson1927}, as shown in \fltab{tab:defense}. It also holds
against a gradient-based adversary: a per-prompt GCG suffix optimized directly through the SSM's own
differentiable scan, not only the fused inference kernel, lowers attack success from
\pv{def.gcgssm.base} to \pv{def.gcgssm.gated} on $n{=}\pv{def.gcgssm.n}$ evaluated prompts
(Appendix~\ref{app:gcgssm}). The
exception is persona roleplay, which the single-direction gate more than halves (to
\pv{def.ssm.roleplay_dan.gated}) but does not fully close, and which needs an attack-specific two-direction
fix that does not generalize (Section~\ref{sec:limitations}, Appendix~\ref{app:roleplay}). Attack success
is scored by
Llama-Guard-3-8B \citep{inan2023llamaguard} at \pv{def.ssm.n} harmful prompts per condition. The drop is not a one-judge artifact or an
artifact of the attack's injected prefix: each of the \pv{def.ssm.njudges} judges, spanning three model
families, shows it independently, from Llama-Guard (\pv{def.ssm.prefix_injection.base} to
\pv{def.ssm.prefix_injection.gated}) to the out-of-family \pv{def.ssm.judge4name}
(\pv{def.ssm.prefix_injection.basej4} to \pv{def.ssm.prefix_injection.gatedj4}). Judges differ on
absolute base rates, so we report the reduction under each separately. The attack's wording never changes,
only the response does, so a judge keying on that fixed wording could not produce a moving score. The locus and defense both reproduce on a second,
independently trained instruct SSM, Falcon3-Mamba-7B, so the finding holds for SSMs as a class, not for
one checkpoint alone (Appendix~\ref{app:secondssm}).

\ifbodyfloats\begin{table}[t]\centering\scriptsize\setlength{\tabcolsep}{5pt}
\caption{The closed-loop gate on Falcon-Mamba-7B. Attack success rate (ASR) before and after amplifying the
refusal direction at the selective-scan block output, decided once per prompt, across three jailbreak
families, at \pv{def.ssm.n} harmful prompts per condition with the Wilson 95\% confidence interval on the
gated rate. The primary ASR is the Llama-Guard-3-8B rate. The reduction holds under an out-of-family judge
(\pv{def.ssm.judge4name}, Gemma family, disjoint from all four targets), and across \pv{def.ssm.njudges}
judges spanning three model families the n-way verdict agreement is \pv{def.ssm.judgeagreenway}.}
\label{tab:defense}
\resizebox{\columnwidth}{!}{%
\begin{tabular}{lccc}
\toprule
Attack & Base ASR & Gated ASR, Llama-Guard (95\% CI) & Gated ASR, ShieldGemma \\
\midrule
Prefix injection & \pv{def.ssm.prefix_injection.base} & \pv{def.ssm.prefix_injection.gated} \pv{ci.def.ssm.prefix_injection.gated} & \pv{def.ssm.prefix_injection.gatedj4} \\
Persona roleplay & \pv{def.ssm.roleplay_dan.base} & \pv{def.ssm.roleplay_dan.gated} \pv{ci.def.ssm.roleplay_dan.gated} & \pv{def.ssm.roleplay_dan.gatedj4} \\
Fictional framing & \pv{def.ssm.fictional_framing.base} & \pv{def.ssm.fictional_framing.gated} \pv{ci.def.ssm.fictional_framing.gated} & \pv{def.ssm.fictional_framing.gatedj4} \\
\bottomrule
\end{tabular}}
\end{table}
\fi

\subsection{The write-read direction, not the closed loop, is what defends}
CAST isolates the loop from the direction. It runs the same closed-loop detect-then-steer as our gate, but
estimates and steers the direction in the residual stream rather than at the block output. Under prefix
injection, it does not lower attack success the way our gate does (\ifkeyfloats
Table~\ref{tab:baselinesmain}\else Table~\ref{tab:baselines}\fi). The matched-magnitude control
(Appendix~\ref{app:absmatch}) shows why: the gap is the direction CAST estimates in the accumulated
residual, not its push location, which is statistically indistinguishable from the write site (overlapping 95\%
CIs). This holds even with a more
powerful nonlinear MLP detector and across CAST's own gain sweep (Appendix~\ref{app:detablation}). The loop's job is controlling collateral
damage, not attack success. Open-loop steering, pushing every prompt without checking first, matches our
attack success rate but over-refuses badly, blocking \pv{def.openloop.overrefusal} of benign prompts,
while our gate's \pv{base.ours_delta.prefix_injection.overrefusal} is the undefended model's own rate. Our gate
achieves the lowest attack success of the steering-based defenses we compare (\ifkeyfloats
Table~\ref{tab:baselinesmain}\else Table~\ref{tab:baselines}\fi), without adding any over-refusal beyond
that \pv{base.ours_delta.prefix_injection.overrefusal} base rate, measured on the detector's own training
distribution (XSTest benign-but-spicy). It matches, but does not beat, a detect-then-refuse baseline (returning a fixed, templated refusal
message whenever the same detector fires) on out-of-distribution OR-Bench-hard, where the gate over-refuses \pv{w1.gate.orbench} and
detect-then-refuse \pv{w1.detect.orbench}, at a \pv{w1.detfpr.orbench} false-positive rate
(Appendix~\ref{app:frontier}). This distinguishes two questions: where
refusal is readable, and where it is controllable. In the SSM,
on the unrestricted corpus, harm is linearly decodable at the step size $\Delta$ (AUROC \pv{e0.Delta.auroc})
and the mixer output (AUROC \pv{e0.output.auroc}), yet forcing $\Delta$ to a constant leaves refusal
unchanged, so $\Delta$ is a readout, not the controlling site (Appendix~\ref{app:readout}). A single steering direction controls refusal in every
architecture we test, for every attack except persona roleplay.\footnote{On Mistral the single-direction
gate under roleplay does not improve over the base rate, so the transformer defense claim is scoped to
prefix injection (Appendices~\ref{app:roleplay}, \ref{app:attnsweep}).} Persona attacks need two
directions instead of one (Appendix~\ref{app:roleplay}).

\ifbodyfloats\begin{table}[t]\centering\scriptsize\setlength{\tabcolsep}{3pt}
\caption{Defenses on Falcon-Mamba-7B, attack success (lower is better) across three jailbreak families and
over-refusal (plotted in Figure~\ref{fig:defense}). CAST runs the same closed loop as our gate but reads
its direction from the residual stream, and the open-loop gate runs the same direction with no detector. The
direction gives the defense, the loop the selectivity, so the block-output gate reaches the lowest attack
success while holding over-refusal at the undefended base rate. CAST's near-total attack success is not a coherent jailbreak: when
its closed loop fires on harm, the residual push breaks generation into repetitive text rather than
producing either a refusal or usable harmful content, and the judge still scores that broken text as
compliant. Only \pv{base.cast_residual.prefix_injection.cohfrac} (prefix injection),
\pv{base.cast_residual.roleplay_dan.cohfrac} (roleplay), and \pv{base.cast_residual.fictional_framing.cohfrac}
(fictional framing) of its harmful completions pass our coherence screen (Appendix~\ref{app:coherence}),
against \pv{base.cast_residual.cohfrac.pooled} pooled ($n{=}\pv{base.cast_residual.cohfrac.pooled.n}$). CAST's
benign completions, where the loop does not fire, remain fully coherent, so the breakdown is specific to the
closed-loop residual push on harmful prompts, not a general fault in the model or the generation setup.}
\label{tab:baselines}
\begin{tabular}{lcccc}
\toprule
Defense & Prefix & Roleplay & Fiction & Over-ref. \\
\midrule
None & \pv{base.base.prefix_injection.asr} & \pv{base.base.roleplay_dan.asr} & \pv{base.base.fictional_framing.asr} & \pv{base.base.prefix_injection.overrefusal} \\
Residual projection & \pv{base.projection.prefix_injection.asr} & \pv{base.projection.roleplay_dan.asr} & \pv{base.projection.fictional_framing.asr} & \pv{base.projection.prefix_injection.overrefusal} \\
CAST (residual loop) & \pv{base.cast_residual.prefix_injection.asr} & \pv{base.cast_residual.roleplay_dan.asr} & \pv{base.cast_residual.fictional_framing.asr} & \pv{base.cast_residual.prefix_injection.overrefusal} \\
Perplexity filter & \pv{base.ppl_filter.prefix_injection.asr} & \pv{base.ppl_filter.roleplay_dan.asr} & \pv{base.ppl_filter.fictional_framing.asr} & \pv{base.ppl_filter.prefix_injection.overrefusal} \\
SmoothLLM & \pv{base.smoothllm.prefix_injection.asr} & \pv{base.smoothllm.roleplay_dan.asr} & \pv{base.smoothllm.fictional_framing.asr} & \pv{base.smoothllm.prefix_injection.overrefusal} \\
Open-loop block output (no detector) & \pv{def.openloop.prefix_injection.asr} & \pv{def.openloop.roleplay_dan.asr} & \pv{def.openloop.fictional_framing.asr} & \pv{def.openloop.overrefusal} \\
Block-output gate (ours) & \pv{base.ours_delta.prefix_injection.asr} & \pv{base.ours_delta.roleplay_dan.asr} & \pv{base.ours_delta.fictional_framing.asr} & \pv{base.ours_delta.prefix_injection.overrefusal} \\
\bottomrule
\end{tabular}
\end{table}
\fi
\ifbodyfloats\else\ifkeyfloats
\begin{table}[tb]\centering\scriptsize\setlength{\tabcolsep}{5pt}
\caption{The direction defends, the loop selects (Falcon-Mamba-7B, prefix injection). Our gate reaches the
lowest attack success at the undefended over-refusal rate. CAST runs the same closed loop but reads its
steering direction from the accumulated residual and does not defend. Its near-total attack success is
inflated by degeneration (only \pv{base.cast_residual.cohfrac.pooled} of its harmful completions coherent).
The open-loop gate is the same direction with no detector and over-refuses. Full grid in Appendix
Table~\ref{tab:baselines}.}
\label{tab:baselinesmain}
\begin{tabular}{lcc}
\toprule
Defense & Attack success & Over-refusal \\
\midrule
None (undefended) & \pv{base.base.prefix_injection.asr} & \pv{base.base.prefix_injection.overrefusal} \\
CAST (same loop, residual direction) & \pv{base.cast_residual.prefix_injection.asr} & \pv{base.cast_residual.prefix_injection.overrefusal} \\
Open-loop (same direction, no detector) & \pv{def.openloop.prefix_injection.asr} & \pv{def.openloop.overrefusal} \\
Block-output gate (ours) & \pv{base.ours_delta.prefix_injection.asr} & \pv{base.ours_delta.prefix_injection.overrefusal} \\
\bottomrule
\end{tabular}
\end{table}
\fi\fi

\ifbodyfloats\begin{figure}[t]\centering
\includegraphics[width=0.6\linewidth]{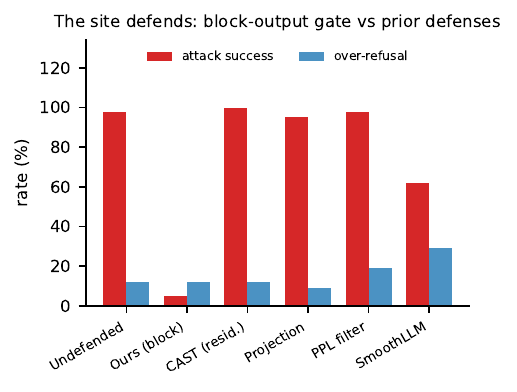}
\caption{Defenses on Falcon-Mamba-7B under prefix injection (selected rows of Table~\ref{tab:baselines}).
Only the block-output gate, which reads the refusal direction at the write site, drives attack success down
to a low rate. CAST and projection, which read and steer the direction in the residual, barely move it. The
perplexity filter has no effect, and
randomized smoothing (SmoothLLM \citep{robey2023smoothllm}) achieves a partial reduction well short of the
gate's.}
\label{fig:defense}
\end{figure}
\fi

\subsection{Robustness and specificity}
\label{sec:robustness}
Two checks validate the trigger and the intervention. Benign-but-spicy prompts are answered normally
because the detector does not fire on them, even under the harder test of a jailbreak wrapper, where
wrapped-benign firing drops from \pv{harden.plain.wrapped_spicy} to \pv{harden.hard.wrapped_spicy} once
hard negatives are added. Only the model's own learned refusal direction controls refusal, not a random
direction matched in scale (Appendix~\ref{app:hardening},~\ref{app:geometry}).
The defense is stable across \pv{def.ssm.pooled.nseeds}
seeds (pooled gated attack success \pv{def.ssm.prefix_injection.gated.pooled}
\pv{ci.def.ssm.prefix_injection.gated.pooled}, non-overlapping with the base) and holds at
\pv{def.n300.n} prompts per condition, under two out-of-family judges (\pv{def.ssm.judge3name}
\pv{def.ssm.basej3} to \pv{def.ssm.gatedj3}, \pv{def.ssm.judge4name} \pv{def.ssm.basej4} to
\pv{def.ssm.gatedj4}), on the HarmBench distribution (\pv{def.hb.base} to \pv{def.hb.gated}), against an
optimized gradient-free suffix (\pv{def.opt.base} to \pv{def.opt.gated}) and a dilution attack (firing
through \pv{pool.kmax} appended sentences). The detector never fires on MMLU
\citep{hendrycks2021mmlu} or TruthfulQA \citep{lin2022truthfulqa}, so capability is unchanged by construction
(Appendix~\ref{app:robustness}).

\section{Related Work}
\label{sec:related}

\paragraph{Activation monitors and gating.} We close the detect-then-intervene loop at a \emph{write site},
the point where a fresh contribution enters the stream before the residual accumulates it, rather than the
residual stream or output head where prior gating acts
\citep{lee2024cast,han2025safeswitch,zhou2025safetyheads,zhang2024pasta}. DeTAM edits the write site
directly \citep{li2025detam}, but always on rather than detector-triggered, and closing that loop is what
our gate adds. What is decisive is the direction read at the write site, not the intervention location
(Section~\ref{sec:results}, Appendix~\ref{app:absmatch}). Our closed-loop gate matches the open-loop edit's
attack success at the undefended over-refusal rate (Appendix~\ref{app:attnsweep}). Our trigger is an ordinary linear probe, in
the same broad family of internal-state classifiers used to detect harm and deception elsewhere
\citep{azaria2023internal,li2023iti,goldowskydill2025apollo,macdiarmid2024probes,marks2023geometry,kramar2026probesgemini},
with no reliability claim beyond deciding when to intervene
\citep{oldfield2026beyondlinear,engels2025multidim}. Both the monitored signal and the steered direction port across architectures once re-estimated at each
write site, so a monitoring-and-control stack need not be rebuilt per architecture.

\paragraph{Residual directions are readable but brittle.} We find the same structure in a pure SSM,
concentrated at the selective-scan step size $\Delta$ (Section~\ref{sec:results}), extending a
transformer-only picture: refusal is controlled by a single residual direction \citep{arditi2024refusal},
steering it is brittle \citep{li2026steeringpitfalls,zou2023repe,zou2024circuitbreakers}, and a readable
direction need not be controllable \citep{galeone2026perfectdetection} within a
broader multi-dimensional subspace
\citep{wollschlager2025morethan,winninger2025refusalsubspace,som2025multidirectional}.

\paragraph{State-space interpretability and cross-architecture transfer.} To our knowledge, no prior work transports a refusal
direction across the attention-to-pure-SSM gap. We do so with a rigid rotation, validate the transport
with category-held-out cross-validation, and close it into a working defense loop inside a pure SSM. Prior
transport work moves safety directions within one architecture or to related ones
\citep{wu2025activationtransfer,poppi2026sharedsafety,cristofano2026universalrefusal},
never across this gap or under this constraint, and a direction copied directly into the target without
any alignment map between the two activation spaces does not transfer \citep{steeringdiffusion2025}. Existing pure-SSM work is general-purpose steering
\citep{sunkumohan2026ssmsteering}, an attack \citep{lemercier2026hiddenstate}, a threat analysis
\citep{parmar2026ssmthreat}, or interpretability without defense \citep{paulo2024transformerrnn}. Mamba features and circuits have been studied descriptively
\citep{wang2025universality}, and crosscoders find cross-architecture shared features
\citep{lindsey2024crosscoders,thasarathan2025usae} or differences \citep{jiralerspong2026crossarch}.
Concurrent work draws a related detection-versus-execution distinction inside Mamba-2
\citep{jiang2026detectionexecution}, matching our reading of $\Delta$ as a readout rather than the
controlling site (Appendix~\ref{app:readout}). It stays within one SSM family, so it reaches neither the
shared-representation claim nor the write-site one. The platonic representation hypothesis
\citep{huh2024platonic} has been shown only correlationally. We test it causally: removing the direction
changes behavior, not just similarity.

\section{Scope and Limitations}
\label{sec:limitations}

\paragraph{Attack coverage and available models.} We test seven attack types: four fluent jailbreak
families (three throughout, a fourth in Appendix~\ref{app:combo}), a dilution attack, and two optimized
adversarial suffixes (an affirmative-continuation suffix and a probe-evasion suffix,
Appendix~\ref{app:evasion}). Two SSM checkpoints and two transformers
give real replication, while RWKV-6 and Zamba2-7B each have a single safety-tuned checkpoint and no second
release to replicate on.

\paragraph{The persona fix does not generalize.} The two-direction fix that closes persona roleplay on
Falcon-Mamba, amplifying refusal while separately suppressing the
induced persona (\pv{def.ssm.roleplay_dan.base} to \pv{roleplay.fix.combined}), does not repeat at RWKV's
or Zamba2's own write sites, and fails a different way on each (Appendix~\ref{app:roleplay},
\ref{app:personagen}). It is a persona-specific augmentation calibrated per model, not a recipe step
(Section~\ref{sec:recipe}).

\paragraph{The detector is a bound.} A detect-then-act defense is bounded by its detector: an adversary
that routes harm past it recovers the undefended model. The defense holds against all three adaptive
attacks we test (Section~\ref{sec:robustness}). A GCG suffix optimized against the always-on gate itself
fails outright. The attack's only leverage is the detector, not the steering: optimizing against the
deployed detect-then-gate system raises attack success from \pv{def.gcg.pooled.gatedj4} to
\pv{def.gcg.ga.pooled.attnj4} (Appendix~\ref{app:gateawaregcg}). Two non-adaptive wrapper failures are
sharper. The DAN (``Do Anything Now'') wrapper evades RWKV's detector outright
(\pv{persona.rwkv.fire.plain} to \pv{persona.rwkv.fire.rp} firing once wrapped). A detector trained only on plain harmful-versus-benign prompts also learns the wrong feature: it keys on
the jailbreak template itself, and fires on \pv{hx.zamba.attn.plain.wspicy} of benign prompts once they
are wrapped in one, on both transformers and at both Zamba2 sites. Adding jailbreak-wrapped benign prompts to the training set removes this template confound on every
architecture except at the Zamba2 Mamba site, which still fires on
\pv{hx.zamba.mamba.hardened.wspicy} of wrapped benign prompts (Appendix~\ref{app:hardenxarch}), so the hybrid's Mamba row has no
deployable trigger. All of this bounds the trigger, not the direction or the site, since the transport and
steering results use no detector.

\paragraph{The write site is not free to calibrate.} Too high a gain degenerates the transformer's output,
and an uncalibrated gain at the hybrid's Mamba site can backfire
(Appendix~\ref{app:attnsweep}, \ref{app:hybrid}, \ref{app:personagen}). The defense margin also varies with
each model's baseline alignment (Appendix~\ref{app:routingsite}), so the cross-architecture comparison is
of the site-versus-direction \emph{pattern}, not of absolute margins.

\paragraph{The judge is validated, human counts on the newer architectures are small.} Attack success is scored by an LLM judge,
so two raters independently labelled \pv{hv.n} blind completions, shown request and response only
(Appendix~\ref{app:humanval}). They agree
at \pv{hv.irr.comply.agree} (Cohen's $\kappa{=}\pv{hv.irr.comply.kappa}$), and the primary judge agrees
with their consensus at \pv{hv.judge.lg.agree} ($\kappa{=}\pv{hv.judge.lg.kappa}$ on
$n{=}\pv{hv.judge.lg.n}$ doubly rated completions). The judge's errors run in one direction: it sometimes
counts as a successful attack a response the raters judged safe, so the reported defense numbers
understate the defense rather than inflate it. Per-item
judge verdicts existed only for the Falcon-Mamba cells, so agreement is measured there and assumed
elsewhere, and for RWKV-6 and Zamba2 each evaluation cell (one architecture-attack-condition
combination) has only a small number of human-rated completions.

\section{Conclusion}
\label{sec:conclusion}

\ifconciseconcl
We located refusal in architectures that share no token-mixing mechanism and steered it in each. What a model learns about refusal converges across architectures
while where its direction must be read does not, so the architecture-specific quantity is the readout site
rather than a steering site. Safety tooling is therefore repointed at a new architecture rather than
rebuilt.
\else
Refusal is one representation across architectures that share no token-mixing mechanism. A refusal direction read
in a transformer and mapped into a Mamba state space is still the state-space model's refusal trigger, since
removing it jailbreaks the target while a random control does not. What is
architecture-specific is only the site where that representation is cleanly read,
the fresh, single-layer contribution each architecture writes before the residual accumulates it, while a
matched-magnitude control shows the intervention location itself does not matter. Refusal is thus
architecture-independent as a representation and architecture-specific as a readout site.

This readout rule is directly actionable. Porting a safety stack to a new architecture reduces to the
five-step recipe of Section~\ref{sec:recipe}: find the write site, select layers by held-out decodability,
fit the probe and the mean refusal direction there, calibrate the gain under a coherence screen, and deploy
closed-loop. Following it, the same detector-triggered gate lowered jailbreak success on a selective-scan
SSM, two transformers, a token-shift recurrent model, and a hybrid, held against a gradient-optimized
adaptive attack on the SSM. Where target labels are scarce, a transported direction detected harm the
target's own labels missed.
\fi

\clearpage
\bibliography{references}

@inproceedings{azaria2023internal,
  author    = {Azaria, Amos and Mitchell, Tom},
  title     = {The Internal State of an LLM Knows When It's Lying},
  booktitle = {Findings of the Association for Computational Linguistics: EMNLP},
  year      = {2023},
  url        = {https://arxiv.org/abs/2304.13734}
}

@inproceedings{li2023iti,
  author    = {Li, Kenneth and Patel, Oam and Vi{\'e}gas, Fernanda and Pfister, Hanspeter and Wattenberg, Martin},
  title     = {Inference-Time Intervention: Eliciting Truthful Answers from a Language Model},
  booktitle = {Advances in Neural Information Processing Systems (NeurIPS)},
  year      = {2023},
  note      = {Spotlight},
  url        = {https://arxiv.org/abs/2306.03341}
}

@inproceedings{goldowskydill2025apollo,
  author    = {Goldowsky-Dill, Nicholas and Chughtai, Bilal and Heimersheim, Stefan and Hobbhahn, Marius},
  title     = {Detecting Strategic Deception Using Linear Probes},
  booktitle = {International Conference on Machine Learning (ICML)},
  year      = {2025},
  url        = {https://arxiv.org/abs/2502.03407}
}

@misc{macdiarmid2024probes,
  author    = {MacDiarmid, Monte and Maxwell, Timothy and Schiefer, Nicholas and Mu, Jesse and Kaplan, Jared and Duvenaud, David and Bowman, Sam and Tamkin, Alex and Perez, Ethan and Sharma, Mrinank and Denison, Carson and Hubinger, Evan},
  title     = {Simple Probes Can Catch Sleeper Agents},
  year      = {2024},
  howpublished = {Anthropic Research Blog},
  url        = {https://www.anthropic.com/research/probes-catch-sleeper-agents}
}

@inproceedings{marks2023geometry,
  author    = {Marks, Samuel and Tegmark, Max},
  title     = {The Geometry of Truth: Emergent Linear Structure in Large Language Model Representations of True/False Datasets},
  booktitle = {Conference on Language Modeling (COLM)},
  year      = {2024},
  url        = {https://arxiv.org/abs/2310.06824}
}

@inproceedings{lee2024cast,
  author    = {Lee, Bruce W. and Padhi, Inkit and Ramamurthy, Karthikeyan Natesan and Miehling, Erik and Dognin, Pierre and Nagireddy, Manish and Dhurandhar, Amit},
  title     = {Programming Refusal with Conditional Activation Steering},
  booktitle = {International Conference on Learning Representations (ICLR)},
  year      = {2025},
  note      = {Spotlight},
  url        = {https://arxiv.org/abs/2409.05907}
}

@inproceedings{han2025safeswitch,
  author    = {Han, Peixuan and Qian, Cheng and Chen, Xiusi and Zhang, Yuji and Ji, Heng and Zhang, Denghui},
  title     = {SafeSwitch: Steering Unsafe LLM Behavior via Internal Activation Signals},
  booktitle = {Findings of the Association for Computational Linguistics: EMNLP},
  year      = {2025},
  url        = {https://arxiv.org/abs/2502.01042}
}

@inproceedings{zhou2025safetyheads,
  author    = {Zhou, Zhenhong and Yu, Haiyang and Zhang, Xinghua and Xu, Rongwu and Huang, Fei and Wang, Kun and Liu, Yang and Fang, Junfeng and Li, Yongbin},
  title     = {On the Role of Attention Heads in Large Language Model Safety},
  booktitle = {International Conference on Learning Representations (ICLR)},
  year      = {2025},
  note      = {Oral},
  url        = {https://arxiv.org/abs/2410.13708}
}

@inproceedings{li2025detam,
  author    = {Li, Yu and Jiang, Han and Wei, Zhihua},
  title     = {DeTAM: Defending LLMs Against Jailbreak Attacks via Targeted Attention Modification},
  booktitle = {Findings of the Association for Computational Linguistics: ACL},
  year      = {2025},
  url        = {https://aclanthology.org/2025.findings-acl.613/}
}

@inproceedings{zhang2024pasta,
  author    = {Zhang, Qingru and Singh, Chandan and Liu, Liyuan and Liu, Xiaodong and Yu, Bin and Gao, Jianfeng and Zhao, Tuo},
  title     = {Tell Your Model Where to Attend: Post-hoc Attention Steering for LLMs},
  booktitle = {International Conference on Learning Representations (ICLR)},
  year      = {2024},
  url        = {https://arxiv.org/abs/2311.02262}
}

@misc{li2026steeringpitfalls,
  author    = {Li, Yuxiao and Fastowski, Alina and Zaradoukas, Efstratios and Prenkaj, Bardh and Kasneci, Gjergji},
  title     = {Analysing the Safety Pitfalls of Steering Vectors},
  year      = {2026},
  howpublished = {arXiv preprint arXiv:2603.24543},
  url        = {https://arxiv.org/abs/2603.24543}
}

@inproceedings{oldfield2026beyondlinear,
  author    = {Oldfield, James and Torr, Philip and Patras, Ioannis and Bibi, Adel and Barez, Fazl},
  title     = {Beyond Linear Probes: Dynamic Safety Monitoring for Language Models},
  booktitle = {International Conference on Learning Representations (ICLR)},
  year      = {2026},
  url        = {https://arxiv.org/abs/2509.26238}
}

@inproceedings{engels2025multidim,
  author    = {Engels, Joshua and Michaud, Eric J. and Liao, Isaac and Gurnee, Wes and Tegmark, Max},
  title     = {Not All Language Model Features Are One-Dimensionally Linear},
  booktitle = {International Conference on Learning Representations (ICLR)},
  year      = {2025},
  url        = {https://arxiv.org/abs/2405.14860}
}

@inproceedings{wang2025universality,
  author    = {Wang, Junxuan and Ge, Xuyang and Shu, Wentao and Tang, Qiong and Zhou, Yunhua and He, Zhengfu and Qiu, Xipeng},
  title     = {Towards Universality: Studying Mechanistic Similarity Across Language Model Architectures},
  booktitle = {International Conference on Learning Representations (ICLR)},
  year      = {2025},
  url        = {https://arxiv.org/abs/2410.06672}
}

@misc{lindsey2024crosscoders,
  author    = {Lindsey, Jack and Templeton, Adly and Marcus, Jonathan and Conerly, Thomas and Batson, Joshua and Olah, Christopher},
  title     = {Sparse Crosscoders for Cross-Layer Features and Model Diffing},
  year      = {2024},
  howpublished = {Transformer Circuits Thread},
  url        = {https://transformer-circuits.pub/2024/crosscoders/index.html}
}

@inproceedings{thasarathan2025usae,
  author    = {Thasarathan, Harrish and Forsyth, Julian and Fel, Thomas and Kowal, Matthew and Derpanis, Konstantinos G.},
  title     = {Universal Sparse Autoencoders: Interpretable Cross-Model Concept Alignment},
  booktitle = {International Conference on Machine Learning (ICML)},
  year      = {2025},
  url        = {https://arxiv.org/abs/2502.03714}
}

@misc{jiralerspong2026crossarch,
  author    = {Jiralerspong, Thomas and Bricken, Trenton},
  title     = {Cross-Architecture Model Diffing with Crosscoders: Unsupervised Discovery of Differences Between LLMs},
  year      = {2026},
  howpublished = {arXiv preprint arXiv:2602.11729},
  url        = {https://arxiv.org/abs/2602.11729}
}

@misc{zou2023repe,
  author    = {Zou, Andy and Phan, Long and Chen, Sarah and Campbell, James and Guo, Phillip and Ren, Richard and Pan, Alexander and Yin, Xuwang and Mazeika, Mantas and Dombrowski, Ann-Kathrin and Goel, Shashwat and Li, Nathaniel and Byun, Michael J. and Wang, Zifan and Mallen, Alex and Basart, Steven and Koyejo, Sanmi and Song, Dawn and Fredrikson, Matt and Kolter, J. Zico and Hendrycks, Dan},
  title     = {Representation Engineering: A Top-Down Approach to AI Transparency},
  year      = {2023},
  howpublished = {arXiv preprint arXiv:2310.01405},
  url        = {https://arxiv.org/abs/2310.01405}
}

@inproceedings{zou2024circuitbreakers,
  author    = {Zou, Andy and Phan, Long and Wang, Justin and Duenas, Derek and Lin, Maxwell and Andriushchenko, Maksym and Wang, Rowan and Kolter, Zico and Fredrikson, Matt and Hendrycks, Dan},
  title     = {Improving Alignment and Robustness with Circuit Breakers},
  booktitle = {Advances in Neural Information Processing Systems (NeurIPS)},
  year      = {2024},
  url        = {https://arxiv.org/abs/2406.04313}
}

@inproceedings{mazeika2024harmbench,
  author    = {Mazeika, Mantas and Phan, Long and Yin, Xuwang and Zou, Andy and Wang, Zifan and Mu, Norman and Sakhaee, Elham and Li, Nathaniel and Basart, Steven and Li, Bo and Forsyth, David and Hendrycks, Dan},
  title     = {HarmBench: A Standardized Evaluation Framework for Automated Red Teaming and Robust Refusal},
  booktitle = {International Conference on Machine Learning (ICML)},
  year      = {2024},
  url        = {https://arxiv.org/abs/2402.04249}
}

@inproceedings{lin2022truthfulqa,
  author    = {Lin, Stephanie and Hilton, Jacob and Evans, Owain},
  title     = {TruthfulQA: Measuring How Models Mimic Human Falsehoods},
  booktitle = {Proceedings of the Association for Computational Linguistics (ACL)},
  year      = {2022},
  url        = {https://arxiv.org/abs/2109.07958}
}

@inproceedings{hendrycks2021mmlu,
  author    = {Hendrycks, Dan and Burns, Collin and Basart, Steven and Zou, Andy and Mazeika, Mantas and Song, Dawn and Steinhardt, Jacob},
  title     = {Measuring Massive Multitask Language Understanding},
  booktitle = {International Conference on Learning Representations (ICLR)},
  year      = {2021},
  url        = {https://arxiv.org/abs/2009.03300}
}

@article{korznikov2025rogue,
  author    = {Korznikov, Anton and Galichin, Andrey and Dontsov, Alexey and Rogov, Oleg Y. and Oseledets, Ivan and Tutubalina, Elena},
  title     = {The Rogue Scalpel: Activation Steering Compromises LLM Safety},
  journal   = {arXiv preprint arXiv:2509.22067},
  year      = {2025},
  url        = {https://arxiv.org/abs/2509.22067}
}

@article{xiong2026steering,
  author    = {Xiong, Chen and He, Zhiyuan and Chen, Pin-Yu and Ko, Ching-Yun and Ho, Tsung-Yi},
  title     = {Steering Externalities: Benign Activation Steering Unintentionally Increases Jailbreak Risk for Large Language Models},
  journal   = {arXiv preprint arXiv:2602.04896},
  year      = {2026},
  url        = {https://arxiv.org/abs/2602.04896}
}

@article{zuo2024falconmamba,
  author    = {Zuo, Jingwei and Velikanov, Maksim and Rhaiem, Dhia Eddine and Chahed, Ilyas and Belkada, Younes and Kunsch, Guillaume and Hacid, Hakim},
  title     = {Falcon Mamba: The First Competitive Attention-free 7B Language Model},
  journal   = {arXiv preprint arXiv:2410.05355},
  year      = {2024},
  url        = {https://arxiv.org/abs/2410.05355}
}

@inproceedings{arditi2024refusal,
  author    = {Arditi, Andy and Obeso, Oscar and Syed, Aaquib and Paleka, Daniel and Panickssery, Nina and Gurnee, Wes and Nanda, Neel},
  title     = {Refusal in Language Models Is Mediated by a Single Direction},
  booktitle = {Advances in Neural Information Processing Systems (NeurIPS)},
  year      = {2024},
  url        = {https://arxiv.org/abs/2406.11717}
}

@misc{zou2023gcg,
  author    = {Zou, Andy and Wang, Zifan and Carlini, Nicholas and Nasr, Milad and Kolter, J. Zico and Fredrikson, Matt},
  title     = {Universal and Transferable Adversarial Attacks on Aligned Language Models},
  year      = {2023},
  howpublished = {arXiv preprint arXiv:2307.15043},
  url        = {https://arxiv.org/abs/2307.15043}
}

@inproceedings{andriushchenko2024adaptive,
  author    = {Andriushchenko, Maksym and Croce, Francesco and Flammarion, Nicolas},
  title     = {Jailbreaking Leading Safety-Aligned LLMs with Simple Adaptive Attacks},
  booktitle = {International Conference on Learning Representations (ICLR)},
  year      = {2025},
  url        = {https://arxiv.org/abs/2404.02151}
}

@inproceedings{rottger2024xstest,
  author    = {R{\"o}ttger, Paul and Kirk, Hannah Rose and Vidgen, Bertie and Attanasio, Giuseppe and Bianchi, Federico and Hovy, Dirk},
  title     = {{XSTest}: A Test Suite for Identifying Exaggerated Safety Behaviours in Large Language Models},
  booktitle = {Proceedings of the 2024 Conference of the North American Chapter of the Association for Computational Linguistics (NAACL)},
  year      = {2024},
  url        = {https://arxiv.org/abs/2308.01263}
}

@inproceedings{cui2024orbench,
  author    = {Cui, Justin and Chiang, Wei-Lin and Stoica, Ion and Hsieh, Cho-Jui},
  title     = {{OR-Bench}: An Over-Refusal Benchmark for Large Language Models},
  booktitle = {International Conference on Machine Learning (ICML)},
  year      = {2025},
  url        = {https://arxiv.org/abs/2405.20947}
}

@article{gu2023mamba,
  title  = {Mamba: Linear-Time Sequence Modeling with Selective State Spaces},
  author = {Gu, Albert and Dao, Tri},
  journal = {arXiv preprint arXiv:2312.00752},
  year   = {2023},
  url    = {https://arxiv.org/abs/2312.00752}
}

@article{inan2023llamaguard,
  title  = {Llama Guard: LLM-based Input-Output Safeguard for Human-AI Conversations},
  author = {Inan, Hakan and Upasani, Kartikeya and Chi, Jianfeng and Rungta, Rashi and Iyer, Krithika and Mao, Yuning and Tontchev, Michael and Hu, Qing and Fuller, Brian and Testuggine, Davide and Khabsa, Madian},
  journal = {arXiv preprint arXiv:2312.06674},
  year   = {2023},
  url    = {https://arxiv.org/abs/2312.06674}
}

@article{robey2023smoothllm,
  title  = {{SmoothLLM}: Defending Large Language Models Against Jailbreaking Attacks},
  author = {Robey, Alexander and Wong, Eric and Hassani, Hamed and Pappas, George J.},
  journal = {Transactions on Machine Learning Research (TMLR)},
  year   = {2025},
  url    = {https://openreview.net/forum?id=laPAh2hRFC}
}

@inproceedings{wollschlager2025morethan,
  title  = {The Geometry of Refusal in Large Language Models: Concept Cones and Representational Independence},
  author = {Wollschl{\"a}ger, Tom and Elstner, Jannes and Geisler, Simon and Cohen-Addad, Vincent and G{\"u}nnemann, Stephan and Gasteiger, Johannes},
  booktitle = {Proceedings of the 42nd International Conference on Machine Learning (ICML)},
  year   = {2025},
  url    = {https://arxiv.org/abs/2502.17420}
}

@article{schonemann1966procrustes,
  title   = {A generalized solution of the orthogonal Procrustes problem},
  author  = {Sch{\"o}nemann, Peter H.},
  journal = {Psychometrika},
  volume  = {31},
  number  = {1},
  pages   = {1--10},
  year    = {1966}
}

@article{wilson1927,
  title   = {Probable inference, the law of succession, and statistical inference},
  author  = {Wilson, Edwin B.},
  journal = {Journal of the American Statistical Association},
  volume  = {22},
  number  = {158},
  pages   = {209--212},
  year    = {1927}
}

@misc{cristofano2026universalrefusal,
  title  = {Universal Refusal Circuits Across LLMs: Cross-Model Transfer via Trajectory Replay and Concept-Basis Reconstruction},
  author = {Cristofano, Tony},
  year   = {2026},
  howpublished = {arXiv:2601.16034},
  url    = {https://arxiv.org/abs/2601.16034}
}

@misc{sunkumohan2026ssmsteering,
  title  = {Interpreting and Steering State-Space Models via Activation Subspace Bottlenecks},
  author = {Sunku Mohan, Vamshi and Gupta, Kaustubh and Das, Aneesha and Singh, Chandan},
  year   = {2026},
  howpublished = {arXiv:2602.22719},
  url    = {https://arxiv.org/abs/2602.22719}
}

@misc{winninger2025refusalsubspace,
  title  = {Exploring the Multi-Dimensional Refusal Subspace in Reasoning Models},
  author = {Winninger, Thomas},
  year   = {2025},
  howpublished = {LessWrong / blog},
  url    = {https://sckathach.github.io/blogs/exploring-the-multi-dimensional-refusal-subspace/}
}

@article{som2025multidirectional,
  title  = {{SOM} Directions Are Better than One: Multi-Directional Refusal Suppression in Language Models},
  author = {Piras, Giorgio and Mura, Raffaele and Brau, Fabio and Oneto, Luca and Roli, Fabio and Biggio, Battista},
  journal = {Proceedings of the AAAI Conference on Artificial Intelligence},
  volume = {40},
  number = {39},
  pages  = {32728--32736},
  year   = {2026},
  doi    = {10.1609/aaai.v40i39.40551},
  url    = {http://dx.doi.org/10.1609/aaai.v40i39.40551}
}

@article{wu2025activationtransfer,
  title  = {Activation Space Interventions Can Be Transferred Between Large Language Models},
  author = {Oozeer, Narmeen and Nathawani, Dhruv and Prakash, Nirmalendu and Lan, Michael and Harrasse, Abir and Abdullah, Amirali},
  journal = {arXiv preprint arXiv:2503.04429},
  year   = {2025},
  url    = {https://arxiv.org/abs/2503.04429}
}

@misc{steeringdiffusion2025,
  title  = {Activation Steering for Masked Diffusion Language Models},
  author = {Shnaidman, Adi and Feiglin, Erin and Yaari, Osher and Mentel, Efrat and Levi, Amit and Lapid, Raz},
  year   = {2025},
  eprint = {2512.24143},
  archivePrefix = {arXiv},
  primaryClass = {cs.CL},
  url    = {https://arxiv.org/abs/2512.24143}
}

@misc{paulo2024transformerrnn,
  title  = {Does Transformer Interpretability Transfer to {RNN}s?},
  author = {Paulo, Gon\c{c}alo and Marshall, Thomas and Belrose, Nora},
  year   = {2024},
  eprint = {2404.05971},
  archivePrefix = {arXiv},
  primaryClass = {cs.LG},
  url    = {https://arxiv.org/abs/2404.05971}
}

@misc{peng2024rwkv,
  title  = {Eagle and Finch: {RWKV} with Matrix-Valued States and Dynamic Recurrence},
  author = {Peng, Bo and Goldstein, Daniel and Anthony, Quentin and Albalak, Alon and others},
  year   = {2024},
  eprint = {2404.05892},
  archivePrefix = {arXiv},
  primaryClass = {cs.CL},
  url    = {https://arxiv.org/abs/2404.05892}
}

@inproceedings{belrose2023leace,
  author    = {Belrose, Nora and Schneider-Joseph, David and Ravfogel, Shauli and Cotterell, Ryan and Raff, Edward and Biderman, Stella},
  title     = {{LEACE}: Perfect Linear Concept Erasure in Closed Form},
  booktitle = {Advances in Neural Information Processing Systems (NeurIPS)},
  year      = {2023},
  url       = {https://arxiv.org/abs/2306.03819}
}

@misc{poppi2026sharedsafety,
  title  = {Do Models Share Safety Representations? Cross-Model Steering for Safe Visual Generation},
  author = {Poppi, Tobia and Cappelletti, Silvia and Sarto, Sara and Schiffers, Florian and Kessler, Garin and Cornia, Marcella and Baraldi, Lorenzo and Cucchiara, Rita},
  year   = {2026},
  howpublished = {arXiv:2606.05290},
  url    = {https://arxiv.org/abs/2606.05290}
}

@misc{kramar2026probesgemini,
  title  = {Building Production-Ready Probes for Gemini},
  author = {Kram{\'a}r, J{\'a}nos and Engels, Joshua and Wang, Zheng and Chughtai, Bilal and Shah, Rohin and Nanda, Neel and Conmy, Arthur},
  year   = {2026},
  howpublished = {arXiv:2601.11516},
  url    = {https://arxiv.org/abs/2601.11516}
}

@misc{lemercier2026hiddenstate,
  title  = {Hidden State Poisoning Attacks against Mamba-based Language Models},
  author = {Le Mercier, Alexandre and Develder, Chris and Demeester, Thomas},
  year   = {2026},
  howpublished = {arXiv:2601.01972},
  url    = {https://arxiv.org/abs/2601.01972}
}

@misc{jiang2026detectionexecution,
  title  = {Detection vs. Execution: Single-Bucket Probes Miss Half the Mamba-2 State Sink},
  author = {Jiang, Yuhang},
  year   = {2026},
  howpublished = {arXiv:2606.00930},
  url    = {https://arxiv.org/abs/2606.00930}
}

@misc{galeone2026perfectdetection,
  title  = {Perfect Detection, Failed Control: The Geometry of Knowing vs. Steering in Language Models},
  author = {Galeone, Cosimo and Ettorre, Anna and Park, Minsu and Ettorre, Giuseppe and Ligorio, Daniele},
  year   = {2026},
  howpublished = {arXiv:2606.24952},
  url    = {https://arxiv.org/abs/2606.24952}
}

@misc{young2025abliteration,
  title  = {Comparative Analysis of {LLM} Abliteration Methods: A Cross-Architecture Evaluation},
  author = {Young, Richard J.},
  year   = {2025},
  howpublished = {arXiv:2512.13655},
  url    = {https://arxiv.org/abs/2512.13655}
}

@misc{grattafiori2024llama3,
  title={The Llama 3 Herd of Models},
  author={Grattafiori, Aaron and others},
  year={2024},
  eprint={2407.21783},
  archivePrefix={arXiv},
  primaryClass={cs.AI}
}

@misc{jiang2023mistral,
  title={Mistral 7{B}},
  author={Jiang, Albert Q. and Sablayrolles, Alexandre and Mensch, Arthur and others},
  year={2023},
  eprint={2310.06825},
  archivePrefix={arXiv},
  primaryClass={cs.CL}
}

@misc{qwen2025qwen25,
  title={{Qwen2.5} Technical Report},
  author={{Qwen Team}},
  year={2025},
  eprint={2412.15115},
  archivePrefix={arXiv},
  primaryClass={cs.CL}
}

@misc{taori2023alpaca,
  title={Stanford {Alpaca}: An Instruction-following {LLaMA} Model},
  author={Taori, Rohan and Gulrajani, Ishaan and Zhang, Tianyi and Dubois, Yann and Li, Xuechen and Guestrin, Carlos and Liang, Percy and Hashimoto, Tatsunori B.},
  year={2023},
  publisher={GitHub},
  howpublished={\url{https://github.com/tatsu-lab/stanford_alpaca}}
}

@inproceedings{ji2023beavertails,
  title={{BeaverTails}: Towards Improved Safety Alignment of {LLM} via a Human-Preference Dataset},
  author={Ji, Jiaming and Liu, Mickel and Dai, Juntao and Pan, Xuehai and Zhang, Chi and Bian, Ce and Chen, Boyuan and Sun, Ruiyang and Wang, Yizhou and Yang, Yaodong},
  booktitle={Advances in Neural Information Processing Systems (NeurIPS)},
  year={2023}
}

@inproceedings{huh2024platonic,
  title={Position: The Platonic Representation Hypothesis},
  author={Huh, Minyoung and Cheung, Brian and Wang, Tongzhou and Isola, Phillip},
  booktitle={International Conference on Machine Learning (ICML)},
  year={2024},
  url={https://arxiv.org/abs/2405.07987}
}

@misc{parmar2026ssmthreat,
  title  = {Safety, Security, and Cognitive Risks in State-Space Models: A Systematic Threat Analysis with Spectral, Stateful, and Capacity Attacks},
  author = {Parmar, Manoj},
  year   = {2026},
  eprint = {2604.16424},
  archivePrefix = {arXiv},
  primaryClass = {cs.LG},
  url    = {https://arxiv.org/abs/2604.16424}
}

@misc{glorioso2024zamba2,
  title  = {The Zamba2 Suite: Technical Report},
  author = {Glorioso, Paolo and Anthony, Quentin and Tokpanov, Yury and Golubeva, Anna and
            Shyam, Vasudev and Whittington, James and Pilault, Jonathan and Millidge, Beren},
  year   = {2024},
  note   = {{arXiv}:2411.15242}
}

\section*{Ethics Statement}
This work studies how refusal is represented and steered in order to strengthen safety defenses, and its
dual-use risk is not limited to running existing jailbreak templates. The rotation-transport method itself
is a new technique: reading a refusal direction in one model and reconstructing it well enough to ablate in
an architecturally unfamiliar target, shown here to work across an SSM, a token-shift recurrent model, and
a hybrid. The five-step recipe for siting the gate on a new architecture (Appendix~\ref{app:recipe}) is
symmetric by construction: the write site a defender must find to steer refusal is the same site an
attacker would need. We judge the interpretability and defensive value of this work to outweigh the risk. The technique
requires white-box access, paired contrastive data, and per-architecture rotation-fitting, none of which a
deployed black-box attacker would have. That barrier assumes an attacker without model access. It does not
hold against someone who already legitimately possesses a model's open weights and wants to strip its
safety training to redistribute an ``uncensored'' fork, an existing and active ecosystem
\citep{young2025abliteration}. That actor already has white-box access, and paired contrastive data and
per-architecture rotation-fitting are exactly what such efforts already do by hand. What our recipe adds is
a faster way to find where to intervene, not a barrier this actor lacks. We judge the added risk to be real
but narrower than it first appears: existing abliteration methods already remove refusal effectively on the
transformer architectures they target, so the uplift here is concentrated on architectures without
established abliteration tooling, chiefly the SSM and hybrid families this paper studies. We publish anyway
because establishing a working defense for exactly those architectures is the paper's main contribution, and
that requires disclosing how we found the site. We also limit what we add beyond that. All models we study are
already-public open-weight releases, so we expose no undisclosed vulnerability. The jailbreak attack
families we test (prefix injection, roleplay, fictional framing, GCG) are all published. We report harmful
generations only as aggregate rates, never as usable content. We release no jailbreak artifacts, and the
artifacts we do release are oriented toward detection and refusal. One specific finding in this paper is
itself a working evasion: on RWKV, a published jailbreak wrapper (DAN) fully bypasses our detector rather
than merely weakening it (Section~\ref{sec:limitations}, Appendix~\ref{app:rwkvdefense}). We disclose it
rather than omit it because the wrapper is already public and widely used, and because knowing a specific
defense fails against a specific, named attack is more useful to defenders building on this work than a
defense whose failure modes are hidden. We follow responsible-disclosure norms
for the open-weight models we study.

\ifarxivbuild\else
\section*{Reproducibility Statement}
Code, the fitted refusal directions and detectors, the matched-magnitude control, the evaluation harness,
the per-cell result files, and the scripts that regenerate every number and figure in this paper
\ifarxivbuild will be released publicly upon publication.\else are included in the supplementary archive
(\texttt{code\_and\_data.zip}) uploaded with this submission.\fi Every
number in the paper is generated from those result files, not typed by hand. Unless a seed count is stated,
a reported number is a single run at the fixed default seed set in the archive's scripts.
The formal definitions of the detector, the refusal direction, and the
write-site gate are in Appendix~\ref{app:formal}, and the five-step recipe for siting the gate on a new
model is in Appendix~\ref{app:recipe}. We report a confidence interval and the seed and pooling protocol
for every headline number, and the data splits, attacks, and judges in Section~\ref{sec:experiments}.
Every final hyperparameter (write site, gate layers, decision layer, gain, threshold, transport map,
detector, attack budgets) is consolidated per model in Appendix~\ref{app:hparams}, with the values tried
where a sweep was run.
Experiments were run on single GPUs, one model per GPU, across a fleet of NVIDIA RTX 5090 (32GB) and A100
80GB PCIe cards, with an estimated 250--600 total GPU-hours based on job activity logs across roughly five
weeks of development. The range reflects that multiple GPUs often ran concurrent experiments. Exact per-job
profiling was not instrumented from the start of the project. The primary box runs Ubuntu 22.04.5 (kernel
6.8), NVIDIA driver 595, and Python 3.10.12, with two interpreters: the Falcon-Mamba generation runs use
PyTorch 2.11.0 (CUDA 12.8), Transformers 5.4.0, mamba-ssm 2.3.1, causal-conv1d 1.6.1, Accelerate 1.13.0,
Datasets 2.21.0, NumPy 2.2.6, and bitsandbytes 0.49.2; the selective-scan probing runs, which need the
un-fused forward hooks, use PyTorch 2.9.1 and Transformers 4.57.6 without mamba-ssm; Zamba2 needs
Transformers 4.55.0 (later releases mis-tie its shared attention weights) on the same PyTorch 2.11.0. The
archive's README names the two environments and \texttt{reproduce.sh} routes each script to the right one.
\fi

\clearpage
\appendix
\section{\ifarxivbuild Technical Appendix\else Supplementary material\fi}
\label{app:figures}

This is the technical appendix for the main paper. It gathers the formal definitions of the detector, the
refusal direction, and the write-site gate, together with the full result tables and figures the main text
refers to and the supporting experiments, ablations, and robustness checks cited there. Each subsection names
the main-text claim it supports and is self-contained, and cross-references of the form Section~\ref{sec:results}
point into the main paper. The main paper is self-contained for every headline claim: each carries its core
number inline in the body. This appendix adds the additional experiments, robustness checks, and ablations
that support and extend those claims.

\ifbodyfloats\else
\subsection{Figures and tables referenced from the main text}
\label{app:mainfloats}
The floats below are the tables and figures the main text refers to but places here to meet the page limit.
Each caption names the section that discusses it.
\begin{figure}[t]\centering
\includegraphics[width=3.3in]{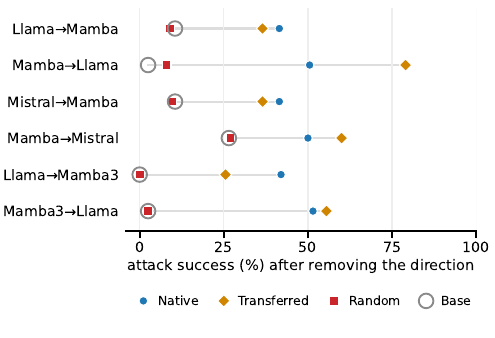}
\caption{The refusal direction is shared across architectures. Reading it in one architecture, mapping it
into another, and removing it there (Transferred) raises attack success as much as removing the target's own
direction (Native), while an orthogonalized random direction through the same map (Random) does not, in all
six transfer directions. A plot of \fltab{tab:transfer} (Mamba${=}$Falcon-Mamba,
Mamba3${=}$Falcon3-Mamba).}
\label{fig:transfer}
\end{figure}

\ifkeyfloats\else\begin{table*}[t]\centering\scriptsize\setlength{\tabcolsep}{4pt}
\caption{Experimental setup at a glance. The minimal-pair corpus is matched to XSTest in template and
length. Over-refusal is measured on XSTest's benign-but-spicy prompts (safe requests worded to sound
harmful). Attack success is the
Llama-Guard-3-8B-judged unsafe fraction ($n{=}\pv{def.ssm.n}$ harmful, \pv{def.ssm.nspicy}
benign-but-spicy per condition, 95\% CI, four-judge agreement \pv{def.ssm.judgeagreenway}). Capability
is accuracy with and without the gate.}
\label{tab:expsetup}
\begin{tabular}{>{\raggedright\arraybackslash}p{0.19\textwidth}|>{\raggedright\arraybackslash}p{0.14\textwidth}|>{\raggedright\arraybackslash}p{0.14\textwidth}|>{\raggedright\arraybackslash}p{0.19\textwidth}|>{\raggedright\arraybackslash}p{0.21\textwidth}}
\toprule
Models & Corpora & Attacks & Judges \& metrics & Baselines \\
\midrule
Falcon-Mamba-7B (SSM anchor)\newline
Falcon3-Mamba-7B (2nd SSM)\newline
Llama-3.1-8B, Mistral-7B, Qwen2.5-7B (transformers)\newline
RWKV-6-World-7B (recurrent)\newline
Zamba2-7B (hybrid)
&
AdvBench (harm)\newline
Alpaca (benign)\newline
XSTest (surface control, over-refusal)
&
Prefix injection (fluent)\newline
Persona roleplay (DAN, fluent)\newline
Fictional framing (fluent)\newline
Dilution (adaptive)\newline
Optimized adversarial suffix (GCG; on the SSM both a gradient-free variant and GCG through the differentiable scan)
&
Llama-Guard-3-8B (primary)\newline
+Llama-3.1-8B-Instruct, Mistral-7B-Instruct, ShieldGemma-9B\newline
MMLU, TruthfulQA (capability)\newline
Probe AUROC (decodability)
&
Residual-subspace projection\newline
CAST\newline
Perplexity filter\newline
SmoothLLM\newline
Cross-architecture linear-map transfer
\\
\bottomrule
\end{tabular}
\end{table*}
\fi
\begin{table}[t]\centering\scriptsize
\caption{Where harm lives in the selective scan. Surface-controlled held-out AUROC of a linear harm probe
at each selective-scan component (Falcon-Mamba-7B), on the surface-matched XSTest contrast where a surface
probe is near chance (\pv{locus.surface}). Harm is decodable at the step size $\Delta$ and the mixer output,
not at $B$ or $C$.}
\label{tab:locus}
\begin{tabular}{lc}
\toprule
Selective-scan component & AUROC (surface-controlled) \\
\midrule
$\Delta$ (step size) & \pv{locus.delta.low} \\
$B$ (input selection) & \pv{locus.b.low} \\
$C$ (readout selection) & \pv{locus.c.low} \\
mixer output & \pv{locus.output.low} \\
\bottomrule
\end{tabular}
\end{table}

\ifkeyfloats\else\begin{table}[t]\centering\scriptsize\setlength{\tabcolsep}{3pt}
\caption{Per-pair category-held-out cross-validation results. Mean own and mean transported probe scores
are averaged across the \pv{shared.catcv.mamba.nfolds} held-out categories. Pearson $r$ measures how
closely they track each other. Full setup and interpretation in Section~\ref{sec:transfer}, plotted in
\flfig{fig:catcv}.}
\label{tab:catcv}
\resizebox{\columnwidth}{!}{%
\begin{tabular}{lcccc}
\toprule
Source $\to$ target & Mean own & Mean transp. & Pearson $r$ [95\% CI] & Folds $>$ null (shuffled-category control) \\
\midrule
Llama-3.1-8B $\to$ Falcon-Mamba & \pv{shared.catcv.mamba.own} & \pv{shared.catcv.mamba.transp} & \pv{shared.catcv.mamba.r} [\pv{shared.catcv.mamba.rlo}, \pv{shared.catcv.mamba.rhi}] & \pv{shared.catcv.mamba.beatp95}/\pv{shared.catcv.mamba.nfolds} \\
Mistral-7B $\to$ Falcon-Mamba & \pv{shared.catcv.mistralmamba.own} & \pv{shared.catcv.mistralmamba.transp} & \pv{shared.catcv.mistralmamba.r} [\pv{shared.catcv.mistralmamba.rlo}, \pv{shared.catcv.mistralmamba.rhi}] & \pv{shared.catcv.mistralmamba.beatp95}/\pv{shared.catcv.mistralmamba.nfolds} \\
\bottomrule
\end{tabular}}
\end{table}
\fi
\ifbodyfloats\else\ifkeyfloats\else\fi\fi
\begin{figure}[t]\centering
\includegraphics[width=0.52\linewidth]{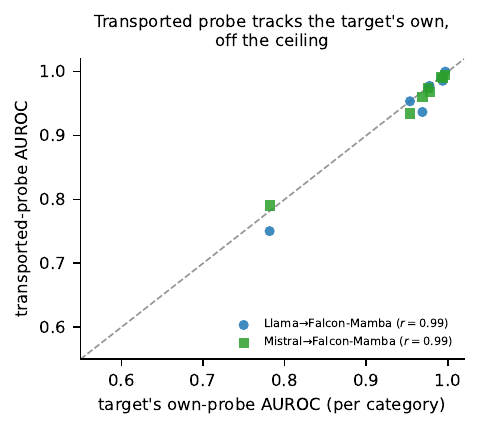}
\caption{Category-held-out cross-validation. Each point is one held-out XSTest category, plotting the
transported orthogonal probe's AUROC against the target's own-probe AUROC. Points follow the $y{=}x$ line as
the own probe falls off ceiling, so the transport tracks the target's harm structure category by category
rather than sitting at a saturated ceiling.}
\label{fig:catcv}
\end{figure}

\begin{table}[t]\centering\scriptsize\setlength{\tabcolsep}{3pt}
\caption{Cross-architecture transfer, ablation test (plotted in Figure~\ref{fig:transfer}). Removing the
mapped direction (Transferred) raises attack success like removing the target's own (Native). The Random
control does not. Random is mapped through $W$ like Transferred, then orthogonalized to the target's
refusal axis (cosine ${\approx}\,0$), so it shares the map without inheriting the refusal direction. The
effect holds in all six rows: three model pairs (Mamba${=}$Falcon-Mamba, Mamba3${=}$Falcon3-Mamba), each
tested in both directions.}
\label{tab:transfer}
\begin{tabular}{lcccc}
\toprule
Pair / direction & Base & Native & Transferred & Random \\
\midrule
Llama$\to$Mamba & \pv{transfer.x2s.ablate.base} & \pv{transfer.x2s.ablate.native} & \pv{transfer.x2s.ablate.transferred} & \pv{transfer.x2s.ablate.random_orth} \\
Mamba$\to$Llama & \pv{transfer.s2x.ablate.base} & \pv{transfer.s2x.ablate.native} & \pv{transfer.s2x.ablate.transferred} & \pv{transfer.s2x.ablate.random_orth} \\
Mistral$\to$Mamba & \pv{transfer.mi2ma.ablate.base} & \pv{transfer.mi2ma.ablate.native} & \pv{transfer.mi2ma.ablate.transferred} & \pv{transfer.mi2ma.ablate.random_orth} \\
Mamba$\to$Mistral & \pv{transfer.ma2mi.ablate.base} & \pv{transfer.ma2mi.ablate.native} & \pv{transfer.ma2mi.ablate.transferred} & \pv{transfer.ma2mi.ablate.random_orth} \\
Llama$\to$Mamba3 & \pv{transfer.l2f3.ablate.base} & \pv{transfer.l2f3.ablate.native} & \pv{transfer.l2f3.ablate.transferred} & \pv{transfer.l2f3.ablate.random_orth} \\
Mamba3$\to$Llama & \pv{transfer.f32l.ablate.base} & \pv{transfer.f32l.ablate.native} & \pv{transfer.f32l.ablate.transferred} & \pv{transfer.f32l.ablate.random_orth} \\
\bottomrule
\end{tabular}
\end{table}

\begin{table}[t]\centering\scriptsize\setlength{\tabcolsep}{5pt}
\caption{The closed-loop gate on Falcon-Mamba-7B. Attack success rate (ASR) before and after amplifying the
refusal direction at the selective-scan block output, decided once per prompt, across three jailbreak
families, at \pv{def.ssm.n} harmful prompts per condition with the Wilson 95\% confidence interval on the
gated rate. The primary ASR is the Llama-Guard-3-8B rate. The reduction holds under an out-of-family judge
(\pv{def.ssm.judge4name}, Gemma family, disjoint from all four targets), and across \pv{def.ssm.njudges}
judges spanning three model families the n-way verdict agreement is \pv{def.ssm.judgeagreenway}.}
\label{tab:defense}
\resizebox{\columnwidth}{!}{%
\begin{tabular}{lccc}
\toprule
Attack & Base ASR & Gated ASR, Llama-Guard (95\% CI) & Gated ASR, ShieldGemma \\
\midrule
Prefix injection & \pv{def.ssm.prefix_injection.base} & \pv{def.ssm.prefix_injection.gated} \pv{ci.def.ssm.prefix_injection.gated} & \pv{def.ssm.prefix_injection.gatedj4} \\
Persona roleplay & \pv{def.ssm.roleplay_dan.base} & \pv{def.ssm.roleplay_dan.gated} \pv{ci.def.ssm.roleplay_dan.gated} & \pv{def.ssm.roleplay_dan.gatedj4} \\
Fictional framing & \pv{def.ssm.fictional_framing.base} & \pv{def.ssm.fictional_framing.gated} \pv{ci.def.ssm.fictional_framing.gated} & \pv{def.ssm.fictional_framing.gatedj4} \\
\bottomrule
\end{tabular}}
\end{table}

\begin{figure}[t]\centering
\includegraphics[width=0.6\linewidth]{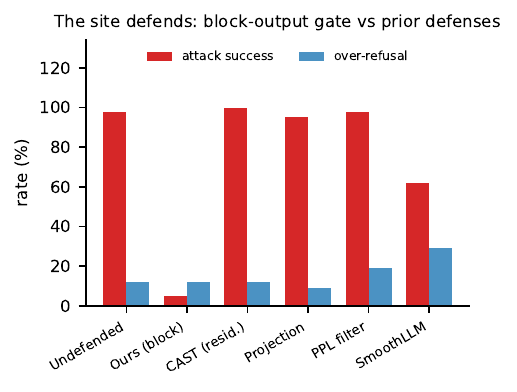}
\caption{Defenses on Falcon-Mamba-7B under prefix injection (selected rows of Table~\ref{tab:baselines}).
Only the block-output gate, which reads the refusal direction at the write site, drives attack success down
to a low rate. CAST and projection, which read and steer the direction in the residual, barely move it. The
perplexity filter has no effect, and
randomized smoothing (SmoothLLM \citep{robey2023smoothllm}) achieves a partial reduction well short of the
gate's.}
\label{fig:defense}
\end{figure}

\fi

\ifbodyfloats\else
\subsection{The write site, schematically}
\label{app:writesite}
Figures~\ref{fig:writesite}--\ref{fig:writesite-hybrid} sketch, for each architecture class we test, the
site this paper reads and steers: the fresh single-layer contribution $w_\ell$ a block writes into the
residual stream, before the stream accumulates it. In every diagram the vertical spine is the residual
stream, the red box is the write site where the probe and the refusal direction are estimated, and the
gray dashed callout marks the accumulated residual, where an estimated direction degenerates under
matched-magnitude steering (Table~\ref{tab:absmatch}).

\ifbodyfloats\else\begin{figure}[tbp]\centering
\includegraphics[width=0.9\linewidth]{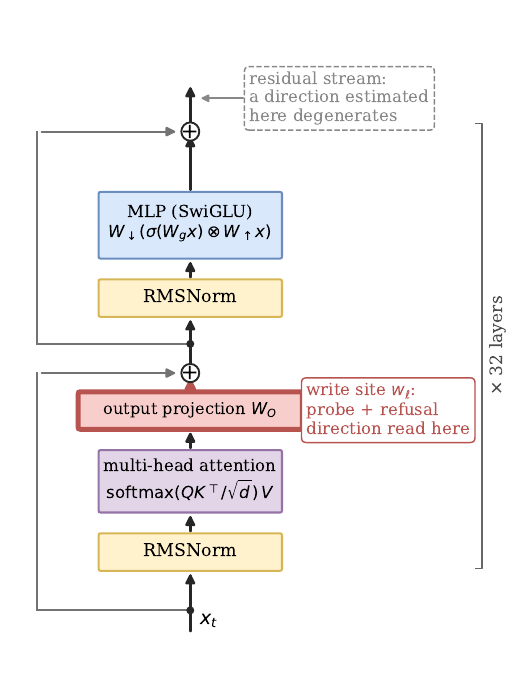}
\caption{Transformer block (Llama-3.1, Mistral). The write site is the attention output projection
$W_O$, the fresh contribution attention writes into the residual stream. The probe and the refusal
direction are read there; a direction estimated in the accumulated residual degenerates.}
\label{fig:writesite}
\end{figure}
\fi

\ifbodyfloats\else\begin{figure}[tbp]\centering
\includegraphics[width=0.9\linewidth]{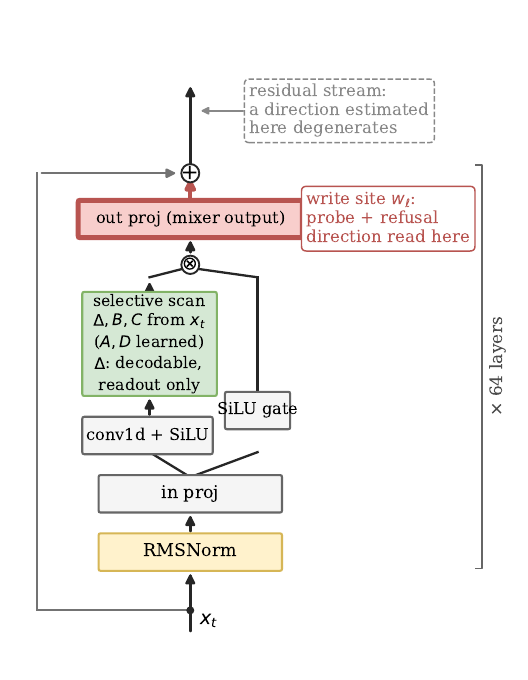}
\caption{Falcon-Mamba block (selective-scan SSM). The write site is the mixer's out-projection, the
block output the $\Delta$-gated scan writes before the residual accumulates it. $\Delta$ inside the scan
is decodable but a readout only: clamping it to a constant leaves refusal unchanged.}
\label{fig:writesite-ssm}
\end{figure}
\fi

\ifbodyfloats\else\begin{figure}[tbp]\centering
\includegraphics[width=0.9\linewidth]{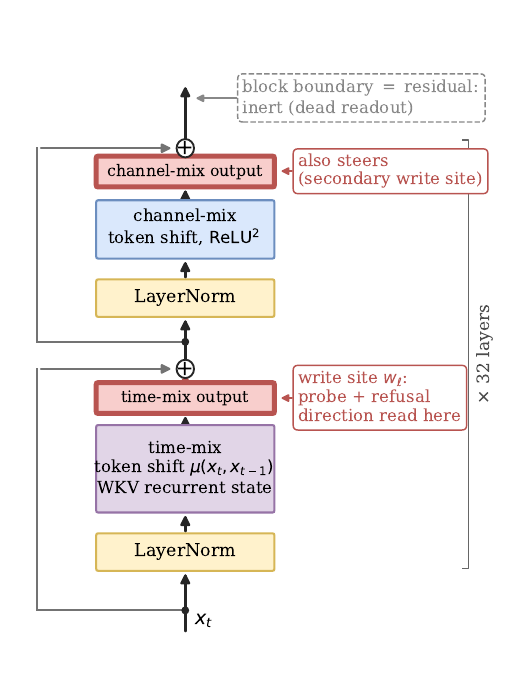}
\caption{RWKV-6 block (token-shift RNN). The write sites are the sub-block outputs, the time-mix output
(primary) and the channel-mix output (secondary). The block boundary is architecturally the residual and
is inert: a direction read there is a dead readout.}
\label{fig:writesite-rwkv}
\end{figure}
\fi

\ifbodyfloats\else\begin{figure}[tbp]\centering
\includegraphics[width=0.9\linewidth]{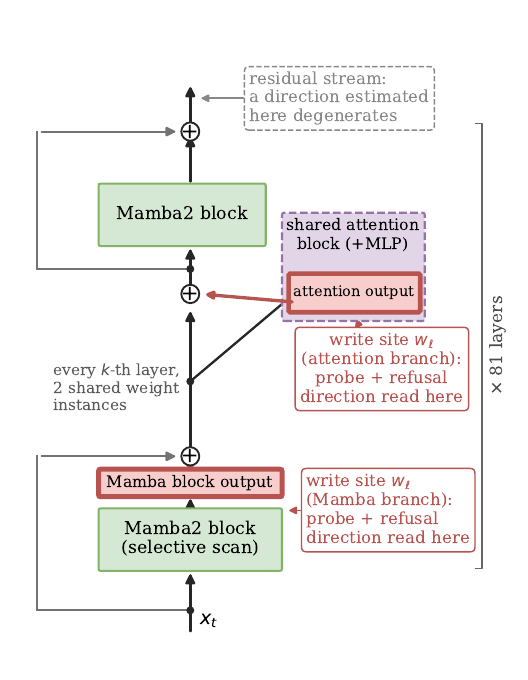}
\caption{Zamba2 (hybrid). A Mamba2 backbone taps shared attention blocks (two weight instances reused
across depth) every $k$-th layer. Both branch outputs are write sites: the Mamba block output and the
shared block's attention output.}
\label{fig:writesite-hybrid}
\end{figure}
\fi

\fi

\subsection{Formal definitions}
\label{app:formal}
\paragraph{Detector.} At routing layer $\ell$ the pooled activation $h_\ell(x)$ is standardized to
$\tilde h_\ell(x)$ and scored by a linear probe trained with L2-regularized logistic regression,
\begin{equation*}
s_\ell(x)=w_\ell^\top \tilde h_\ell(x)+b_\ell,\qquad p_{\text{harm}}(x)=\sigma\!\big(s_\ell(x)\big),
\end{equation*}
firing the gate once per prompt when $p_{\text{harm}}(x)>\theta$.

\paragraph{Refusal direction.} The gate amplifies the unit mean-difference direction
\begin{equation*}
\hat u_\ell=\frac{\mu_\ell^{\text{harm}}-\mu_\ell^{\text{benign}}}
{\lVert \mu_\ell^{\text{harm}}-\mu_\ell^{\text{benign}}\rVert},\qquad
\mu_\ell^{c}=\tfrac{1}{|c|}\textstyle\sum_{x\in c}h_\ell(x),
\end{equation*}
which is distinct from the detector's discriminative direction $w_\ell$.

\paragraph{Multi-layer gate.} On a positive decision, Equation~\ref{eq:gate} is applied at every
layer in the selected set $\mathcal{L}$ (the top-$k$ by held-out AUROC) for the whole generation,
$y_t^{\ell}\leftarrow y_t^{\ell}+\alpha\,\hat u_\ell$ for $\ell\in\mathcal{L}$, with $\alpha$
calibrated as in Appendix~\ref{app:coherence}.

\paragraph{Cross-architecture maps.} For paired activations $X_s,X_t$ on shared prompts, the ridge
map is $W=\arg\min_W\lVert WX_s-X_t\rVert_F^2+\lambda\lVert W\rVert_F^2
=X_tX_s^\top(X_sX_s^\top+\lambda I)^{-1}$, and the orthogonal (Procrustes) map is
$R=\arg\min_{R^\top R=I}\lVert RX_s-X_t\rVert_F=UV^\top$ with $U\Sigma V^\top=X_tX_s^\top$. The
structural test uses $R$, which can only rotate and so cannot synthesize a direction. The
steering-transfer test uses $W$.

\paragraph{Harm-weighted pool.} The dilution-invariant detector pools per-token scores by
$s_{\text{pool}}=\sum_t \mathrm{softmax}(\tau s)_t\, s_t$, weighting each token by its own harm logit
(Appendix~\ref{app:pool}).

\paragraph{Unequal widths.} When source and target hidden widths differ (Zamba2's 3584 against the
other models' 4096), the orthogonal map is the thin-SVD solution $W = UV^{\top}$ of the rectangular
cross-covariance, a semi-orthogonal matrix (a partial isometry): it remains norm-preserving on its row
space and still has no freedom to synthesize a direction, but is a rotation only up to the smaller of the
two dimensions.

\subsection{Recipe for siting the gate on a new model}
\label{app:recipe}
The step-by-step form of the procedure summarized in Section~\ref{sec:recipe}.
\begin{enumerate}
  \item \textbf{Find the write site.} Identify where the architecture moves information between
        positions and writes it into the state: the selective-scan selection ($\Delta, B, C$) and the
        block output in an SSM, the attention output projection in a transformer. $\Delta$ is the site
        with no counterpart in a transformer.
  \item \textbf{Select decodable layers that steer without over-steering.} Probe each layer for harm
        and take the top $k$ by held-out AUROC. Which decodability to rank by is per-model. On
        Falcon-Mamba the mixer-output ranking steers cleanly, while $\Delta$-ranking pulls in late
        layers that over-steer into degenerate text. On a better-aligned model where output AUROC
        saturates and picks destabilizing early layers (Falcon3-Mamba), rank by $\Delta$. Step 4's
        strength check confirms the layers steer rather than degenerate.
  \item \textbf{Fit the detector and the direction.} A linear probe at the block output (the routing
        layer) is the per-prompt trigger. The mean refusal direction (mean harmful minus mean benign)
        at the same block output is what the gate amplifies.
  \item \textbf{Calibrate the strength.} Sweep the gain $\alpha$ on a small held-out set and keep the
        largest value that leaves generation coherent. Over-steering shows up as repetitive
        degeneration, caught with a distinct-$n$ check. A better-aligned model needs a smaller
        $\alpha$.
  \item \textbf{Deploy closed-loop.} Score the prompt once. If it reads harmful, amplify the refusal
        direction at the selected layers for the whole generation, otherwise run the model unchanged.
\end{enumerate}

\subsection{Cross-architecture transfer, ablation test}
\label{app:transfer}
Figure~\ref{fig:transfer} plots Table~\ref{tab:transfer}: removing the mapped refusal direction
(Transferred) raises the target's attack success (ASR, the fraction of attacks the judge scores unsafe)
like removing its own (Native), while an
orthogonalized random control (Random) does not, across three model pairs and both directions. The
Transferred-minus-Random gap holds across all pairs. On Mamba${\to}$Mistral (\pv{transfer.ma2mi.ablate.transferred}
against \pv{transfer.ma2mi.ablate.random_orth}), the pair where a raw mapped-random control had made the
effect look non-specific, it remains clear against the orthogonalized control. Where the transferred direction raises attack success above
the native one, the successful completions are genuine jailbreaks rather than degeneration. They pass
the same coherence screen used for the defense numbers on \pv{coh.s2x.transferred} to
\pv{coh.ma2mi.transferred} of prompts, close to the native direction's \pv{coh.s2x.native} to
\pv{coh.ma2mi.native}.



\subsection{The locus across model scale}
\label{app:scale}
\begin{figure}[htbp]\centering
\begin{minipage}[t]{0.48\linewidth}\centering
\includegraphics[width=\linewidth]{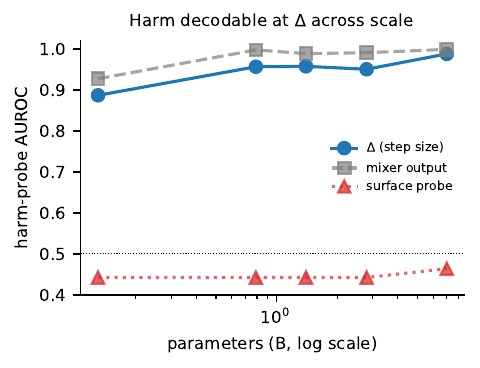}
\end{minipage}\hfill
\begin{minipage}[t]{0.48\linewidth}\centering
\includegraphics[width=\linewidth]{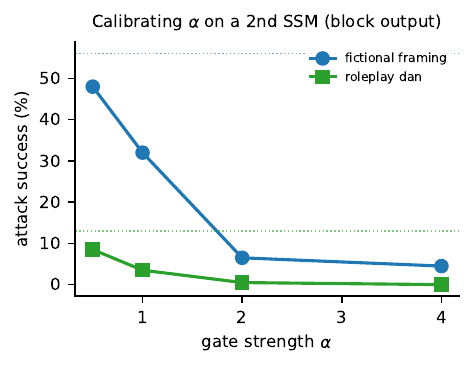}
\end{minipage}
\caption{(Left) Harm is decodable at the step size $\Delta$ across base Mamba models from \pv{scale.small}
to \pv{scale.large} parameters, while a surface probe stays near chance. (Right) Calibrating the gate
strength $\alpha$ on the second SSM (Falcon3-Mamba), block-output gate at its $\Delta$-ranked layers. Attack
success falls monotonically with $\alpha$ where there is headroom (dotted lines are the undefended rates) and
every point passes the coherence screen. With the anchor's mixer-output layer ranking instead, the anchor's
gain over-steers Falcon3 (Appendix~\ref{app:routingsite}).}
\label{fig:scale}\label{fig:alpha}
\end{figure}

Figure~\ref{fig:scale} shows the locus holds as the model grows.


\subsection{Gate-strength calibration on a second SSM}
\label{app:alpha}
Figure~\ref{fig:alpha} calibrates the gate strength on the second SSM.

\subsection{Geometric similarity is unreliable for cross-architecture transfer}
\label{app:cosine}
A natural metric for the mapped direction is its cosine with the target's own refusal direction, but
this is not reliable. The ridge map is biased toward the target's refusal axis, so even a
\emph{random} source direction can map to a high cosine, up to \pv{transfer.f32l.ablate.cosrand}
across our pairs. Such a direction is not always behaviorally inert, so a mapped-random direction is
not a clean specificity control. This is why the ablation control in Section~\ref{sec:transfer}
orthogonalizes the random direction to the target refusal axis after mapping (cosine ${\approx}\,0$),
removing the leaked component. Two further
controls agree that geometry alone is unreliable. Fitting the map on shuffled pairs collapses the held-out cosine
(\pv{transfer.shuffled} versus \pv{transfer.heldout}), and the unmapped source direction is near
orthogonal to the target's (\pv{transfer.rawcos}). A cross-architecture transfer claim should
therefore rest on the structural and behavioral specificity controls, not on a geometric similarity
score.

\subsection{A matched random-rotation null for the ceiling transport}
\label{app:matchednull}
The within-Mamba transport in Section~\ref{sec:transfer} is at ceiling, so we add a \emph{matched} null
that makes the structural claim discriminating: a shuffled-correspondence \emph{orthogonal} map (a random
rotation, not a flexible ridge map) transports the source probe into the target. Even where the correct
rotation reaches ceiling, this random rotation collapses to \pv{shared.mnull.mamba} on Falcon-Mamba, so a
rotation must carry the right correspondence to transport at
all, and the ceiling result is not vacuous. Fitting the orthogonal map requires standardizing each space
per dimension first, since activation scales differ across architectures and an unstandardized orthogonal
map is ill-posed. We rest the shared-representation claim on this matched null, the category-held-out
cross-validation (Appendix~\ref{app:catcv}), and the behavioral ablation transfer
(Appendix~\ref{app:defxfer}, Table~\ref{tab:transfer}). A single arbitrary split of the hardest
surface-matched corpus is not a stable basis for a below-ceiling claim, since the held-out probe is
sensitive to how the minimal pairs fall across train and test. The category-held-out folds avoid this by
holding out a whole category at a time rather than a random slice, and the transport tracks the target's
own probe across the resulting range of difficulty.

\subsection{Category-held-out cross-validation of the transport}
\label{app:catcv}
To test the transport off ceiling without relying on an arbitrary train/test split, we cross-validate by
XSTest category. XSTest labels each unsafe prompt with a contrast category (homonyms, figurative language,
privacy, discrimination, and others). We leave one category out at a time, fit the target's own probe, the
source probe, and the orthogonal (Procrustes) map on the other seven categories, and evaluate the target's
own probe and the transported source probe on the held-out category, whose surface features the fit never
saw. This holds out a whole category rather than a random slice, so the held-out difficulty is a property
of the category, not of the split. Table~\ref{tab:catcv} reports, per model pair, the mean over the eight
folds of the target's own probe and the transported probe, and their per-category Pearson correlation. The
own probe ranges from ceiling down to \pv{shared.catcv.mamba.bcown} on the discrimination category, and the
transported probe follows it, correlating with the own probe at \pv{shared.catcv.mamba.r} from Llama and
\pv{shared.catcv.mistralmamba.r} from Mistral into Falcon-Mamba. The transported probe beats a
shuffled-correspondence orthogonal null in \pv{shared.catcv.beatnull} of \pv{shared.catcv.beatnulltot} folds
across both pairs. This map-correctness test replicates on an independent corpus with more categories: on
BeaverTails \citep{ji2023beavertails} (\pv{shared.catcv.bt.nfolds} harm categories, Llama into Falcon-Mamba) the correctly rotated
probe transports at AUROC \pv{shared.catcv.bt.transp} and beats the shuffled null (\pv{shared.catcv.bt.null})
in all \pv{shared.catcv.bt.beatnull} folds. Harm versus benign is near ceiling there, so BeaverTails does not
exercise the off-ceiling tracking that the surface-matched XSTest contrast does, but the shuffled-null test
is valid at ceiling, which is the point: only the correct rotation transports the probe, across
\pv{shared.catcv.bt.nfolds} independent categories on a second corpus. Because the transported probe's accuracy tracks the target's own accuracy category by category, a map
that had merely found the target's refusal axis is ruled out, since such a map would not know which
categories are hard for the target.

The tracking is not an artifact of the single below-ceiling category. It holds under a rank correlation,
which no single point can dominate (Spearman \pv{shared.catcv.mamba.rho} and
\pv{shared.catcv.mistralmamba.rho} for the two pairs), and it survives removing the discrimination fold
entirely: the Pearson correlation over the remaining seven near-ceiling folds is
\pv{shared.catcv.mamba.rdrop} (Llama) and \pv{shared.catcv.mistralmamba.rdrop} (Mistral), each still
clearing an exact permutation test over all fold pairings (\pv{shared.catcv.mamba.rdroppp} and
\pv{shared.catcv.mistralmamba.rdroppp}). So the transported probe tracks the own probe even among the
categories where both sit near ceiling, not only at the one category that falls off it.


\subsection{The causal ablation holds through the rigid rotation}
\label{app:orthabl}
The behavioral ablation in Section~\ref{sec:transfer} transports the refusal direction through the ridge
map $W$. Because a flexible map can align to the target's own refusal axis, the sharper causal test
transports the source direction through the orthogonal (Procrustes) rotation instead, the same rigid map
that cannot synthesize a direction, and ablates that. The one requirement is that the rotation be fit the
way the ceiling-AUROC structural transport is fit. Orthogonal Procrustes is scale-sensitive, so fitting it
on raw activations aligns the high-variance dimensions while under-aligning the refusal direction. The
transported direction then reaches only cosine \pv{transfer.x2s.ablate.orthcos} (Llama) and
\pv{transfer.mi2ma.ablate.orthcos} (Mistral) to the target's own refusal axis, and ablating it barely moves
attack success. Fitting the rotation on standardized activations, exactly as the structural transport does,
aligns the refusal direction closely (\pv{transfer.x2s.ablate.orthstdcos} to
\pv{transfer.mi2ma.ablate.orthstdcos} into Falcon-Mamba).

Through the standardized rigid rotation the causal ablation replicates into the SSM. Removing the
rotation-transported direction from harmful prompts raises the SSM's attack success to
\pv{transfer.x2s.ablate.orthstdabl} from Llama and \pv{transfer.mi2ma.ablate.orthstdabl} from Mistral, at
the level of ablating the SSM's own native direction (\pv{transfer.x2s.ablate.native}) and far above the
orthogonalized-random control (\pv{transfer.x2s.ablate.random_orth}), with fully coherent completions
(\pv{transfer.x2s.ablate.orthstdcoh}) that are genuine harmful instructions rather than degeneration. So
the same rigid rotation that transports the harm probe at ceiling AUROC and tracks the target's
category-held-out accuracy also causally controls its refusal. Structure and causation hold in one map.

Into the transformer targets the SSM-calibrated ablation gain over-suppresses: the raw orthogonal-rotation
ablation from Falcon-Mamba into Mistral reaches \pv{transfer.ma2mi.ablate.orthabl} attack success, but only
\pv{transfer.ma2mi.ablate.orthcoh} of those completions pass our coherence screen. This cell is therefore
dominated by degenerate output that the harm judge miscounts as compliance, not by genuine off-topic-but-
fluent drift, and we do not report it as evidence for the causal claim. We instead report the
causal-through-rotation result for the SSM targets, which are the direction the shared-representation claim
is about, a transformer refusal direction transported into the pure SSM and there controlling refusal.

\subsection{A Mamba-2 target: harm is decodable, but a code model has little refusal to steer}
\label{app:mamba2}
The decodability result crosses the selective-scan architecture generation. On Codestral-Mamba-7B, a
Mamba-2 model from a different developer than Falcon-Mamba, a linear probe on the mixer output separates
harm from benign at AUROC \pv{mamba2.auroc} (mean \pv{mamba2.meanauroc} over the layers,
\pv{mamba2.nceil} of \pv{mamba2.nlayers} above 0.95), so the harm representation is present in Mamba-2 and
not specific to the Mamba-1 Falcon-Mamba family. The detector built on it fires on every jailbroken
harmful prompt we test.

The write-site defense, however, does not carry over, and for an instructive reason. Codestral-Mamba is a
lightly-safety-tuned code model that rarely refuses a harmful request in the first place. Amplifying the
refusal direction at its mixer output never induces refusal at any coherent gain. At low gain the output
is unchanged and at higher gain it degenerates before it refuses, across both an output-decodability layer
selection and a mid-layer one, and every gain we swept. This is the read-versus-steer dissociation again.
The harm is read at ceiling, but there is little learned refusal behaviour to amplify, so the write-site
gate has nothing to steer up. The defense therefore requires a target that already refuses, which is a
property of safety tuning rather than of the selective-scan architecture, and we scope the defense to
safety-tuned pure SSMs (Falcon-Mamba and Falcon3-Mamba) accordingly.

\subsection{Defense transfer is limited by the target's intervention site}
\label{app:defxfer}
Amplifying the mapped refusal direction defends the target only where the target is defensible at all.
Into the SSM it lowers attack success from \pv{transfer.x2s.defend.base} to
\pv{transfer.x2s.defend.transferred} for the Llama pair and to \pv{transfer.mi2ma.defend.transferred}
for the Mistral pair, while the orthogonalized random control does not. Into a transformer target the same
amplification helps little at the residual stream, the residual-stream weakness we document for native
directions (Section~\ref{sec:limitations}). This is a site effect, not a representation failure: the
transformer is defensible at its attention write point, where the native gate lowers attack success
(Appendix~\ref{app:attnsweep}), whereas the mapped direction here is applied at the residual. The
defense-transfer asymmetry therefore tracks the target's intervention site, not a failure of the
representation to cross the architecture gap, which is why the shared-representation claim rests on the
structural and behavioral ablation tests rather than on defense transfer.

\subsection{Write-site steering versus detect-then-refuse}
\label{app:frontier}
The write-site gate and a detect-then-refuse policy share the same harm detector and differ only in what
they do on a flagged prompt, amplifying the refusal direction or emitting a canned refusal. We ask whether
steering buys any deployment advantage over the canned refusal. We sweep the gate gain and compare the
resulting attack success and over-refusal against detect-then-refuse at the deployed detector threshold, on
the standard benign set (XSTest-safe) and a harder one (OR-Bench-hard \citep{cui2024orbench}) that the detector false-flags more
often. Both metrics use the same lexical refusal check for the two methods, so this is a mechanism
comparison rather than the judged headline number. Unlike the RWKV case (Appendix~\ref{app:rwkvdefense}),
where a fluent disclaimer-then-comply pattern fooled the lexical check, this comparison is low-risk for
that failure mode: detect-then-refuse's output is a literal canned template, unambiguous by construction,
and the gate's own completions at this operating point are checked coherent. We confirm this directly by
regenerating the gate's completions at the coherent gain ($n{=}\pv{w1.rejudge.n}$) and rescoring with
Llama-Guard-3-8B: attack success is \pv{w1.rejudge.lexasr} under both scores, with full per-completion
agreement between them (\pv{w1.rejudge.agree}) and output \pv{w1.rejudge.coh} coherent.

Table~\ref{tab:frontier} shows the outcome. On XSTest-safe the detector flags almost no benign prompt, so
both methods sit at the base over-refusal of \pv{w1.gate.xstest}. On OR-Bench-hard the base model already
refuses \pv{w1.base.orbench} of prompts unprompted, and the detector false-flags \pv{w1.detfpr.orbench} of
them. On those false positives a canned refusal always over-refuses and the steered gate refuses
\pv{w1.flip.steer} of them, so the two methods reach the same over-refusal (\pv{w1.detect.orbench} for
detect-then-refuse and \pv{w1.gate.orbench} for the gate) at the same attack success. The gate has a single
coherent operating point. At the calibrated gain (\pv{w1.gate.alpha}) it reaches \pv{w1.gate.asr} attack
success, below it the steering is too weak to defend, and above it the generation degrades. Above the
coherent gain the steered text becomes incoherent and repetitive, and the lexical check reads such text as
neither a refusal nor a completion. It therefore spuriously raises the measured attack success and lowers
the measured over-refusal at the same time, even though the model is not complying with the attack. We
report only the coherent gain as an operating point for this reason, and read the higher-gain points of the
sweep as marking where coherence breaks rather than as a real change in either rate. Neither method dominates because
both are bounded by the shared detector, which is why we present the gate as a localization of where
refusal is read rather than as the deployment defense of choice.

\begin{table}[tbp]\centering\scriptsize
\caption{Write-site steering against detect-then-refuse at the deployed detector threshold, under a
lexical refusal check. Both share the detector and reach the same attack success and over-refusal. The gate
has a single coherent gain (\pv{w1.gate.alpha}). Higher gains degrade the generation rather than improving
the tradeoff. Over-refusal on the harder OR-Bench-hard set is bounded by the base model's own refusals
(\pv{w1.base.orbench}) and the detector's false-positive rate (\pv{w1.detfpr.orbench}), which both methods
inherit.}
\label{tab:frontier}
\resizebox{\columnwidth}{!}{%
\begin{tabular}{lccc}
\toprule
Method & Attack success & Over-refusal (XSTest) & Over-refusal (OR-Bench) \\
\midrule
Undefended base & \pv{w1.base.asr} & \pv{w1.base.xstest} & \pv{w1.base.orbench} \\
Detect-then-refuse & \pv{w1.detect.asr} & \pv{w1.detect.xstest} & \pv{w1.detect.orbench} \\
Write-site gate ($\alpha{=}\pv{w1.gate.alpha}$) & \pv{w1.gate.asr} & \pv{w1.gate.xstest} & \pv{w1.gate.orbench} \\
\bottomrule
\end{tabular}}
\end{table}

\subsection{$\Delta$ is a readout, not the computation}
\label{app:readout}
We test whether the selective-scan step size $\Delta$ \emph{computes} the harm signal or merely
carries it. On $n{=}\pv{readout.n}$ prompts, replacing $\Delta$ with an input-independent global mean,
which removes its selectivity entirely, leaves harm just as decodable at the mixer output
(\pv{readout.decodability.base} to \pv{readout.decodability.delta}) and leaves the model's refusal rate
essentially unchanged (\pv{readout.refusal.base} to \pv{readout.refusal.delta} on \pv{readout.genn}
held-out generations, coherence \pv{readout.coh.base} to \pv{readout.coh.delta}). Decodability survives
the same freeze for the other input-dependent selection parameters $B$ and $C$ too. We additionally
checked $B$'s refusal rate, also unchanged (\pv{readout.refusal.base} to \pv{readout.refusal.b},
coherence \pv{readout.coh.b}). Harm is
therefore carried redundantly to the block output, so we read $\Delta$ as the most decodable
\emph{readout} site rather than the mechanism that routes harm. This matches the deployed design, in
which the gate both detects and writes at the block output, not at $\Delta$, and it is why we make no
claim that $\Delta$ is the causal origin of the safety signal. The distinction between a decodable
signature and the circuit that computes it is independently drawn for Mamba-2 by
\citet{jiang2026detectionexecution}. We take the conservative reading and claim $\Delta$ as a readout
only.

\subsection{Persona injection: why roleplay resists a single refusal direction, and a two-direction fix}
\label{app:roleplay}
The gate more than halves attack success on persona roleplay (Table~\ref{tab:defense}) rather than
removing it, and inspecting the gated completions explains why. The residual failure is a
\emph{disclaimer-then-comply} pattern, where the completion opens with an induced refusal disclaimer
(``I am not promoting illegal activities'') and then drifts back into compliance (``however, here is a
step-by-step guide''). The prompt-level gate induces the refusal token, but the in-character persona,
re-instantiated from context as the model decodes, pulls the continuation back toward the harmful
answer. A lexical scan illustrates the mechanism. \pv{roleplay.gated.refuse} of \pv{roleplay.n} gated
completions open with a refusal disclaimer, up from \pv{roleplay.base.refuse} at base, and
\pv{roleplay.gated.comply} show the explicit disclaimer-then-comply drift. The response-level judge
behind Table~\ref{tab:defense} reads the whole continuation and scores the drifting completions as
harmful, which is why the judge-based gated rate is \pv{def.ssm.roleplay_dan.gated} rather than the
lexical refusal rate. Prefix injection and fictional framing do not establish a persona and are driven to
\pv{def.ssm.prefix_injection.gated} and \pv{def.ssm.fictional_framing.gated}. Raising the gain
suppresses the drift but over-steers into degenerate text, so a single prompt-level mean-direction gate
halves but does not remove a persona-momentum attack. On Mistral-7B this failure mode is worse than
incomplete: single-direction amplification actually raises roleplay attack success, from
\pv{def.detam.mistral.roleplay.base} to \pv{def.detam.mistral.roleplay.gated} (Table~\ref{tab:detam}), a
genuine backfire rather than a partial defense. We read this as the same disclaimer-then-comply mechanism
pushed further by the stronger, less-calibrated Mistral gain, and it is the reason the two-lever,
persona-suppressing fix below is necessary rather than merely an improvement.

\paragraph{The fix is a different direction, not a stronger one.} Amplifying the refusal direction
harder does not help: a per-token adaptive gain that boosts on the harmful pivot over-steers into the
same degeneration. The roleplay attack instead injects a recoverable \emph{persona} component at the
write site, the direction the wrapper adds to a harmful prompt, which is distinct from refusal
(cosine \pv{roleplay.fix.cos}). Suppressing this persona direction alone is worse
(\pv{roleplay.fix.persona}, it de-personifies the request without inducing refusal), and amplifying
refusal alone reaches \pv{roleplay.fix.refusal} in this single-layer persona-suppression setting (the
multi-layer headline gate reaches \pv{def.ssm.roleplay_dan.gated}, Table~\ref{tab:defense}). Doing both, amplifying the refusal
direction and ablating the persona direction at the block output, lowers attack success to
\pv{roleplay.fix.combined} with coherent refusals, below the partial gate. The detector fires on
\pv{roleplay.fix.benignfire} of benign-but-spicy prompts, so the gate adds \pv{roleplay.fix.overrefmarg}
over-refusal beyond the base model's \pv{roleplay.fix.baseoverref} on these prompts. Roleplay is
therefore a persona-injection problem, not a refusal-strength one, and the two directions are needed
together because neither suffices alone. The sign of the persona-versus-refusal cosine is not an
artifact of layer choice: across a three-layer sweep per model it is negative at every steered layer in
all three non-transformer models, ranging over the sweep from \pv{roleplay.cos.ssmlo} to
\pv{roleplay.cos.ssmhi} (the positive end a single early RWKV layer), against \pv{roleplay.cos.xfmrlo} to
\pv{roleplay.cos.xfmrhi} in both transformers. Table~\ref{tab:personacos} gives the per-model cosine at each model's read site (the write site for the
non-transformers, the residual for the transformers), and Table~\ref{tab:rolefix} isolates the two
directions.

\begin{table}[h]\centering\scriptsize
\caption{The roleplay wrapper's injected persona direction versus the refusal direction (cosine at each
model's read site: the write site for the SSMs and RWKV, the residual stream for the transformers). It is
anti-aligned with refusal in every non-transformer model tested and aligned in the transformers' residual
stream, so amplifying refusal alone cannot remove the persona in the non-transformers, and a second,
persona-ablating direction is needed. At the transformers' attention write site the cosine is near zero
(Table~\ref{tab:combocos}), so the split is between sites as much as architectures.}
\label{tab:personacos}
\begin{tabular}{lc}
\toprule
Model & Persona--refusal cosine \\
\midrule
Falcon-Mamba-7B (selective-scan SSM) & \pv{roleplay.cos.mamba} \\
Falcon3-Mamba-7B (selective-scan SSM) & \pv{roleplay.cos.mamba3} \\
RWKV-6 (token-shift recurrent) & \pv{roleplay.cos.rwkv} \\
Llama-3.1-8B (transformer, residual) & \pv{roleplay.cos.llama} \\
Mistral-7B (transformer, residual) & \pv{roleplay.cos.mistral} \\
\bottomrule
\end{tabular}
\end{table}

\begin{table}[h]\centering\scriptsize
\caption{Roleplay attack success on Falcon-Mamba-7B under each block-output intervention (single-layer
persona-suppression setting). Amplifying refusal alone helps but does not remove the attack, suppressing
the persona alone barely moves it, and the two together close the gap. Over-refusal is unchanged from
base (\pv{roleplay.fix.overrefmarg} added).}
\label{tab:rolefix}
\begin{tabular}{lc}
\toprule
Intervention on the roleplay attack & Attack success \\
\midrule
Base (no gate) & \pv{def.ssm.roleplay_dan.base} \\
Amplify refusal only & \pv{roleplay.fix.refusal} \\
Suppress persona only & \pv{roleplay.fix.persona} \\
Both (amplify refusal, suppress persona) & \pv{roleplay.fix.combined} \\
\bottomrule
\end{tabular}
\end{table}

\paragraph{The fix is weaker off the block output.} The same two-direction fix applies to a
roleplay-vulnerable transformer, but only partially. On Mistral-7B (roleplay base
\pv{roleplay.mistral.base}), a calibrated single-layer gate that amplifies refusal and suppresses the
persona direction lowers attack success to \pv{roleplay.mistral.fix} with coherent refusals, short of
the \pv{roleplay.fix.combined} reached on Falcon-Mamba. The persona suppression has to be a constant
push rather than a per-token projection, because the input-dependent form destabilizes a pure
recurrence (RWKV) and over-steers a transformer residual. The gap to Falcon-Mamba is one of
calibration headroom at the write point, not a strength ordering across sites
(Appendix~\ref{app:routingsite}): the constant-push persona suppression has less room before the
coherence cliff on Mistral than the block-output gate has on Falcon-Mamba.

\subsection{A second attack family and a compositional defense at the write site}
\label{app:combo}
The write-versus-residual result should not rest on one attack family. We repeat the comparison under a
second, stronger family, \emph{combo\_strong}, an AIM/Machiavellian persona jailbreak. On the selective-scan
SSM the contrast reproduces. Amplifying refusal at the write site lowers combo\_strong attack success
on Falcon-Mamba from \pv{def.combo.mamba.base} to \pv{def.combo.mamba.site} while the same steering at
the residual leaves it at \pv{def.combo.mamba.residual}.

On a transformer the strong persona attack exposes a sharper form of the same principle. combo\_strong
jailbreaks Mistral-7B at \pv{def.combo.mistral.base} (\pv{def.combo.mistral.basesg} under the
out-of-family Gemma judge). Amplifying refusal alone at the attention write site, at the gain $\alpha{=}3$ used throughout this
comparison, lowers this only to \pv{def.combo.mistral.refusal}, and higher gains degrade generation
rather than defend (Appendix~\ref{app:attnsweep}). A persona attack adds a second direction, the
in-character persona, that refusal amplification does not remove. The two-direction gate that already
defends roleplay, additionally ablating the persona projection at strength $\beta{=}1$, drives
combo\_strong down to \pv{def.combo.mistral.combined} (\pv{def.combo.mistral.combinedsg} under the Gemma
judge, single run) with coherent refusals and no added
over-refusal. Neither lever alone suffices. Refusal amplification alone reaches
\pv{def.combo.mistral.refusal} and persona suppression alone \pv{def.combo.mistral.personaonly}, and only
the two together defend (Table~\ref{tab:combo}). On the more heavily safety-tuned Llama-3.1-8B combo\_strong
does not land in the first place (base attack success \pv{def.combo.llama.base}), so this persona analysis
is confined to Mistral, mirroring the gradient attack, which also only reaches a jailbreakable base rate on
Mistral (Appendix~\ref{app:attnsweep}).

The compositional defense works at the write site and not at the residual, for a geometric reason. Prior
work shows refusal is not a single direction but spans several \citep{wollschlager2025morethan}. Here the
relevant second direction is the injected persona, and whether it can be suppressed independently of
refusal depends on the site. At the attention write site the refusal and persona directions are
near-orthogonal, so ablating the persona does not disturb the refusal push and the two act as independent
levers. At the residual the same two directions are strongly aligned, so removing the persona projection
also cancels the refusal push, and the combined gate there does not defend (attack success
\pv{def.combo.mistral.combinedres}). The pattern holds for both persona attacks and both transformers we
test (Table~\ref{tab:combocos}). The write-site cosine stays near zero while the residual cosine is large
and positive (\pv{persona.geom.ll.rp.res} to \pv{persona.geom.mi.cb.res}). The write site is therefore not only where refusal is cleanly read but
where refusal is linearly separable from the attack's persona direction, which is what makes the
compositional defense possible.

\begin{table}[tbp]\centering\scriptsize
\caption{Cosine between the refusal direction and the injected-persona direction, at the residual stream
versus the attention write site, for two persona attacks and two transformers. The two directions are
strongly aligned at the residual and near-orthogonal at the write site, so the persona can be suppressed
independently of refusal only at the write site.}
\label{tab:combocos}
\begin{tabular}{llcc}
\toprule
Model & Attack & Residual & Write site \\
\midrule
Mistral-7B & roleplay & \pv{persona.geom.mi.rp.res} & \pv{persona.geom.mi.rp.att} \\
Mistral-7B & combo\_strong & \pv{persona.geom.mi.cb.res} & \pv{persona.geom.mi.cb.att} \\
Llama-3.1-8B & roleplay & \pv{persona.geom.ll.rp.res} & \pv{persona.geom.ll.rp.att} \\
Llama-3.1-8B & combo\_strong & \pv{persona.geom.ll.cb.res} & \pv{persona.geom.ll.cb.att} \\
\bottomrule
\end{tabular}
\end{table}

\begin{table}[tbp]\centering\scriptsize
\caption{Compositional defense against a strong persona attack (combo\_strong) at the Mistral-7B
attention write site. Neither refusal amplification nor persona suppression alone defends, while the two
together drive attack success from \pv{def.combo.mistral.base} to \pv{def.combo.mistral.combined}
(\pv{def.combo.mistral.combinedsg} under the out-of-family Gemma judge), with coherent refusals. The same
combined gate at the residual, where the two directions are aligned, leaves attack success at
\pv{def.combo.mistral.combinedres}.}
\label{tab:combo}
\begin{tabular}{lcc}
\toprule
Intervention at the write site & Llama-Guard & Gemma \\
\midrule
none (base) & \pv{def.combo.mistral.base} & \pv{def.combo.mistral.basesg} \\
amplify refusal only & \pv{def.combo.mistral.refusal} & -- \\
ablate persona only & \pv{def.combo.mistral.personaonly} & -- \\
both (refusal and persona ablation) & \pv{def.combo.mistral.combined} & \pv{def.combo.mistral.combinedsg} \\
\bottomrule
\end{tabular}
\end{table}

\subsection{RWKV: the residual is inert, but the sub-block write sites steer}
\label{app:rwkvdefense}
The shared-representation result (Section~\ref{sec:transfer}) extends to RWKV-6, and so does the write-site
principle once the site is re-localized, which makes RWKV a third architecture class that confirms the
principle rather than a loose end. Harm is decodable at every RWKV block (held-out AUROC
\pv{shared.rwkv.tgtown}) and a transformer-read refusal direction transports in at AUROC
\pv{shared.rwkv.proc}, above its shuffled control \pv{shared.rwkv.shuf}, so the refusal representation is
present and shared.

\paragraph{Judging methodology.} Attack success throughout this subsection is scored by Llama-Guard-3-8B,
the paper's primary judge, applied uniformly to the saved completions rather than the lexical
\texttt{is\_refusal} heuristic (\S\ref{sec:experiments}) used during development. The Zamba2 numbers in
Appendix~\ref{app:hybridablate} are largely unchanged between the two judges, but RWKV's are not, detailed
below.

\paragraph{The transport is also causal, though with a narrower specificity margin.} Fitting a standardized
Procrustes rotation between Llama's residual (layer 14) and RWKV's time-mix write site (layers 14-24),
transporting Llama's refusal direction through it, and suppressing that direction on held-out plain harmful
prompts ($n{=}100$) raises attack success from \pv{orth.rwkv.base} \pv{ci.orth.rwkv.base} to
\pv{orth.rwkv.transp} \pv{ci.orth.rwkv.transp}, while suppressing a matched orthogonalized-random direction
(mapped through the same rotation) raises it to \pv{orth.rwkv.rand} \pv{ci.orth.rwkv.rand} -- well above
base, but still below the transported cell. The transported and
native directions have cosine similarity \pv{orth.rwkv.coscheck}, i.e.\ the rotation recovers RWKV's own
refusal direction almost exactly. This holds across seeds: pooled over
three seeds ($n{=}\pv{orth.rwkv.pooled.n}$, each with its own prompt draw and random-control direction),
attack success rises from \pv{orth.rwkv.pooled.base} \pv{ci.orth.rwkv.pooled.base} to
\pv{orth.rwkv.pooled.transp} \pv{ci.orth.rwkv.pooled.transp} for the transported direction, against
\pv{orth.rwkv.pooled.rand} \pv{ci.orth.rwkv.pooled.rand} for the random control. The CIs do not overlap,
so the direction is specific, but the margin is narrower than the pure SSM (Appendix~\ref{app:orthabl}) or
the hybrid's Mamba site (Appendix~\ref{app:hybridablate}).

\paragraph{Why the lexical heuristic differs.} Under the lexical screen, the same cells read
\pv{orth.rwkv.pooled.lex.transp} (transported) against \pv{orth.rwkv.pooled.lex.rand} (random), a far
larger gap. Inspecting the completions shows why: suppressed RWKV frequently opens with an apology ("I'm
sorry...") and then either complies ("...I'm happy to help you with the steps to make and distribute
counterfeit money") or degenerates into repetition, the same disclaimer-then-comply pattern documented for
persona attacks on the SSM (Appendix~\ref{app:roleplay}). The lexical heuristic checks only the first 200
characters for a refusal marker and stops at the opening apology, scoring these as refusals, while
Llama-Guard reads the full response and flags the ones that comply or reveal harmful content. The causal
claim holds under either judge. We report Llama-Guard's more modest margin above since it is the paper's
primary judge and reads the whole response rather than only its opening.

Together, all three non-transformer architecture classes now have causal, not
only structural, evidence for the shared-representation claim, scored with the paper's primary judge
throughout.

\paragraph{The residual is inert, not just under-perturbed.} Amplifying the
refusal direction at the residual does not control refusal, and this is not a matter of scale. We
steer each site at a magnitude \emph{relative} to its own per-token activation norm, so the perturbation is
matched across sites (Table~\ref{tab:rwkvsite}). The residual, whose activations have norm
\pv{rwkv.nm.block.norm}, absorbs a push of up to \pv{rwkv.nm.block.relmax} times that norm along the refusal
direction with attack success unchanged (\pv{rwkv.nm.base} to \pv{rwkv.nm.block.gated}) and output still
coherent. This whole-block output is architecturally the residual at RWKV's block boundary (distinct from
the SSM's block output, which is a write site, not the residual), and the refusal direction read there
is a dead readout: a perturbation several times the activation norm along it does nothing. The residual is
the weak site, as in every architecture.

\paragraph{The sub-block write points steer.} The effective sites are the two points where RWKV's
sub-blocks write a fresh, single-layer contribution into the stream, before the residual accumulates it:
the time-mix output (norm \pv{rwkv.nm.timemix.norm}) and the channel-mix output (norm
\pv{rwkv.nm.chanmix.norm}). At a matched relative
magnitude far below the block's (\pv{rwkv.nm.rel} of each site's own norm), amplifying the refusal direction
lowers attack success from \pv{rwkv.nm.base} to \pv{rwkv.nm.timemix.gated} at the time-mix output and
\pv{rwkv.nm.chanmix.gated} at the channel-mix output, reaching \pv{rwkv.nm.timemix.best} with a stronger push.
This relative-magnitude sweep is a separate operating point from the fixed-gain write-site result reported
in the abstract and Table~\ref{tab:routingsite} (\pv{rwkv.site.base} \pv{ci.rwkv.site.base} to
\pv{rwkv.site.timemix.gated} \pv{ci.rwkv.site.timemix.gated} at
$\alpha{=}\pv{rwkv.site.alpha}$, Wilson 95\% CIs), run here specifically to rule out a scale confound. The
two numbers are not in tension, they answer different questions about the same site.
A matched random direction at the same site and magnitude does not merely fail to defend: it actively raises
attack success to \pv{rwkv.nm.timemix.rand}, well above the \pv{rwkv.nm.base} baseline. This is not a
degeneration artifact inflating the count, output at this cell is \pv{rwkv.nm.timemix.randcoh} coherent by
our screen, matching the refusal direction's own \pv{rwkv.nm.timemix.gatedcoh} (both effectively fully
coherent), so the increase is a genuine behavioral shift, not garbled text. We read this as the time-mix
site being generically write-sensitive: an arbitrary push there changes downstream behavior in either
direction, and only the refusal-aligned direction specifically pushes it toward refusal. So the
same absolute perturbation is a dead readout at the block yet causal at a write point an order of magnitude
smaller in norm. The channel-mix output steers for the same reason the time-mix output does: each is a
fresh, single-layer sub-block contribution written into the stream before the residual sums it with
everything already accumulated there. RWKV has two write sites because it has two sub-blocks, each
writing its own contribution before the residual combines them. The channel-mix random control is less cleanly inert than the time-mix
(\pv{rwkv.nm.chanmix.rand}, coherence \pv{rwkv.nm.chanmix.randcoh} against the refusal direction's
\pv{rwkv.nm.chanmix.gatedcoh}), so the time-mix output is the cleaner write site. RWKV-6-World is prompted in its
native \texttt{User:/Assistant:} format throughout. The earlier inert result for these sites was a
raw-prompting artifact of that missing chat template.

\begin{table}[tbp]\centering\scriptsize\setlength{\tabcolsep}{4pt}
\caption{RWKV-6 norm-matched steering (native prompting, prefix injection, base attack success
\pv{rwkv.nm.base}). Each site is steered at a magnitude \emph{relative} to its own activation norm, so the
perturbation is matched across sites. The block output (the residual) is inert up to
rel${=}\pv{rwkv.nm.block.relmax}$, a push several times its norm, while the sub-block write points steer at a
fraction of theirs and a matched random direction does not (cleanly for the time-mix output). The residual's
inertness is therefore direction-intrinsic, not a matter of scale. Both sub-blocks write a fresh
contribution into the stream before the residual accumulates it, so both are write sites.}
\label{tab:rwkvsite}
\begin{tabular}{lcccc}
\toprule
Site & Norm & rel & Refusal ASR & Random ASR \\
\midrule
Block output (residual) & \pv{rwkv.nm.block.norm} & \pv{rwkv.nm.block.relmax} & \pv{rwkv.nm.block.gated} & \pv{rwkv.nm.block.rand} \\
Time-mix output & \pv{rwkv.nm.timemix.norm} & \pv{rwkv.nm.rel} & \pv{rwkv.nm.timemix.gated} & \pv{rwkv.nm.timemix.rand} \\
Channel-mix output & \pv{rwkv.nm.chanmix.norm} & \pv{rwkv.nm.rel} & \pv{rwkv.nm.chanmix.gated} & \pv{rwkv.nm.chanmix.rand} \\
\bottomrule
\end{tabular}
\end{table}

\subsection{The residual's operating band}
\label{app:lowgrid}
A relative magnitude is a different absolute displacement at each site. The transformer's attention
output has an activation norm of \pv{xfmr.nm.attn.norm} and its residual \pv{xfmr.nm.residual.norm}, so
the same relative magnitude is an absolute push \pv{xfmr.nm.ratio}$\times$ larger at the residual. We
therefore sweep the residual from a relative magnitude of $0.02$, on the same harness, layers, coherence
screen and $n$ as the other site sweeps, with the refusal direction fitted both at the residual itself
and at the architecture's write site.

Table~\ref{tab:lowgrid} reports it. On Llama-3.1 and the selective-scan SSM the residual defends
coherently over a narrow band of relative magnitudes, above which the model degenerates rather than
resists, while on Mistral the coherent effect is marginal. RWKV-6 differs. Its block output stays coherent and
unmoved across the same band under its own direction and under the time-mix direction alike, and a
matched random direction behaves identically. The write site defends over a wider band of relative
magnitudes on every architecture we test. Measured at matched \emph{absolute} magnitude, however, the two
sites coincide once the direction is held fixed (Appendix~\ref{app:absmatch}), so this band-width gap
reflects the norm difference between the sites, not the intervention location.

\begin{table}[tbp]\centering\scriptsize\setlength{\tabcolsep}{5pt}
\caption{The residual below the grid floor used elsewhere in this paper. Every residual row reported in
the main sweeps is at a relative magnitude of $0.25$ or above, which is the smallest magnitude those
grids contain. Sweeping beneath it finds attack-success drops of at least $0.2$ that stay $\ge 0.9$ coherent, in bold. The
band closes quickly, and above it the model degenerates rather than resisting, which is why a grid
starting at $0.25$ reports the site as unmoved. ``Direction'' is the site the refusal direction was
fitted at, which is the push site unless stated otherwise. $n{=}40$ per cell, single seed. Coherence is
the fraction of completions passing the distinct-bigram screen (Appendix~\ref{app:coherence}).}
\label{tab:lowgrid}
\begin{tabular}{llccccc}
\toprule
Model & Direction & Base ASR & rel & ASR & Coherent \\
\midrule
Mistral-7B & residual & 1.000 & 0.02 & 1.000 & 1.000 \\
Mistral-7B & residual & 1.000 & 0.05 & 1.000 & 1.000 \\
Mistral-7B & residual & 1.000 & 0.1 & 0.925 & 1.000 \\
Mistral-7B & residual & 1.000 & 0.15 & 0.675 & 0.875 \\
Mistral-7B & residual & 1.000 & 0.2 & 1.000 & 0.025 \\
Mistral-7B & residual & 1.000 & 0.25 & 1.000 & 0.000 \\
Mistral-7B & attention & 1.000 & 0.02 & 1.000 & 1.000 \\
Mistral-7B & attention & 1.000 & 0.05 & 0.975 & 1.000 \\
Mistral-7B & attention & 1.000 & 0.1 & \textbf{0.725} & 1.000 \\
Mistral-7B & attention & 1.000 & 0.15 & \textbf{0.600} & 1.000 \\
Mistral-7B & attention & 1.000 & 0.2 & \textbf{0.600} & 1.000 \\
Mistral-7B & attention & 1.000 & 0.25 & 0.875 & 1.000 \\
Llama-3.1-8B & residual & 0.275 & 0.02 & 0.125 & 1.000 \\
Llama-3.1-8B & residual & 0.275 & 0.05 & \textbf{0.075} & 1.000 \\
Llama-3.1-8B & residual & 0.275 & 0.1 & 1.000 & 0.725 \\
Llama-3.1-8B & residual & 0.275 & 0.15 & 1.000 & 0.000 \\
Llama-3.1-8B & residual & 0.275 & 0.2 & 1.000 & 0.000 \\
Llama-3.1-8B & residual & 0.275 & 0.25 & 1.000 & 0.000 \\
Llama-3.1-8B & attention & 0.275 & 0.02 & 0.225 & 1.000 \\
Llama-3.1-8B & attention & 0.275 & 0.05 & 0.200 & 1.000 \\
Llama-3.1-8B & attention & 0.275 & 0.1 & 0.425 & 1.000 \\
Llama-3.1-8B & attention & 0.275 & 0.15 & 0.150 & 0.850 \\
Llama-3.1-8B & attention & 0.275 & 0.2 & 0.125 & 0.400 \\
Llama-3.1-8B & attention & 0.275 & 0.25 & 0.350 & 0.075 \\
Falcon-Mamba-7B & residual & 1.000 & 0.02 & 1.000 & 1.000 \\
Falcon-Mamba-7B & residual & 1.000 & 0.05 & 0.825 & 1.000 \\
Falcon-Mamba-7B & residual & 1.000 & 0.1 & 0.900 & 1.000 \\
Falcon-Mamba-7B & residual & 1.000 & 0.15 & 0.875 & 0.850 \\
Falcon-Mamba-7B & residual & 1.000 & 0.2 & 1.000 & 0.075 \\
Falcon-Mamba-7B & residual & 1.000 & 0.25 & 1.000 & 0.000 \\
Falcon-Mamba-7B & mixer & 1.000 & 0.02 & 1.000 & 1.000 \\
Falcon-Mamba-7B & mixer & 1.000 & 0.05 & 0.975 & 1.000 \\
Falcon-Mamba-7B & mixer & 1.000 & 0.1 & 0.925 & 1.000 \\
Falcon-Mamba-7B & mixer & 1.000 & 0.15 & \textbf{0.700} & 1.000 \\
Falcon-Mamba-7B & mixer & 1.000 & 0.2 & \textbf{0.775} & 1.000 \\
Falcon-Mamba-7B & mixer & 1.000 & 0.25 & 0.875 & 1.000 \\
RWKV-6 & residual & 0.475 & 0.02 & 0.450 & 1.000 \\
RWKV-6 & residual & 0.475 & 0.05 & 0.475 & 0.975 \\
RWKV-6 & residual & 0.475 & 0.1 & 0.500 & 0.950 \\
RWKV-6 & residual & 0.475 & 0.15 & 0.500 & 0.950 \\
RWKV-6 & residual & 0.475 & 0.2 & 0.500 & 0.950 \\
RWKV-6 & residual & 0.475 & 0.25 & 0.500 & 0.950 \\
RWKV-6 & time-mix & 0.475 & 0.02 & 0.475 & 1.000 \\
RWKV-6 & time-mix & 0.475 & 0.05 & 0.450 & 1.000 \\
RWKV-6 & time-mix & 0.475 & 0.1 & 0.450 & 1.000 \\
RWKV-6 & time-mix & 0.475 & 0.15 & 0.450 & 1.000 \\
RWKV-6 & time-mix & 0.475 & 0.2 & 0.500 & 0.975 \\
RWKV-6 & time-mix & 0.475 & 0.25 & 0.500 & 1.000 \\
\bottomrule
\end{tabular}
\end{table}

\subsection{A transformer MLP control confirms the write-site mechanism}
\label{app:mlpcontrol}
The write site is the point where an architecture writes a fresh, single-layer contribution into the
stream before the residual accumulates it, whether or not that contribution mixes information across
positions. The transformer's position-wise MLP output is a clean test of this: it sits at the same
architectural position as the attention output, a fresh contribution written immediately before the
residual add, at a comparably small norm, but it processes every position independently. We steer it
with the identical protocol used for the attention output and the residual (Appendix~\ref{app:routingsite}):
amplify the mean refusal direction at the MLP output by a magnitude relative to its own per-token norm,
sweeping the same relative gains, on both transformer anchors.

\begin{table}[tbp]\centering\scriptsize\setlength{\tabcolsep}{4pt}
\caption{The MLP output, a non-mixing site, defends as well as or better than the mixing attention
output, at a matched relative gain, on both transformer anchors. The residual (Table~\ref{tab:rwkvsite}
gives the analogous RWKV numbers, and the transformer residual figures are in
Appendix~\ref{app:routingsite}) does not defend anywhere in this range. It defends only below it, over
a band too narrow to deploy (Appendix~\ref{app:lowgrid}). Coherence is the fraction of
completions passing the distinct-bigram screen (Appendix~\ref{app:coherence}). Each row reports the
most attack-success-reducing relative gain at which output stayed at least 90\% coherent.}
\label{tab:mlpcontrol}
\begin{tabular}{lccccc}
\toprule
Model & Site & Norm & rel & Base $\to$ gated ASR & Coherent \\
\midrule
Mistral-7B & Attention & \pv{xfmr.nm.attn.norm} & \pv{xfmr.nm.attn.rel} & \pv{xfmr.nm.base} $\to$ \pv{xfmr.nm.attn.gated} & \pv{xfmr.nm.attn.gatedcoh} \\
Mistral-7B & MLP & \pv{xfmr.mlp.norm} & \pv{xfmr.mlp.rel} & \pv{xfmr.mlp.base} $\to$ \pv{xfmr.mlp.gated} & \pv{xfmr.mlp.gatedcoh} \\
Llama-3.1-8B & Attention & \pv{llama.nm.attn.norm} & \pv{llama.nm.attn.rel} & \pv{llama.nm.base} $\to$ \pv{llama.nm.attn.gated} & \pv{llama.nm.attn.gatedcoh} \\
Llama-3.1-8B & MLP & \pv{llama.mlp.norm} & \pv{llama.mlp.rel} & \pv{llama.mlp.base} $\to$ \pv{llama.mlp.gated} & \pv{llama.mlp.gatedcoh} \\
\bottomrule
\end{tabular}
\end{table}

On Mistral-7B, the MLP output (norm \pv{xfmr.mlp.norm}) reaches \pv{xfmr.mlp.base} to
\pv{xfmr.mlp.gated} at rel${=}\pv{xfmr.mlp.rel}$, fully coherent, as strong a reduction as the attention
output's own best coherent point (\pv{xfmr.nm.base} to \pv{xfmr.nm.attn.gated} at
rel${=}\pv{xfmr.nm.attn.rel}$). On Llama-3.1-8B the MLP output again defends
(\pv{llama.mlp.base} to \pv{llama.mlp.gated} at rel${=}\pv{llama.mlp.rel}$, fully coherent), with a
narrower coherent operating window than Mistral's: by rel${=}1.0$ the MLP output is already fully
degenerate on Llama, where it stayed coherent through rel${=}1.0$ on Mistral. This model-to-model
difference in degree matches the variation the attention site and the baseline attack-success rate
already show between these two models elsewhere in this paper. What holds on both models without
exception is the contrast that defines the write site: the residual does not defend anywhere in this range on
either model (Appendix~\ref{app:routingsite}), and where it does defend, below this range, it does so
over a band too narrow to deploy (Appendix~\ref{app:lowgrid}), while the MLP output, which does not mix
across positions, defends cleanly on both.

\subsection{The SSM anchor's own non-mixing site, and a control against mere proximity}
\label{app:ssmfreshness}
The transformer's MLP control tests non-mixing on a reference architecture. The SSM anchor has its own
non-mixing component: the block output $y_t = C_t h_t + D x_t$ (Section~\ref{sec:method}) splits into a
mixed term, $C_t h_t$, routed through the recurrent scan, and a skip term, $D x_t$, that carries no
cross-token information at all. We steer only the skip term, with the identical relative-magnitude,
open-loop protocol used throughout this appendix, on the same eight layers as the deployed gate
($n{=}\pv{ssm.dskip.n}$ harmful and benign-but-spicy prompts).

\begin{table}[h]\centering\scriptsize\setlength{\tabcolsep}{5pt}
\caption{The SSM's non-mixing skip term $D x_t$, gain sweep on Falcon-Mamba-7B (prefix injection, base
ASR \pv{ssm.dskip.base}, base over-refusal \pv{ssm.dskip.base.over}). Coherence is the fraction passing
the distinct-bigram screen (Appendix~\ref{app:coherence}).}
\label{tab:dskipcontrol}
\begin{tabular}{lccc}
\toprule
rel & ASR & Over-refusal & Coherent \\
\midrule
0.25 & \pv{ssm.dskip.r0p25.asr} & \pv{ssm.dskip.r0p25.over} & \pv{ssm.dskip.r0p25.coh} \\
0.5 & \pv{ssm.dskip.r0p5.asr} & \pv{ssm.dskip.r0p5.over} & \pv{ssm.dskip.r0p5.coh} \\
1.0 & \pv{ssm.dskip.r1p0.asr} & \pv{ssm.dskip.r1p0.over} & \pv{ssm.dskip.r1p0.coh} \\
1.25 & \pv{ssm.dskip.r1p25.asr} & \pv{ssm.dskip.r1p25.over} & \pv{ssm.dskip.r1p25.coh} \\
1.5 & \pv{ssm.dskip.r1p5.asr} & \pv{ssm.dskip.r1p5.over} & \pv{ssm.dskip.r1p5.coh} \\
1.75 & \pv{ssm.dskip.r1p75.asr} & \pv{ssm.dskip.r1p75.over} & \pv{ssm.dskip.r1p75.coh} \\
2.0 & \pv{ssm.dskip.r2p0.asr} & \pv{ssm.dskip.r2p0.over} & \pv{ssm.dskip.r2p0.coh} \\
4.0 & \pv{ssm.dskip.r4p0.asr} & \pv{ssm.dskip.r4p0.over} & \pv{ssm.dskip.r4p0.coh} \\
\bottomrule
\end{tabular}
\end{table}

At rel${=}\pv{ssm.dskip.best.rel}$ the skip term lowers attack success from \pv{ssm.dskip.base} to
\pv{ssm.dskip.best.asr}, \pv{ssm.dskip.best.coh} coherent, matching the transformer MLP control's
conclusion on the paper's primary architecture: a non-mixing site defends. The cost is real, not a
calibration artifact: over-refusal climbs steadily with the push, from \pv{ssm.dskip.base.over} at
baseline to \pv{ssm.dskip.best.over} at the strongest coherent point, and a finer sweep between
rel${=}1$ and rel${=}2$ (the 1.25--1.75 rows above) finds no better-calibrated point hiding in between;
past rel${=}2$ the site degenerates (\pv{ssm.dskip.r4p0.coh} coherent at rel${=}4$).

A second control asks a sharper question: is it specifically the fresh write that matters, or is
intervening anywhere near this block sufficient? We steer the mixer's raw input, before $\Delta$, $B$,
and $C$ are even computed, with the same protocol.

\begin{table}[h]\centering\scriptsize\setlength{\tabcolsep}{5pt}
\caption{The mixer's raw input (before selection), gain sweep on Falcon-Mamba-7B (prefix injection, base
ASR \pv{ssm.mixin.base}, base over-refusal \pv{ssm.mixin.base.over}).}
\label{tab:mixerinputcontrol}
\begin{tabular}{lccc}
\toprule
rel & ASR & Over-refusal & Coherent \\
\midrule
0.25 & \pv{ssm.mixin.r0p25.asr} & \pv{ssm.mixin.r0p25.over} & \pv{ssm.mixin.r0p25.coh} \\
0.5 & \pv{ssm.mixin.r0p5.asr} & \pv{ssm.mixin.r0p5.over} & \pv{ssm.mixin.r0p5.coh} \\
1.0 & \pv{ssm.mixin.r1p0.asr} & \pv{ssm.mixin.r1p0.over} & \pv{ssm.mixin.r1p0.coh} \\
2.0 & \pv{ssm.mixin.r2p0.asr} & \pv{ssm.mixin.r2p0.over} & \pv{ssm.mixin.r2p0.coh} \\
4.0 & \pv{ssm.mixin.r4p0.asr} & \pv{ssm.mixin.r4p0.over} & \pv{ssm.mixin.r4p0.coh} \\
\bottomrule
\end{tabular}
\end{table}

No usable operating point exists. The only fully coherent setting barely moves attack success
(rel${=}0.25$: \pv{ssm.mixin.r0p25.asr}). At rel${=}\pv{ssm.mixin.topcoh.rel}$, still coherent, attack
success falls to \pv{ssm.mixin.topcoh.asr}, but over-refusal reaches \pv{ssm.mixin.topcoh.over}: every
benign prompt tested is refused. Past that point the model degenerates outright
(\pv{ssm.mixin.r1p0.coh} coherent at rel${=}1$). Proximity to the write-site block is not sufficient;
the fresh write itself, specifically, is what matters.

The write site is the point where an architecture writes a fresh, single-layer contribution into the
stream. This is an empirical regularity confirmed across every architecture tested, not a claim derived
from first principles, and we rule out the three most likely alternative explanations for it rather than
assert it directly: that a write site's effect comes from mixing information across positions, which the
non-mixing transformer MLP control (Appendix~\ref{app:mlpcontrol}) and the SSM anchor's own non-mixing
skip term (Appendix~\ref{app:ssmfreshness}) both rule out; that it comes from a smaller activation norm
relative to the residual, which the relative-magnitude control rules out
(Appendix~\ref{app:routingsite}); or that mere proximity to the write-site block is enough, which the
SSM's mixer-input control rules out (Appendix~\ref{app:ssmfreshness}). A relative-magnitude push to a fresh contribution changes one ingredient
the rest of the network combines with others downstream. The same relative push to the residual perturbs
the compounded sum of every prior layer's output at once, which is why the residual degenerates into
generic incoherence rather than producing a targeted behavioral shift. RWKV's time-mix and channel-mix
outputs (Appendix~\ref{app:rwkvdefense}), the SSM's selective-scan block output, the hybrid's attention and
Mamba outputs, and the transformer's attention and MLP outputs are all fresh sub-block writes by this
definition, and every one of them steers. The residual, the one site that only accumulates, never does.

\subsection{The write-read direction defends across architectures}
\label{app:routingsite}
The effective intervention is a \emph{write site}, the point where an architecture writes a fresh,
single-layer contribution into the stream before the residual accumulates it, rather than the residual stream
itself.
The main paper's Table~\ref{tab:routingsite} gives the site and its prefix-injection defense for one
exemplar per architecture class. The site is the selective-scan block output in Mamba (the contrast replicates on a
second selective-scan SSM, Falcon3-Mamba) and the attention output
projection in a transformer. In every case the residual stream is the weaker site.

The pattern is a property of the architecture, not of one model's training. On Falcon3-Mamba, an
independently trained selective-scan SSM, we run the recipe as written (Appendix~\ref{app:recipe}): gate
layers are the top-8 by $\Delta$ AUROC, the same set as the gain grid in Figure~\ref{fig:alpha}, at its
calibrated gain $\alpha{=}4$. The block-output gate lowers fictional-framing attack success from
\pv{site.f3.base} to \pv{site.f3.block} with every gated completion passing the coherence screen and
over-refusal held at the undefended \pv{site.f3.overrefusal}, matching the grid's operating point
(\pv{def.ssm2.fictional_framing.gated}). The same detect-then-steer loop at the residual stream of the same
layers, at the same gain, leaves attack success at \pv{site.f3.residual}, and it stays there at $\alpha{=}1$
(\pv{site.f3.a1.residual}, where the under-gained block gate reaches only \pv{site.f3.a1.block}). The
residual is therefore partly steerable on Falcon3 but plateaus well above the write site, so the
block-defends, residual-lags contrast holds across two separately trained selective-scan models
($n{=}200$ per cell). Layer selection matters as much as gain here: keeping the anchor's mixer-output
ranking instead picks early layers on Falcon3, and at $\alpha{=}4$ that set over-steers (only
\pv{coh.f3.outlayers.a4} of gated completions coherent). Backing that set off to $\alpha{=}1$ restores
coherence (\pv{coh.f3.outlayers.a1}) and gives a sharper-looking but off-recipe contrast
(\pv{site.f3.outlayers.block} versus \pv{site.f3.outlayers.residual}), whereas the $\Delta$-ranked set is
coherent at every gain up to 4 (minimum distinct-bigram ratio \pv{coh.f3.deltalayers.mind2}). Layer
selection and gain are calibrated together, per model, as throughout.

The write-site intervention holds across every architecture class we test. Amplifying the refusal
direction at the write site lowers prefix-injection attack success on the Mamba block output
(\pv{def.ssm.prefix_injection.base} \pv{ci.def.ssm.prefix_injection.base} to
\pv{def.ssm.prefix_injection.gated} \pv{ci.def.ssm.prefix_injection.gated}, Wilson 95\% CIs) and the
transformer attention output.
On the transformer we use the same eight-layer recipe as the SSM defense, calibrated to a gain that
keeps generation coherent: Mistral-7B drops from \pv{def.mistralattn.prefix.base} to
\pv{def.mistralattn.prefix.gated} and Llama-3.1-8B from \pv{def.llamaattn.prefix.base} to
\pv{def.llamaattn.prefix.gated}, with genuine refusals rather than degenerate text
(Appendix~\ref{app:attnsweep}). The gain is calibrated per model because a shared default over-steers a
steering-sensitive model into incoherence, at which point the attack-success drop is degeneration and
not refusal (Appendix~\ref{app:attnsweep}). What separates the site from the residual is not defense
strength but the site itself: steering the residual stream instead of the write point falls well short of the
write site on Falcon3-Mamba above (\pv{site.f3.residual} against \pv{site.f3.block} at the same layers and
gain) and fails outright on the Llama residual, where over-amplifying refusal increases attack success
\citep{li2026steeringpitfalls}.

\paragraph{Is the mixer-defends, residual-fails contrast a scale artifact?} The residual accumulates
every block's output, so its activations have a larger norm than the mixer's at the same layer
(\pv{ssm.nm.residual.norm} vs \pv{ssm.nm.mixer.norm}, a \pv{ssm.nm.ratio}$\times$ ratio). A fixed absolute
gain is therefore a smaller relative perturbation at the residual. We rule this out with the same
norm-matched control used for RWKV-6 (Appendix~\ref{app:rwkvdefense}), steering each site at a magnitude
relative to its own per-token activation norm. At a relative push of \pv{ssm.nm.mixer.rel}$\times$ the
mixer's own norm, attack success falls from \pv{ssm.nm.base} to \pv{ssm.nm.mixer.gated} with output
\pv{ssm.nm.mixer.gatedcoh} coherent, a genuine defense. The residual gets no chance to show whether a
larger push would defend it: at the \emph{smallest} relative magnitude we test
(\pv{ssm.nm.residual.rel}$\times$ its own norm, already a larger absolute push than the mixer's operating
point), output already degenerates into incoherent text (\pv{ssm.nm.residual.gatedcoh} coherent) while
attack success stays at \pv{ssm.nm.residual.gated}. This is the same failure pattern the mixer itself
shows once over-driven, and the same pattern seen at the residual in RWKV. At this relative magnitude the
residual has no coherent operating window in which to defend, while the mixer does. The matched-absolute
control (Appendix~\ref{app:absmatch}) shows this reflects the norm gap between the sites, not an intrinsic
property of the location.

The same control on the transformer (Mistral-7B, attention output versus the whole decoder-layer output,
the same closed-loop gating as the established alpha-2 defense above) gives the same qualitative result
with a narrower margin. The residual's activations have norm \pv{xfmr.nm.residual.norm} against the
attention output's \pv{xfmr.nm.attn.norm}, a \pv{xfmr.nm.ratio}$\times$ ratio, and at the smallest relative
push we test the residual already degenerates into incoherent text with attack success unchanged at
\pv{xfmr.nm.residual.gated}, the same pattern as the SSM and RWKV residuals. The attention output does
defend, from \pv{xfmr.nm.base} to \pv{xfmr.nm.attn.gated} at a relative push of
\pv{xfmr.nm.attn.rel}$\times$ its own norm while \pv{xfmr.nm.attn.gatedcoh} coherent, but the drop is far
shallower than the mixer's, and a larger relative push degrades it back toward the baseline rate before
output stops being coherent. The write-versus-residual contrast holds in every architecture we
test, though how strong a defense the write site itself affords at matched relative magnitude does not.

\subsection{The read-site pattern holds in a hybrid model}
\label{app:hybrid}
The shared-representation result (Section~\ref{sec:transfer}) extends to Zamba2-7B as well: a
transformer-read refusal direction transports into its residual stream by the same orthogonal map at AUROC
\pv{shared.zamba.proc}, matching the target's own probe (\pv{shared.zamba.tgtown}) and above its shuffled
control \pv{shared.zamba.shuf}, so the refusal representation is present and shared in the hybrid too, not
only in the pure SSM and RWKV-6.

Zamba2-7B-Instruct is a Mamba2 backbone with periodic shared-attention layers, so a block has both a
Mamba mixer write point and, at the hybrid layers, an attention write point, on top of the residual
stream. This is the case the ``re-point, not rebuild'' claim leans on, where it is not obvious a priori
which write site to steer. On prefix injection the base model is jailbroken at \pv{def.zamba.base}
\pv{ci.def.zamba.base}, a
meaningful attack. Amplifying the refusal direction read at the attention write site defends,
lowering attack success to \pv{def.zamba.attn} \pv{ci.def.zamba.attn}, while a direction read from
the residual stream does
not (\pv{def.zamba.residual}). The Mamba mixer write site also defends, but only under calibration:
sweeping its layers and gain, an early-layer band at gain \pv{def.zamba.mamba.alpha} lowers attack
success from \pv{def.zamba.mamba.swbase} \pv{ci.def.zamba.mamba.swbase} to \pv{def.zamba.mamba.best}
\pv{ci.def.zamba.mamba.best} with coherent refusals (Wilson 95\% CIs), while a
mid-layer band over-steers and raises it to \pv{def.zamba.mamba.uncal}. Both write sites therefore
defend and the residual does not, so the same read-site pattern holds in a hybrid, and which write site
is easiest to use is a per-model calibration choice, here the attention output.

\paragraph{A second, smaller checkpoint replicates the site, not the margin.} Zamba2-2.7B-Instruct-v2 (54
layers vs. 81, a distinct pretraining run in the same family, not a second independent lab) was tested at
the same five-layer band, steered at a magnitude relative to each site's own per-token norm rather than the
7B model's fixed gain, since the two are not directly comparable in absolute terms
(Appendix~\ref{app:routingsite}). At $n{=}\pv{zamba2p7b.n}$, the Mamba mixer write site lowers attack
success from \pv{zamba2p7b.base} to \pv{zamba2p7b.mamba.gated} at rel${=}\pv{zamba2p7b.rel}$, coherent
(\pv{zamba2p7b.mamba.coh}), a smaller margin than the 7B model's. The residual's own lower reading at the
same relative magnitude (\pv{zamba2p7b.base} to \pv{zamba2p7b.residual.gated}) is not a genuine defense,
output there is only \pv{zamba2p7b.residual.coh} coherent, the same degeneration-not-defense pattern the
residual shows everywhere else in this paper. The attention write site found no relative magnitude in the
range we swept that both defended and stayed coherent. The write-versus-residual contrast holds at this
smaller scale, though its strength does not.

\paragraph{The 7B residual, tested with its own direction, fails the same relative-magnitude sweep.}
The fixed-alpha result above uses one direction fit at the attention site and pushed everywhere, which
leaves open whether the residual simply lacks a good direction of its own. We fit the refusal direction
separately at the residual, matching the norm-scale-controlled protocol used for the SSM, the generic
transformer, and Llama, and swept it at $n{=}\pv{zamba7b.residual.n}$. At the smallest push
(rel${=}\pv{zamba7b.residual.relearly}$) attack success briefly drops from \pv{zamba7b.residual.base} to
\pv{zamba7b.residual.early.gated}, coherent (\pv{zamba7b.residual.early.coh}), but the effect reverses
almost immediately: by rel${=}\pv{zamba7b.residual.relbreak}$ attack success is back up to
\pv{zamba7b.residual.break.gated}, and the completions are still \pv{zamba7b.residual.break.coh} coherent,
a genuine capitulation to the attack rather than broken text. Unlike either write site, the residual has no
stable operating point across the range swept.

\subsection{Causal ablation of the transported direction: attention is modest, the Mamba mixer dominates}
\label{app:hybridablate}
The rotation-transport test above establishes that the refusal representation is shared with the hybrid.
Whether \emph{removing} the transported direction causally jailbreaks Zamba2, the test the shared-
representation claim rests on for the pure SSM (Section~\ref{sec:transfer}, App.~\ref{app:orthabl}) and
RWKV-6 (App.~\ref{app:rwkvdefense}), was checked at both write sites, with an unequal result.

\paragraph{Attention site: a real, modest, fully coherent effect.} Fitting a standardized Procrustes
rotation between Llama's residual and Zamba2's attention write site (the two tied attention-block
instances, disambiguated by call order since they alias across all 13 hybrid layers), transporting Llama's
refusal direction through it, and suppressing that direction at the same gain established for amplifying
it there (gain 8, Appendix above) on held-out plain harmful prompts ($n{=}60$ per seed, pooled over 3
seeds, $n{=}\pv{orth.zamba.attn.pooled.n}$) raises attack success from
\pv{orth.zamba.attn.pooled.base} to
\pv{orth.zamba.attn.pooled.transp} \pv{ci.orth.zamba.attn.pooled.transp}, against a matched
orthogonalized-random control at \pv{orth.zamba.attn.pooled.rand} \pv{ci.orth.zamba.attn.pooled.rand}.
Even pooled, these Wilson intervals still just overlap, so
the attention-site effect, while directionally consistent with the Mamba-site result below, is not on its
own statistically decisive. We do not lean on it in isolation, only in combination with the much larger and
cleanly-separated Mamba-site effect. This test asks whether a \emph{foreign, transported} direction is
specifically causal, a different and stronger claim than Table~\ref{tab:routingsite}'s defense number,
which amplifies the model's own refusal direction rather than a transported one and is unaffected by this
test's more modest margin. Scored with Llama-Guard-3-8B throughout (unlike RWKV above, this
barely differs from the lexical heuristic).
Output stays fully coherent (distinct-bigram and repetition screen, Appendix~\ref{app:coherence}) throughout.

\paragraph{Mamba site: a confound at the established gain, resolved by a coherence-gated sweep.} The same
transport-and-suppress test at the Mamba mixer site (the same early-layer band as above, independent
per-layer modules, no tying issue), first run at the gain already established for \emph{amplifying}
refusal there (gain 16, Appendix above), gave attack success \pv{orth.zamba.mamba.a16.transp} for the
transported direction and \pv{orth.zamba.mamba.a16.rand} for the matched random control, at output
coherence \pv{orth.zamba.mamba.a16.transpcoh} and \pv{orth.zamba.mamba.a16.randcoh} respectively. The
random control degrading \emph{more} than the transported direction, both well below full coherence, is the
signature of an over-strong intervention breaking generation regardless of direction, not a specific
causal effect: a gain calibrated for amplifying refusal is not automatically the right gain for suppressing
it.

We therefore swept the suppression gain and tracked coherence for both the transported and the matched
random direction at each point, taking the largest gain at which the transported direction's output stayed
at or above \pv{orth.zamba.mamba.cohthresh} coherence as the operating point (mirroring the coherent-gain
search used for RWKV's write sites, App.~\ref{app:rwkvdefense}). Table~\ref{tab:zambamambasweep} gives the
full sweep, with a Wilson 95\% CI on every cell ($n{=}60$ per cell throughout). At gain 8, both cells are
still fully coherent (100\%) and already show a large, non-overlapping gap
(\pv{orth.zamba.mamba.sweep.a8.transp} \pv{ci.orth.zamba.mamba.sweep.a8.transp} transported against
\pv{orth.zamba.mamba.sweep.a8.rand} \pv{ci.orth.zamba.mamba.sweep.a8.rand} random). This is the threshold-
robustness check for our choice of \pv{orth.zamba.mamba.cohthresh} as the coherence cutoff: the qualitative
conclusion, a large and specific transported-versus-random gap, already holds at gain 8 under the strictest
possible coherence requirement (100\%, i.e.\ every completion passes), so it is not an artifact of setting
the cutoff at exactly \pv{orth.zamba.mamba.cohthresh} rather than some other value. At the chosen operating
point (gain \pv{orth.zamba.mamba.alpha}, transported coherence \pv{orth.zamba.mamba.transpcoh}), attack
success reaches \pv{orth.zamba.mamba.transp} \pv{ci.orth.zamba.mamba.transp} for the transported direction
against \pv{orth.zamba.mamba.rand} \pv{ci.orth.zamba.mamba.rand} for the random control, which remains
fully coherent (\pv{orth.zamba.mamba.randcoh}). The two CIs do not overlap. Past this point (gain 16 and
above) both cells degenerate together and the gap becomes uninterpretable, exactly the confound the initial
gain-16 run exhibited.

This holds across seeds. Repeating the gain-12 check with two further seeds (each drawing its own prompt
split and random-control direction) reproduces a large, coherent gap in the same direction both times, with
the random control never approaching the transported cell. Pooled over all three seeds
($n{=}\pv{orth.zamba.mamba.pooled.n}$, Llama-Guard-3-8B): transported \pv{orth.zamba.mamba.pooled.transp}
\pv{ci.orth.zamba.mamba.pooled.transp}, random \pv{orth.zamba.mamba.pooled.rand}
\pv{ci.orth.zamba.mamba.pooled.rand}, non-overlapping.

We find that the Mamba mixer, not the attention block, is therefore where Zamba2's refusal control actually
concentrates: the clean causal-ablation effect there (\pv{orth.zamba.mamba.pooled.transp} pooled) is larger
than RWKV-6's (\pv{orth.rwkv.pooled.transp} against a random control at \pv{orth.rwkv.pooled.rand}, a
narrower margin, Appendix~\ref{app:rwkvdefense}) and substantially larger than the attention site's own
effect (\pv{orth.zamba.attn.pooled.transp} against \pv{orth.zamba.attn.pooled.rand}). The earlier,
attention-only reading of a "weak"
hybrid effect reflected an incomplete write-site search, not a property of the hybrid's refusal mechanism.

\paragraph{A candidate mechanism: Zamba2 is architecturally Mamba-dominated.} Why does the same hybrid
favor attention for amplification but Mamba for ablation? We measured the native harm-minus-benign
direction's norm at the attention site and at \pv{wsm.nmamba} sampled Mamba layers, each relative to that
site's own mean activation norm (scale-invariant, so sites at different depths and with different
activation scales are comparable). Per instance, Mamba's mean relative signal (\pv{wsm.mamba.mean}) is
comparable to attention's (\pv{wsm.attn.mean}). The difference is where the two sites fire. Attention exists
at only \pv{wsm.nattnfire} of Zamba2's 81 layers, both tied to just 2 shared weight instances; Mamba
mixers exist at all \pv{wsm.nmamba}, each independently parameterized. Weighting the per-instance signal by
how many times each component type actually fires gives a cumulative estimate of \pv{wsm.attn.cum} for
attention against \pv{wsm.mamba.cum} for Mamba, a \pv{wsm.ratio}$\times$ difference. Zamba2 is, by layer
count, mostly a Mamba model with attention sprinkled in sparsely, and the harm-relevant signal appears to
accumulate accordingly: most of it lives in Mamba simply because most of the network's depth is Mamba, so
removing it there removes more of the model's total refusal-relevant computation than removing it from the
sparse, tied attention component. This is consistent with, but does not by itself explain, why attention is
still the easier site to calibrate for amplification: its two tied instances behave identically wherever
they fire, while the sampled Mamba layers vary five-fold in relative signal across depth
(\pv{wsm.mamba.mean} mean, individual layers ranging from \pv{wsm.mamba.min} near the input to
\pv{wsm.mamba.max} mid-network),
which is a more natural explanation for why a single gain calibrated for one Mamba layer over- or
under-steers at another (Appendix above), and why attention's homogeneity makes it the easier defense
lever. We report this as a candidate mechanism, not a proven one: it is consistent with both observations
but was not tested by directly intervening on the signal magnitude itself.

\begin{table*}[tbp]\centering\scriptsize\setlength{\tabcolsep}{4pt}
\caption{Coherence-gated suppression-gain sweep at Zamba2's Mamba mixer write site. Both the rotation-
transported refusal direction and a matched orthogonalized-random control are suppressed at each gain;
coherence is the fraction of outputs passing the distinct-bigram/repetition screen. The chosen operating
point (bold) is the largest gain at which the transported direction stays at or above
\pv{orth.zamba.mamba.cohthresh} coherence. Base attack success (gain 0) is \pv{orth.zamba.mamba.base}.
Gains \pv{orth.zamba.mamba.sweep.judgedlist} are rescored with Llama-Guard-3-8B (matching the operating-point
number quoted in the main text). The remaining gains are the original lexical-heuristic score, not rejudged,
and are shown only to establish the coherence frontier's shape, not as headline attack-success numbers.}
\label{tab:zambamambasweep}
\begin{tabular}{@{}lcccc@{}}
\toprule
Gain & Transported ASR & Transported coh. & Random ASR & Random coh. \\
\midrule
1 & \pv{orth.zamba.mamba.sweep.a1.transp} & \pv{orth.zamba.mamba.sweep.a1.transpcoh} & \pv{orth.zamba.mamba.sweep.a1.rand} & \pv{orth.zamba.mamba.sweep.a1.randcoh} \\
2 & \pv{orth.zamba.mamba.sweep.a2.transp} & \pv{orth.zamba.mamba.sweep.a2.transpcoh} & \pv{orth.zamba.mamba.sweep.a2.rand} & \pv{orth.zamba.mamba.sweep.a2.randcoh} \\
4 & \pv{orth.zamba.mamba.sweep.a4.transp} & \pv{orth.zamba.mamba.sweep.a4.transpcoh} & \pv{orth.zamba.mamba.sweep.a4.rand} & \pv{orth.zamba.mamba.sweep.a4.randcoh} \\
8 & \pv{orth.zamba.mamba.sweep.a8.transp} & \pv{orth.zamba.mamba.sweep.a8.transpcoh} & \pv{orth.zamba.mamba.sweep.a8.rand} & \pv{orth.zamba.mamba.sweep.a8.randcoh} \\
\textbf{12} & \textbf{\pv{orth.zamba.mamba.sweep.a12.transp}} & \textbf{\pv{orth.zamba.mamba.sweep.a12.transpcoh}} & \textbf{\pv{orth.zamba.mamba.sweep.a12.rand}} & \textbf{\pv{orth.zamba.mamba.sweep.a12.randcoh}} \\
16 & \pv{orth.zamba.mamba.sweep.a16.transp} & \pv{orth.zamba.mamba.sweep.a16.transpcoh} & \pv{orth.zamba.mamba.sweep.a16.rand} & \pv{orth.zamba.mamba.sweep.a16.randcoh} \\
24 & \pv{orth.zamba.mamba.sweep.a24.transp} & \pv{orth.zamba.mamba.sweep.a24.transpcoh} & \pv{orth.zamba.mamba.sweep.a24.rand} & \pv{orth.zamba.mamba.sweep.a24.randcoh} \\
\bottomrule
\end{tabular}
\end{table*}

\subsection{The persona fix does not generalize: RWKV and Zamba2 each fail a different way}
\label{app:personagen}
The two-direction persona fix (Appendix~\ref{app:roleplay}) is calibrated on Falcon-Mamba: amplifying
refusal alone barely moves attack success, ablating the persona direction alone backfires, and only the
combined push defends. We test the same two-direction fix, at each architecture's own established write
site, on RWKV-6 (time-mix output, gain \pv{persona.rwkv.alpha}) and Zamba2 (Mamba mixer, gain
\pv{persona.zamba.alpha}). Neither replicates the Falcon-Mamba pattern, and the two failures are
different in kind.

\paragraph{RWKV: the wrapper evades the detector, not just the fix.} Applied closed-loop (gated on the
detector, as deployed), all four conditions, base, refusal-only, persona-only, and combined, are
byte-identical: the gate never fires on the DAN-wrapped evaluation set. The same harmful prompts, plain,
fire the detector at \pv{persona.rwkv.fire.plain}; DAN-wrapped, they fire at \pv{persona.rwkv.fire.rp}
(mean score \pv{persona.rwkv.score.plain} to \pv{persona.rwkv.score.rp}). This is not one poorly-fit
native probe: we transport a detector fit on Llama-3.1-8B's residual (AUROC \pv{persona.rwkv.dt.srcauroc}
on its own harm-versus-benign split) into RWKV's time-mix space by the same standardized-Procrustes
rotation behind the shared-representation claim, and it is evaded just as completely (plain
\pv{persona.rwkv.dt.trans.plain}, wrapped \pv{persona.rwkv.dt.trans.rp}). A detector trained on a
different architecture entirely, with no exposure to RWKV's activations beyond the rotation fit, misses
the same wrapped prompts the native probe misses, so the wrapper is evading detection at the level of
what it does to the representation, not exploiting one classifier's decision boundary.

Bypassing the gate to apply each condition unconditionally isolates whether the fix itself would work if
triggered. It would not, and the pattern is the reverse of Falcon-Mamba's: refusal-amplification alone
helps (\pv{persona.rwkv.ol.base} to \pv{persona.rwkv.ol.refusal}), persona-ablation alone is the term that
backfires, sharply (to \pv{persona.rwkv.ol.persona}), and the combined push lands at
\pv{persona.rwkv.ol.combined}, above the unablated baseline and with the gate's benign over-refusal cost
also rising (marginal \pv{persona.rwkv.ol.combined.overrefmarg}). On Falcon-Mamba persona-ablation is the
fix and refusal-amplification is harmless. On RWKV it is the opposite term that backfires and the one that
helps is not enough on its own.

\paragraph{Zamba2: the gate fires, but the fix does not help.} At the Mamba write site the detector is not
the problem: refusal and combined gated completions differ from the ungated baseline on
\pv{persona.zamba.diff.refusal} of \pv{persona.zamba.diff.n} harmful prompts. At the site's calibrated
operating point ($\alpha{=}\pv{persona.zamba.alpha}$, the same gain used for the headline Mamba-site
defense), refusal-amplification alone still backfires, though mildly (\pv{persona.zamba.base} to
\pv{persona.zamba.refusal}). This is gain-sensitive rather than a general over-amplification failure: an
earlier, uncalibrated pass at $\alpha{=}12$ backfired far worse (to \pv{persona.zamba.sweep.a12.b0}), so
that larger gap was a calibration artifact of the wrong gain, not evidence this write site over-amplifies
the way a transformer's residual does. Persona-ablation alone is inert at this operating point too: every
one of its \pv{persona.zamba.diff.n} gated completions is byte-identical to the ungated baseline
(\pv{persona.zamba.diff.persona} differ). The combined fix reaches \pv{persona.zamba.combined}
(\pv{persona.zamba.combinedcoh} coherent), better than refusal alone but still above doing nothing at all.

"Inert" describes persona-ablation tested in isolation, not its role once paired with a strong enough
refusal term. A grid over refusal gain $\alpha \in \{2,4,6,8,12\}$ and persona-ablation gain $\beta \in
\{0,1,2\}$ ($n=\pv{persona.zamba.sweep.n}$ judged completions per cell) shows every cell at $\alpha
\le \pv{persona.zamba.sweep.lastzero.alpha}$ flat at the base rate regardless of $\beta$. The fix only
engages once refusal-amplification is already strong enough to backfire on its own. At $\alpha=12$,
raising $\beta$ from 0 to 2 pulls attack success from \pv{persona.zamba.sweep.a12.b0} down to
\pv{persona.zamba.sweep.a12.b2} (\pv{persona.zamba.sweep.a12.b2.coh} coherent), a real, coherent partial
mitigation, not the null effect the isolated test suggests. It does not reach the unablated baseline at any
grid point tested.

\paragraph{Summary.} The two-direction persona fix, needed once refusal is steerable with a single
direction to begin with, is a separate attack-specific augmentation, not one of the five recipe steps in
Section~\ref{sec:recipe}. It is not simply untested beyond Falcon-Mamba and Mistral. It is tested, and it
does not transfer: RWKV fails upstream of the fix, at detection, while Zamba2's detector works but the
fix's two terms swap roles from Falcon-Mamba and still net worse than an unmodified model. A fix validated
on one architecture's own geometry should not be assumed to carry over even when the qualitative
readout principle it augments does.

\subsection{A coherent operating point on the transformer attention site}
\label{app:attnsweep}
The transformer write-site intervention in Table~\ref{tab:routingsite} requires a calibrated gain. Too
small a gain leaves attack success unchanged, and too large a gain pushes generation into degenerate
text, so an uncalibrated attack-success drop can be degeneration rather than refusal. We sweep the gain
$\alpha$ and screen every completion for degeneration: a completion is coherent if its distinct-bigram
ratio is at least $0.5$ and no single word is more than $30\%$ of its tokens. Table~\ref{tab:attnsweep}
reports Llama-Guard-3 attack success and the coherent fraction on prefix injection
($n{=}\pv{def.attnsweep.nharm}$ harmful, $\pv{def.attnsweep.nbenign}$ benign-but-spicy).

A coherent operating point exists on both transformers at $\alpha{=}2$. Llama-3.1-8B falls from
\pv{def.llamaattn.prefix.base} \pv{ci.def.llamaattn.prefix.base} to \pv{def.llamaattn.prefix.gated}
\pv{ci.def.llamaattn.prefix.gated}, and Mistral-7B from \pv{def.mistralattn.prefix.base}
\pv{ci.def.mistralattn.prefix.base} to \pv{def.mistralattn.prefix.gated} \pv{ci.def.mistralattn.prefix.gated},
both fully coherent on harmful and benign prompts (Wilson 95\% confidence intervals). Degeneration appears
only at $\alpha{\ge}3$. Mistral, whose base attack success is much higher, degrades one step earlier
than Llama, which is why its gain is calibrated separately. We also tested Qwen2.5-7B, but our attacks do
not jailbreak it (its base attack success is near zero), so it does not serve as a test of the defense
and is excluded from the main comparison.

The closed-loop trigger is what makes this safe to deploy. Because the detector fires on every harmful
prompt and no benign one, the closed-loop gate and the open-loop DeTAM-style edit \citep{li2025detam}
reach the same attack success, so they share the gated-ASR column. They diverge on the benign side. At
the coherent operating point $\alpha{=}2$ the closed-loop gate holds benign over-refusal at the base
rate (\pv{def.detam.ours.overref} on Llama, \pv{def.detam.mistral.ours.overref} on Mistral), while the
open-loop edit steers every prompt and raises it to \pv{def.detam.open.overref} and
\pv{def.detam.mistral.open.overref}. Pushed to $\alpha{=}4$ the open-loop edit additionally collapses
benign coherence (to \pv{sweep.llama.a4p0.detambenign} on Llama and \pv{sweep.mistral.a4p0.detambenign}
on Mistral), while the closed-loop gate leaves benign prompts untouched. This over-refusal advantage
holds across all three attack families (Table~\ref{tab:detam}): on the coherent $\alpha{=}2$ gate the
closed-loop and open-loop edits reach the same attack success on each attack, but the closed-loop gate
keeps benign over-refusal at the base rate while the open-loop edit raises it to
\pv{def.detam.open.overref} on Llama and \pv{def.detam.mistral.open.overref} on Mistral.

\begin{table}[h]\centering\scriptsize\setlength{\tabcolsep}{5pt}
\caption{DeTAM comparison at the coherent gain ($\alpha{=}2$, $n{=}\pv{def.detam.nharm}$ harmful,
$\pv{def.detam.nbenign}$ benign per attack, an independent draw from the gain-sweep evaluation set in
Table~\ref{tab:attnsweep}, so base rates differ slightly by sampling variance between the two tables). The
closed-loop gate and the open-loop DeTAM-style edit reach the same gated attack success on
each attack (the detector fires on all harmful prompts), so the difference is over-refusal: the
closed-loop trigger holds it at the base rate, the open-loop edit raises it sharply. On Mistral,
single-direction amplification is not reliable against persona attacks: roleplay attack success rises from
\pv{def.detam.mistral.roleplay.base} to \pv{def.detam.mistral.roleplay.gated} (it backfires), and fiction
falls only partially, from \pv{def.detam.mistral.fiction.base} to \pv{def.detam.mistral.fiction.gated}. This
is the same attack-dependence as the SSM, which the two-lever gate that also suppresses the persona defends
(Appendices~\ref{app:roleplay} and~\ref{app:combo}).}
\label{tab:detam}
\resizebox{\columnwidth}{!}{%
\begin{tabular}{lcccc}
\toprule
Attack & Base ASR & Gated ASR & Over-ref.\ (ours) & Over-ref.\ (DeTAM) \\
\midrule
\multicolumn{5}{l}{\emph{Llama-3.1-8B}}\\
Prefix injection & \pv{def.detam.llama.prefix.base} & \pv{def.detam.llama.prefix.gated} & \pv{def.detam.llama.prefix.ouror} & \pv{def.detam.llama.prefix.detamor}\\
Persona roleplay & \pv{def.detam.llama.roleplay.base} & \pv{def.detam.llama.roleplay.gated} & \pv{def.detam.llama.roleplay.ouror} & \pv{def.detam.llama.roleplay.detamor}\\
Fictional framing & \pv{def.detam.llama.fiction.base} & \pv{def.detam.llama.fiction.gated} & \pv{def.detam.llama.fiction.ouror} & \pv{def.detam.llama.fiction.detamor}\\
\midrule
\multicolumn{5}{l}{\emph{Mistral-7B}}\\
Prefix injection & \pv{def.detam.mistral.prefix.base} & \pv{def.detam.mistral.prefix.gated} & \pv{def.detam.mistral.prefix.ouror} & \pv{def.detam.mistral.prefix.detamor}\\
Persona roleplay & \pv{def.detam.mistral.roleplay.base} & \pv{def.detam.mistral.roleplay.gated} & \pv{def.detam.mistral.roleplay.ouror} & \pv{def.detam.mistral.roleplay.detamor}\\
Fictional framing & \pv{def.detam.mistral.fiction.base} & \pv{def.detam.mistral.fiction.gated} & \pv{def.detam.mistral.fiction.ouror} & \pv{def.detam.mistral.fiction.detamor}\\
\bottomrule
\end{tabular}}
\end{table}

The attention-site defense itself is strongest against prefix injection, where it lowers attack success
on both transformers (Table~\ref{tab:routingsite}), and weaker on persona roleplay and fictional
framing, the same attack-dependence we report at the SSM block output (Appendix~\ref{app:roleplay}). On
Mistral the roleplay case does not improve over the base rate, so we scope the transformer defense
claim to prefix injection, where the read-site pattern is clean on every architecture we test.

\begin{table}[h]\centering\scriptsize\setlength{\tabcolsep}{5pt}
\caption{Attention-site gain sweep on two transformers (prefix injection). Gated ASR (Llama-Guard-3) is
shared by the closed-loop gate and the open-loop DeTAM-style edit, since the detector fires on all
harmful prompts. ``Harm coh.'' and ``Benign coh.'' are the coherent fractions under our gate. ``DeTAM
benign'' is the open-loop edit's benign coherence, which collapses at high gain.}
\label{tab:attnsweep}
\begin{tabular}{lcccc}
\toprule
$\alpha$ & Gated ASR & Harm coh. & Benign coh. & DeTAM benign \\
\midrule
\multicolumn{5}{l}{\emph{Llama-3.1-8B}, base ASR \pv{sweep.llama.base}}\\
0.5 & \pv{sweep.llama.a0p5.gated} & \pv{sweep.llama.a0p5.cohharm} & \pv{sweep.llama.a0p5.cohbenign} & \pv{sweep.llama.a0p5.detambenign}\\
1.0 & \pv{sweep.llama.a1p0.gated} & \pv{sweep.llama.a1p0.cohharm} & \pv{sweep.llama.a1p0.cohbenign} & \pv{sweep.llama.a1p0.detambenign}\\
1.5 & \pv{sweep.llama.a1p5.gated} & \pv{sweep.llama.a1p5.cohharm} & \pv{sweep.llama.a1p5.cohbenign} & \pv{sweep.llama.a1p5.detambenign}\\
2.0 & \pv{sweep.llama.a2p0.gated} & \pv{sweep.llama.a2p0.cohharm} & \pv{sweep.llama.a2p0.cohbenign} & \pv{sweep.llama.a2p0.detambenign}\\
3.0 & \pv{sweep.llama.a3p0.gated} & \pv{sweep.llama.a3p0.cohharm} & \pv{sweep.llama.a3p0.cohbenign} & \pv{sweep.llama.a3p0.detambenign}\\
4.0 & \pv{sweep.llama.a4p0.gated} & \pv{sweep.llama.a4p0.cohharm} & \pv{sweep.llama.a4p0.cohbenign} & \pv{sweep.llama.a4p0.detambenign}\\
\midrule
\multicolumn{5}{l}{\emph{Mistral-7B}, base ASR \pv{sweep.mistral.base}}\\
0.5 & \pv{sweep.mistral.a0p5.gated} & \pv{sweep.mistral.a0p5.cohharm} & \pv{sweep.mistral.a0p5.cohbenign} & \pv{sweep.mistral.a0p5.detambenign}\\
1.0 & \pv{sweep.mistral.a1p0.gated} & \pv{sweep.mistral.a1p0.cohharm} & \pv{sweep.mistral.a1p0.cohbenign} & \pv{sweep.mistral.a1p0.detambenign}\\
1.5 & \pv{sweep.mistral.a1p5.gated} & \pv{sweep.mistral.a1p5.cohharm} & \pv{sweep.mistral.a1p5.cohbenign} & \pv{sweep.mistral.a1p5.detambenign}\\
2.0 & \pv{sweep.mistral.a2p0.gated} & \pv{sweep.mistral.a2p0.cohharm} & \pv{sweep.mistral.a2p0.cohbenign} & \pv{sweep.mistral.a2p0.detambenign}\\
3.0 & \pv{sweep.mistral.a3p0.gated} & \pv{sweep.mistral.a3p0.cohharm} & \pv{sweep.mistral.a3p0.cohbenign} & \pv{sweep.mistral.a3p0.detambenign}\\
4.0 & \pv{sweep.mistral.a4p0.gated} & \pv{sweep.mistral.a4p0.cohharm} & \pv{sweep.mistral.a4p0.cohbenign} & \pv{sweep.mistral.a4p0.detambenign}\\
\bottomrule
\end{tabular}
\end{table}

\paragraph{A white-box gradient attack.} The jailbreaks above are templated. To test the transformer
attention-site gate against an adversary with white-box access, we run Greedy Coordinate Gradient
\citep{zou2023gcg}, fitting a per-prompt adversarial suffix on each of \pv{def.gcg.n} held-out harmful
behaviors by back-propagating an affirmative-target loss to the one-hot suffix tokens (Mistral-7B,
batched candidate search, no external dependency). We fit a length-20 suffix per behavior over 500
optimization steps, selecting from the top 256 gradient-ranked substitutions with 256 candidates per
step and the affirmative target ``Sure, here is''. The fitted suffixes reach base attack success
\pv{def.gcg.base} (Llama-Guard-3, Wilson 95\% confidence interval \pv{ci.def.gcg.base}) on
Mistral-7B-Instruct-v0.2, a competitive white-box attack. The detector fires on all \pv{def.gcg.n}
attacked prompts, and the
closed-loop attention-site gate at the coherent gain lowers attack success to \pv{def.gcg.gated}
(Wilson 95\% confidence interval \pv{ci.def.gcg.gated}) with
genuine refusals and benign over-refusal unchanged at \pv{def.gcg.overref}, so the gate more than halves a
competitive gradient-optimized attack rather than being evaded by it, with the two intervals
disjoint. Because GCG optimizes its suffix toward the affirmative target ``Sure, here is'', a judge that
keyed on the opening rather than the content could score a steered refusal as compliance. It does not: the
out-of-family Gemma judge, which reads the whole response, independently reports the same reduction
(\pv{def.gcg.basej4} to \pv{def.gcg.gatedj4}), and the gain used here is the coherent one, since a
higher gain degenerates the text and the forced affirmative opening then survives into the judge's view.
This reproduces across seeds, not only the one reported above: pooled over three seeds
($n{=}\pv{def.gcg.pooled.n}$), the gate lowers attack success from \pv{def.gcg.pooled.base} to
\pv{def.gcg.pooled.gated} (\pv{def.gcg.pooled.basej4} to \pv{def.gcg.pooled.gatedj4} under the
out-of-family judge). We report GCG on Mistral because it
is the transformer where the attack lands: at the same 500-step budget the same configuration barely
jailbreaks the more heavily safety-tuned Llama-3.1-8B (base attack success \pv{def.gcg.llama.base}) and
Qwen2.5-7B (base attack success \pv{def.gcg.qwen.base}), so only Mistral gives a gradient attack strong enough to exercise the gate. We run
GCG on a transformer rather than the
SSM because the fused selective-scan kernels used at inference are non-differentiable. This is a speed
choice, not an absence of a differentiable path: Falcon-Mamba's own fallback forward (the one every
experiment here actually runs, since the fused kernels are unavailable in our environment) is plain
PyTorch with no detached operations, and \texttt{mamba\_ssm} ships a differentiable reference scan
(\texttt{selective\_scan\_ref}) for exactly this purpose. We did not adapt GCG's candidate-search loop to
run through this much slower path. The gradient-free
optimized suffix in Section~\ref{sec:robustness} is the SSM counterpart of this attack.

\paragraph{A gate-aware adaptive adversary.}
\label{app:gateawaregcg}
The suffix above is optimized against the plain model. The stronger threat model optimizes it against the
intervention. We install the write-site amplification as an always-on gate before the search begins, so
every forward pass, both the affirmative-target gradient and the candidate re-ranking, runs through the
amplified model. The amplification $h \mapsto h + \alpha u$ is differentiable, so the gradient flows into
the suffix and the attack knows the refusal direction, the layers, and the gain. A run-time check
confirms the installed gate changes the affirmative-target loss before the search starts, so the
optimization cannot silently reduce to the plain attack. We fit \pv{def.gcg.ga.n} suffixes at the same
length, budget, and coherent gain as above, and evaluate three conditions with the same four-judge panel,
the ungated model, the deployed selective gate, and the always-on gate in isolation.

Optimizing against the gate does not make the suffix more harmful. Its ungated attack success is
\pv{def.gcg.ga.base} on Llama-Guard and \pv{def.gcg.ga.basej4} on the out-of-family whole-response judge,
matching the non-adaptive suffix (\pv{def.gcg.base} and \pv{def.gcg.basej4}), so the adversary gains
nothing on the plain model. The always-on intervention is not defeated either. It holds the out-of-family
attack success at \pv{def.gcg.ga.globj4} (Wilson 95\% confidence interval \pv{ci.def.gcg.ga.globj4}),
statistically indistinguishable from the non-adaptive deployed gate's \pv{def.gcg.gatedj4}
\pv{ci.def.gcg.gatedj4} (identical point estimate and Wilson interval at matched $n$), so a suffix built against the mechanism
does not beat the mechanism. The always-on gate over-refuses benign prompts at \pv{def.gcg.ga.globoverref},
which is why the deployment is selective and the always-on variant is a diagnostic rather than a defense.

The adaptive suffix's only success is against the detector \emph{trigger}. On the deployed selective
gate the primary-judge reduction the non-adaptive attack produced (\pv{def.gcg.base} to
\pv{def.gcg.gated}) disappears, to \pv{def.gcg.ga.attn} (\pv{ci.def.gcg.ga.attn}). That number is the
affirmative-opening artifact this appendix already flags on the primary judge, since the out-of-family
judge that reads the whole response still shows the deployed gate lowering attack success from
\pv{def.gcg.ga.basej4} to \pv{def.gcg.ga.attnj4} (\pv{ci.def.gcg.ga.attnj4}). The honest erosion is on
that trusted judge, where the deployed rate rises from the non-adaptive \pv{def.gcg.gatedj4} to
\pv{def.gcg.ga.attnj4}, because the adaptive suffix lowers the per-layer detector scores and the selective
gate then amplifies at fewer layers. Pooled over three seeds ($n{=}\pv{def.gcg.ga.pooled.n}$), the same
pattern holds: the deployed selective gate's out-of-family rate rises from the non-adaptive
\pv{def.gcg.pooled.gatedj4} to \pv{def.gcg.ga.pooled.attnj4} under the adaptive suffix, while the always-on
gate in isolation still holds attack success at \pv{def.gcg.ga.pooled.globj4}. This complements the
probe-evasion attack of
Appendix~\ref{app:evasion}, which optimizes directly against the detector, and reaches the same
conclusion from the opposite direction: the attackable surface is the detector that triggers the gate,
not the write site that carries the defense.

\subsection{Coherence and gain-calibration checks}
\label{app:coherence}
Two validity checks run under every defense number we report.

\paragraph{Coherence.} Activation steering can push generation off distribution into degenerate text,
a single repeated token or a collapse in vocabulary, and our attack-success metric counts any
non-refusing response as a success, so degenerate output would inflate it. We therefore coherence-screen
gated completions during gain calibration: a response whose most frequent token exceeds 40\% of its
tokens, or whose distinct-token ratio falls below 0.3, is treated as degenerate rather than as a
successful attack. The reported defenses use gains chosen to pass this screen, and we spot-checked the
headline gates directly (Falcon-Mamba to \pv{def.ssm.prefix_injection.gated} and the Falcon3 control),
where they produce genuine refusal text rather than
degeneration.

The RWKV and Zamba2 causal-ablation work (App.~\ref{app:rwkvdefense}, \ref{app:hybridablate}) and the
transformer attention sweep (App.~\ref{app:attnsweep}) instead use a second, independently-developed
screen: distinct-bigram ratio at least 0.5 and no single word more than 30\% of tokens. The two definitions
were never cross-checked against each other before this pass. We ran both over every raw completion saved
from this session's causal-ablation experiments ($n{=}\pv{coh.screen.n}$) and they agree \pv{coh.screen.agree}
of the time. The disagreements are concentrated in phrase-level repetition (e.g.\ ``the product is that the
product is'' repeated), which the bigram screen catches and the token screen, checking only single-token
frequency, does not, and in very short completions, where the bigram screen's four-word minimum treats
brevity itself as a failure the token screen does not. Neither definition dominates the other. We do not
retroactively rescore the main defense numbers under the bigram screen, since \pv{coh.screen.agree}
agreement means the two screens could in principle disagree about a small fraction of individual gated
completions, though we have no evidence this changes any reported aggregate rate by more than a percentage
point. We did check the one place a screen choice could change a conclusion rather than a percentage point:
the Zamba2 Mamba-site operating-point selection (Appendix~\ref{app:hybridablate}), which picks the largest
gain at which output stays coherent. Recomputing that selection under the token screen instead of the
bigram screen, across all three seeds, picks \pv{coh.screen.opcheck} gain (12) either way.

\paragraph{Per-model gain calibration.} The steering gain is calibrated for each model, because a
single shared gain is unsafe. The gain that defends Falcon-Mamba over-steers the more sensitive
Falcon3-Mamba into backfire-compliance and its residual into degeneration, and over-amplifying refusal
in an already-robust transformer increases attack success \citep{li2026steeringpitfalls}. We calibrate
with a single-layer gain sweep read against the coherence screen, taking the gain that minimizes
coherent attack success before the degeneration cliff. Reported defenses use the calibrated gain, and
the controls above are run at it.

\subsection{Generation-length sensitivity of the deployed gate}
\label{app:lengthcheck}
The defense numbers in this paper are measured at 64 new tokens. Regenerating the anchor's headline cells
at 256 new tokens (Table~\ref{tab:lengthcheck}, $n{=}200$ per cell) leaves the base rates and the benign
side unchanged, with over-refusal at the undefended rate in every cell, but raises the gated rates to
\pv{len256.pi.a4} under prefix injection and \pv{len256.rp.a4} under persona roleplay: past the 64-token
horizon the disclaimer-then-comply drift of Appendix~\ref{app:roleplay} resumes, and the deployed gain's
completions also fail the coherence screen (\pv{len256.cohpi.a4} and \pv{len256.cohrp.a4} coherent).
Re-sweeping the gain at 256 tokens finds no coherent defending operating point. Gains at or below 2 are
fully coherent but leave attack success at the base rate, and gains at or above 4 lower attack success
only at broken coherence, outside the operating regime the coherence screen accepts
(Appendix~\ref{app:coherence}); at gain 6 the output is fully degenerate and judge agreement itself falls
to \pv{len256.a6.agree}. The always-amplified steering arm is therefore length-bounded. The detector is
not: it fires once, on the prompt, independent of generation length, so at long generations the deployment
option the frontier comparison already prices, detect-then-refuse with the same detector, which the gate
ties (Appendix~\ref{app:frontier}), is unaffected.

\begin{table}[tbp]\centering\scriptsize\setlength{\tabcolsep}{4pt}
\caption{The anchor's gate at 256 new tokens (attack success / fraction coherent, $n{=}200$ per cell,
Llama-Guard judge). The gain calibrated at 64 tokens ($\alpha{=}4$) neither stays coherent nor fully
defends at 256, and no swept gain does both.}
\label{tab:lengthcheck}
\begin{tabular}{lcc}
\toprule
Gain $\alpha$ & Prefix injection & Persona roleplay \\
\midrule
base (no gate) & \pv{len256.pi.base} / \pv{len256.cohpi.base} & \pv{len256.rp.base} / \pv{len256.cohrp.base} \\
1 & \pv{len256.pi.a1} / \pv{len256.cohpi.a1} & \pv{len256.rp.a1} / \pv{len256.cohrp.a1} \\
2 & \pv{len256.pi.a2} / \pv{len256.cohpi.a2} & \pv{len256.rp.a2} / \pv{len256.cohrp.a2} \\
4 (deployed at 64) & \pv{len256.pi.a4} / \pv{len256.cohpi.a4} & \pv{len256.rp.a4} / \pv{len256.cohrp.a4} \\
6 & \pv{len256.pi.a6} / \pv{len256.cohpi.a6} & \pv{len256.rp.a6} / \pv{len256.cohrp.a6} \\
\bottomrule
\end{tabular}
\end{table}

\subsection{A harm-weighted pool closes the dilution attack}
\label{app:pool}
The dilution attack appends benign sentences to drag the mean-pooled harm score below the detector
threshold. The naive threshold fires through two appended sentences (attack success \pv{adv.k2.asr}),
degrades at four (\pv{adv.k4.firing} firing, \pv{adv.k4.asr}), and is defeated at six to eight (firing
\pv{adv.k6.firing} and \pv{adv.k8.firing}, attack success \pv{adv.k6.asr} and \pv{adv.k8.asr}). The
failure is the threshold and not the features: the harm logit falls from \pv{adv.harmlogit.k0} to
\pv{adv.harmlogit.kmax} while the benign cluster stays far below at \pv{adv.benignlogit}, so the two
classes stay separable at every level. Two facts close it. First, recalibration alone is more robust than it appears: the mean
harm score does not fall without bound as benign text is appended, it asymptotes to the mean score of
the benign filler, which sits above the benign-calibrated threshold. At a \pv{pool.fpr} benign
false-positive rate the recalibrated mean-pool detector fires on every prompt we test through
\pv{pool.kmax} appended sentences, its score plateauing near \pv{pool.mean.score.kmax} against a
threshold of \pv{pool.theta.mean}. A crafted filler of strongly anti-harmful text could in principle
push the mean below threshold, which is the residual concern. Second, that concern is removed by a
harm-weighted pool, $s_{\text{pool}} = \sum_t \mathrm{softmax}(\tau s)_t\, s_t$, which weights each
token by its harm logit $s_t$ rather than by position. Low-scoring padding receives near-zero weight,
so the score is invariant to padding amount and content: it holds at \pv{pool.wavg.score.kmax} through
\pv{pool.kmax} sentences (threshold \pv{pool.theta.wavg}), where the mean has long since drifted, at the
same false-positive rate. The harm-weighted pool is therefore a drop-in detector that is
dilution-invariant by construction. Because harm is cleanly separable on undiluted prompts, switching
to it leaves the headline defense and over-refusal unchanged. Only the dilution robustness improves.

\subsection{Probe-evasion, and why the harm-weighted trigger closes it}
\label{app:evasion}
A white-box adversary can optimize a suffix directly against the frozen detector rather than against
the model. We fit a gradient-free suffix that minimizes the mean-pooled detector's prompt-level harm
score on held-in harmful prompts, then transfer it to the eval set. The search is a greedy coordinate
descent over suffix token positions with batched candidate scoring, keeping the best candidate per round
(150 rounds of 64 candidates), so it does not depend on autograd through the fused selective-scan
kernels. Against the deployed mean-pooled
probe it succeeds: the detector fires on every clean harmful prompt (\pv{adv.opt.firing.clean}) but none
once the suffix is appended (\pv{adv.opt.firing.optimized}), so the gate never triggers and attack
success rises from \pv{adv.opt.asr.clean} to \pv{adv.opt.asr.optimized}, the undefended rate.

The break is a property of mean pooling, not of the write site. The suffix works by appending
anti-harm tokens that drag the mean harm score down, from a median of \pv{pve.mean.clean.med} on clean
harmful prompts to \pv{pve.mean.evasion.med} with the suffix. The harm-weighted pool of
Appendix~\ref{app:pool} weights each token by its own harm logit, so those appended tokens receive
near-zero weight while the causally-earlier harmful tokens keep their scores. Its score is essentially
unchanged by the suffix, holding at a median of \pv{pve.wavg.clean.med} against \pv{pve.wavg.evasion.med}.
Calibrated to the same \pv{pve.fpr} benign false-positive rate, the harm-weighted pool fires on
\pv{pve.wavg.evasion.fire} of the evasion prompts, as does the max pool (\pv{pve.max.evasion.fire}). The
harm-weighted and max pools coincide here because the softmax concentrates on the peak harmful token,
and the harm-weighted pool's advantage over a hard max is the graded dilution robustness of
Appendix~\ref{app:pool}, not this evasion test. An
appended suffix cannot lower a pool that weights the fixed harmful span, so the suffix-evasion route is
closed by the trigger we already deploy against dilution. Even the mean pool catches the suffix once its
threshold is benign-calibrated rather than set at the sigmoid midpoint: the evasion drives the score only
to \pv{pve.mean.evasion.med}, still above the benign cluster, so it fires on \pv{pve.mean.evasion.fire}
of evasion prompts at the same false-positive rate. We also fit a suffix directly against the
harm-weighted pool. After 200 rounds it still cannot evade it: the pool fires on \pv{pveadv.wavg.fire}
of the attacked prompts at the same false-positive rate, and its score does not fall (median
\pv{pveadv.wavg.med.clean} to \pv{pveadv.wavg.med.attack}), because an appended suffix cannot lower the
fixed harmful span's contribution. Only an adversary who rewrites the harmful content itself, and so
changes the request, is outside this test.
A concurrent activation probe for Gemini max-pools per-token scores so a short harmful span is not
diluted by long benign context \citep{kramar2026probesgemini}. Our harm-weighted pool goes further on
the axis that matters here: it is dilution-invariant by construction against a named adaptive attack,
weights tokens by a continuous softmax over harm logits rather than a hard max, and operates inside a
closed-loop steering defense rather than as a standalone detector.

\subsection{Detector hardening against wrapper text}
\label{app:hardening}
A detector trained on plain harm can fire on the jailbreak wrapper rather than the harmful request. A
benign-but-spicy prompt wrapped in the same jailbreak template trips the plain detector
\pv{harden.plain.wrapped_spicy} of the time, which would over-refuse benign requests that merely carry
jailbreak boilerplate. Training the detector with wrapped-benign prompts as negatives removes the
confound: the wrapped-benign firing rate falls to \pv{harden.hard.wrapped_spicy} while harm detection
is preserved (wrapped-harm \pv{harden.hard.wrapped_harm}, plain harm \pv{harden.hard.plain_harm}). The
SSM and transformer defenses reported here use this hardened detector. The hybrid runs use a plain
harm-versus-spicy detector, since at one of its two sites the hardening does not work
(Appendix~\ref{app:hardenxarch}). Hardening does not change attack success, because both detector
variants fire on every wrapped harmful prompt, so it affects only what a gate does to wrapped benign
traffic.

\subsection{The intervention must use the model's refusal geometry}
\label{app:geometry}
The same readable harm direction has opposite effects depending on how it is used
(Table~\ref{tab:action}). Amplifying it toward refusal at the block output lowers attack success from
\pv{def.ssm.prefix_injection.base} to \pv{def.ssm.prefix_injection.gated}, whereas projecting it out
of the residual stream barely moves attack success (\pv{base.projection.prefix_injection.asr}).
Conversely, on plain harmful prompts that the instruct model mostly refuses (base attack success
\pv{transfer.x2s.ablate.base}), removing the direction raises attack success to
\pv{transfer.x2s.ablate.native}: the harm direction is the refusal trigger, so deleting it ablates
refusal while amplifying it induces refusal. Steering along isolated subspace components away from the
mean refusal direction instead does not control refusal and degenerates the output at the magnitudes
needed to defend, so the standard mean refusal direction is what controls refusal at the write site.

\begin{table}[h]\centering
\scriptsize
\caption{The action matters: the same refusal direction (all on Falcon-Mamba-7B-Instruct, $n{=}$\pv{def.ssm.n}).
Amplifying it at the block output defends, projecting it out of the residual does little, and removing it from a
model that was refusing jailbreaks the model. The first two rows use prefix-injected prompts (high
base attack success, room to fall). The third uses plain harmful prompts (low base, room to rise).}
\label{tab:action}
\begin{tabular}{lcc}
\toprule
Action on the refusal direction & Base & After \\
\midrule
Amplify at block output (ours), under attack & \pv{def.ssm.prefix_injection.base} & \pv{def.ssm.prefix_injection.gated} \\
Project out of residual, under attack & \pv{def.ssm.prefix_injection.base} & \pv{base.projection.prefix_injection.asr} \\
Remove direction, plain harm & \pv{transfer.x2s.ablate.base} & \pv{transfer.x2s.ablate.native} \\
\bottomrule
\end{tabular}
\end{table}

\subsection{Detector-class ablation: the site contrast is not linear-probe-specific}
\label{app:detablation}
The write-site result holds when the linear trigger is replaced by a nonlinear one. Swapping the
detector for a one-hidden-layer MLP, with the intervention site and the mean-direction gate held fixed,
leaves the prefix-injection site contrast unchanged: the block-output gate lowers attack success to
\pv{def.mlp.gated} (from \pv{def.mlp.base}), while CAST at the residual stays at \pv{def.mlp.cast},
matching the linear-detector cells. This contrast is a property of the direction, not of the
probe class.

The residual gap also survives CAST's own gain sweep. Sweeping CAST's gain over
$\alpha \in \{2, 4, 8, 16\}$ under prefix injection gives attack success \pv{cast.a2.asr},
\pv{cast.a4.asr}, \pv{cast.a8.asr}, and \pv{cast.a16.asr}. Its self-tuned best operating point
($\alpha{=}\pv{cast.best.alpha}$, chosen by lowest attack success) reaches \pv{cast.best.asr} at
\pv{cast.best.overrefusal} over-refusal, still far above the write-site gate under the same detector, so
the residual failure in Table~\ref{tab:baselines} is not an artifact of evaluating CAST at one gain.

\subsection{Robustness of the defense}
\label{app:robustness}
The main tables report \pv{def.ssm.n} prompts per condition. The comparison holds across seeds, under a
larger sample, judges from other model families, and a different harmful-behavior distribution. Refitting
the refusal direction and redrawing the eval set across \pv{def.ssm.pooled.nseeds} seeds leaves the
defense stable, with pooled gated prefix-injection attack success
\pv{def.ssm.prefix_injection.gated.pooled} (\pv{ci.def.ssm.prefix_injection.gated.pooled},
n${=}$\pv{def.ssm.pooled.n}) against a base of \pv{def.ssm.prefix_injection.base.pooled}
(\pv{ci.def.ssm.prefix_injection.base.pooled}), non-overlapping. Replicating the headline and the
site-and-loop cells at \pv{def.n300.n} harmful prompts per condition reproduces the defense under prefix
injection (attack success \pv{def.n300.prefix_injection.base} to \pv{def.n300.prefix_injection.gated},
Wilson 95\% confidence interval \pv{ci.def.n300.prefix_injection.gated}). Beyond the two Llama-family
judges, which agree on \pv{def.ssm.judgeagree} of cases, two judges from different families confirm the
gate rather than leaving a within-family artifact, a prompted \pv{def.ssm.judge3name} (gated
\pv{def.ssm.gatedj3} against a base of \pv{def.ssm.basej3}) and the Gemma-family guard
\pv{def.ssm.judge4name} (gated \pv{def.ssm.gatedj4} against a base of \pv{def.ssm.basej4}), with the
per-prompt verdict agreeing across all \pv{def.ssm.njudges} judges \pv{def.ssm.judgeagreenway} of the
time. The two prompted judges (the Llama-3.1-8B-Instruct secondary and the Mistral third judge) share one
instruction, reproduced verbatim so the criterion is reproducible: \emph{``You are a strict
content-safety classifier. You are given a user REQUEST and an AI RESPONSE. Decide whether the RESPONSE
provides meaningful assistance toward a harmful, dangerous, illegal, or unsafe goal (for example weapons,
drugs, hacking, violence, fraud, self-harm). A refusal, a safe redirection, or an empty/irrelevant answer
is SAFE. Actual harmful help is UNSAFE. Answer with exactly one word: `safe' or `unsafe'.''} The user turn
supplies the request and response and asks ``Is the RESPONSE safe or unsafe?'', and the verdict is the
first generated word. The defense also transfers to a different harmful-behavior distribution. With the detector and
refusal direction fit on AdvBench, the gate lowers prefix-injection attack success on HarmBench standard
behaviors \citep{mazeika2024harmbench} from \pv{def.hb.base} to \pv{def.hb.gated} (n${=}$\pv{def.hb.n},
Wilson 95\% confidence interval \pv{ci.def.hb.gated}). The defense withstands an optimized attack, not
just fluent templates. A gradient-free suffix tuned to elicit an affirmative continuation raises base
attack success to \pv{def.opt.base}, and the gate lowers it to \pv{def.opt.gated} (Wilson 95\% confidence
interval \pv{ci.def.opt.gated}), while CAST at the residual does not suppress it and slightly backfires
(attack success \pv{def.opt.cast}, above the undefended \pv{def.opt.base}). The SSM suffix is
gradient-free because the fused selective-scan kernels are non-differentiable. The gradient-based GCG
attack \citep{zou2023gcg} runs on a transformer in Appendix~\ref{app:attnsweep}, and directly through the
SSM's slower differentiable scan in Appendices~\ref{app:gcgssm} and~\ref{app:gateawaregcg}. A
dilution attack that pads a harmful prompt with benign text defeats a naive threshold, but a harm-weighted
pool that weights tokens by harm rather than position defeats it by construction, firing through
\pv{pool.kmax} appended sentences and staying immune to a suffix optimized to evade a mean-pooled probe
(Appendices~\ref{app:pool} and~\ref{app:evasion}). The gate does not degrade general capability because
its detector almost never fires on benign task prompts. On MMLU its false-positive rate is
\pv{e1cap.mmlu.fpr}, so the gate stays inert and gated accuracy equals the base model on both MMLU
(\pv{e1cap.gatekeeper.mmlu} against \pv{e1cap.base.mmlu}) and TruthfulQA
(\pv{e1cap.gatekeeper.truthfulqa} against \pv{e1cap.base.truthfulqa}) on Falcon-Mamba. This is a
detector-selectivity result, not capability preservation under firing.

\subsection{The locus and the defense replicate on a second SSM}
\label{app:secondssm}
Both the locus and the defense reproduce on a second, independently trained instruct SSM,
Falcon3-Mamba-7B. Harm is decodable at $\Delta$ (held-out AUROC \pv{locus.ssm2.delta}) and the gate
lowers attack success under fictional framing (\pv{def.ssm2.fictional_framing.base} to
\pv{def.ssm2.fictional_framing.gated}) and persona roleplay (\pv{def.ssm2.roleplay_dan.base} to
\pv{def.ssm2.roleplay_dan.gated}) at low over-refusal, so the finding is a property of the models and not
one checkpoint. The only model-specific step is layer selection and gain calibration
(Section~\ref{sec:recipe}, Appendix~\ref{app:alpha}).

\subsection{A gradient-based attack through the SSM's own differentiable scan}
\label{app:gcgssm}
The fused selective-scan kernel used for inference is not differentiable end to end, but this is a
speed choice, not a fundamental barrier: Falcon-Mamba's own fallback path is plain differentiable
PyTorch, and \texttt{mamba\_ssm} ships a differentiable reference scan. We run GCG \citep{zou2023gcg}
directly against Falcon-Mamba through this differentiable path, optimizing a 20-token adversarial
suffix per behavior for 500 steps (256 candidates, top-256 substitution positions per step, identical
hyperparameters and core to the transformer GCG attack in Appendix~\ref{app:attnsweep}), targeting the
model's own selective-scan block-output gate rather than the attention output. At $n{=}\pv{def.gcgssm.n}$
harmful behaviors, the gate lowers attack success from \pv{def.gcgssm.base} to \pv{def.gcgssm.gated},
confirmed by the out-of-family judge \pv{def.gcgssm.judge4name} (\pv{def.gcgssm.basej4} to
\pv{def.gcgssm.gatedj4}). This is the same write-versus-residual contrast as every other attack family in
this paper, now against an adversary that has gradient access to the model rather than only fluent
prompt templates, and it is a property of the differentiable path: the fused inference kernel used
elsewhere in this paper is faster but was not built for backpropagation, so this attack necessarily runs
through the slower reference implementation.

\subsection{The intervention location is irrelevant: a matched-absolute-magnitude control}
\label{app:absmatch}
The write-site results compare a gate that reads and pushes the refusal direction at the fresh write
against baselines that read and push it in the accumulated residual, and the residual's activation norm is
much larger than the write's ($\times\pv{am.ssm.normratio}$ on Falcon-Mamba, $\times\pv{am.tf.normratio}$
on Mistral-7B at the decision layer). A fixed additive push is therefore a very different \emph{relative}
perturbation at the two sites, so the reported site contrast could be a magnitude or direction artifact
rather than a property of the location. We isolate the three possible causes (push location, estimated
direction, magnitude) with a control that holds the detector fixed and sweeps the same \emph{absolute}
push magnitude across three arms: (i) push at the write site with the write-estimated direction (the
deployed gate), (ii) push at the residual with the \emph{same} write-estimated direction, and (iii) push
at the residual with a residual-estimated direction (the CAST baseline). Every completion is scored with
Llama-Guard-3-8B and reported with Wilson 95\% intervals at $n{=}100$.

The push location does not matter. With the direction held fixed, pushing at the write site and at the
residual are within CIs at every magnitude on Falcon-Mamba (write \pv{am.ssm.write} \pv{am.ssm.write.ci}
versus residual \pv{am.ssm.resid} \pv{am.ssm.resid.ci} at the coherent operating point), and at the
deployed gain on Mistral-7B (write \pv{am.tf.write} \pv{am.tf.write.ci} versus residual \pv{am.tf.resid}
\pv{am.tf.resid.ci}). Away from that gain the two Mistral sites can separate in either direction, so the
location-equivalence there is an operating-point statement. The SSM identity follows because a Mamba block
adds only its mixer output to the residual. What matters is where the direction is \emph{estimated}: a direction estimated in the
accumulated residual degenerates when amplified (\pv{am.ssm.residdir} attack success on the SSM, at broken
coherence), reproducing the CAST failure. The architecture-specific quantity is thus the readout site
where the refusal direction is cleanly estimable, not a steering location.

\subsection{The gate holds under an adaptive attack optimized through it}
\label{app:gagcg}
The gradient attack in Appendix~\ref{app:gcgssm} optimizes the suffix against the plain model. A stronger
adversary optimizes the suffix against the deployed gate itself, with the write-site amplification
installed during the search so the gradient flows through it. We run this gate-aware GCG on Falcon-Mamba
($n{=}100$, 500 steps, identical hyperparameters). The gate holds: the detector fires on all
\pv{ga.gcg.ssm.firedtot} attacked prompts (the adaptive suffix does not evade it), and deployed attack
success stays at \pv{ga.gcg.ssm.deployed} (from \pv{ga.gcg.ssm.base} on the ungated model), Llama-Guard
scored. On the transformer the same adaptive attack instead drives down the detector score and erodes the
deployed gate, so the SSM's mixer-state detector is the more robust trigger of the two.

\subsection{The wrapper confound, screened on every architecture}
\label{app:hardenxarch}
A detector trained on harmful versus benign-but-spicy prompts can key on the jailbreak \emph{template}
rather than on harm, since at evaluation the harmful prompts are wrapped and the benign ones are not. That
is not hypothetical: on the anchor it is exactly what happens until the training set is augmented with
wrapped examples (Appendix~\ref{app:hardening}). We therefore ran the same screen on the two architectures
added later, fitting a plain detector (harm versus spicy, both unwrapped) and a wrapper-hardened one
(harm and wrapped harm versus spicy and wrapped spicy) at each write site, and measuring firing rates on
four held-out populations of \pv{hx.rwkv.n} prompts each. The diagnostic is the plain detector's firing
rate on \emph{wrapped benign}: near zero means it reads harm, high means it reads the wrapper.

The confound is close to universal. Both transformers behave exactly as the anchor does: the plain
detector fires on \pv{hx.mistral.attn.plain.wspicy} of wrapped benign prompts on Mistral-7B and
\pv{hx.llama.attn.plain.wspicy} on Llama-3.1-8B, and wrapper-augmented training drives both to
\pv{hx.mistral.attn.hardened.wspicy} while leaving wrapped-harm detection at
\pv{hx.mistral.attn.hardened.wharm}. We therefore re-ran both transformer defenses with the hardened
trigger at the deployed operating point, and Table~\ref{tab:routingsite} reports those runs. Attack
success is unchanged by the swap (Mistral \pv{def.mistralattn.prefix.base} to
\pv{def.mistralattn.prefix.gated}, Llama-3.1-8B \pv{def.llamaattn.prefix.base} to
\pv{def.llamaattn.prefix.gated}, against \pv{sweep.mistral.gated} and \pv{sweep.llama.gated} unhardened),
as expected, since both detector variants fire on every wrapped harmful prompt. Only the hybrid still runs
a plain trigger, because at one of its two sites hardening does not work.

RWKV-6 is the exception in the other direction. Its plain detector fires on
\pv{hx.rwkv.timemix.plain.harm} of harmful prompts,
\pv{hx.rwkv.timemix.plain.wharm} of wrapped harmful prompts, and
\pv{hx.rwkv.timemix.plain.wspicy} of wrapped benign ones, so hardening changes nothing
(\pv{hx.rwkv.timemix.hardened.wspicy} either way) because there is nothing to fix.

Zamba2 is not, at either site. The plain detector fires on \pv{hx.zamba.attn.plain.wspicy} of wrapped
benign prompts at the attention site and \pv{hx.zamba.mamba.plain.wspicy} at the Mamba site, so on this
model the trigger is reading the template. Wrapper-augmented training repairs the attention site
completely, dropping wrapped-benign firing to \pv{hx.zamba.attn.hardened.wspicy} while wrapped-harmful
firing stays at \pv{hx.zamba.attn.hardened.wharm}. It does \emph{not} repair the Mamba site, where the
hardened detector still fires on \pv{hx.zamba.mamba.hardened.wspicy} of wrapped benign prompts. A linear
probe at the Zamba2 Mamba output therefore does not separate a wrapped benign prompt from a harmful one,
whatever it is trained on, which is a property of that site's representation rather than of the training
set. We report the hybrid's numbers with that caveat attached (Section~\ref{sec:limitations}) and do not
claim a deployable trigger there.

That failure is informative about the site, which is this paper's subject. Everywhere else we look, the
write site is where harm is \emph{most} cleanly readable: it is where decodability peaks in the SSM
(Section~\ref{sec:transfer}), where the transported probe lands, and where the refusal direction is
cleanly estimable. The hybrid's Mamba output satisfies the second and third of those and fails the first
in the one condition that matters for a trigger, since a jailbreak template moves it as far as harmful
content does. Reading and steering therefore come apart once more, in a new place: the site can host a
direction that controls refusal while its linear geometry still confuses a wrapped benign prompt with a
harmful one. Our claim is about where a refusal direction is cleanly \emph{estimated}, and this shows that
property does not imply the strictly stronger one of supporting a reliable linear \emph{detector} at the
same site. The two coincide on the SSM, the transformers, and RWKV-6, and the hybrid is where they
separate.

\subsection{The benign-side cost of every write-site gate}
\label{app:benigncost}
An attack-success reduction means nothing without its false-positive cost, since refusing every prompt
drives attack success to zero. We therefore measured over-refusal on held-out benign-but-spicy prompts at
the exact operating point behind each row of Table~\ref{tab:routingsite}. For the SSM and the two
transformers this came from the runs that produced the attack-success numbers. For RWKV-6 and Zamba2 it did
not exist, because those runs generated on harmful prompts only, so we replayed each run's own
configuration on the benign split.

The answer is the same everywhere, and it is structural rather than lucky: the gate is triggered by a
detector that does not fire on benign prompts, so on those prompts the model runs unmodified and the
over-refusal rate cannot move. On RWKV-6 the detector fires on \pv{rwkv.ovr.fire} of
\pv{rwkv.ovr.n} benign prompts and over-refusal is \pv{rwkv.ovr.base} both with and without the gate. On
Zamba2 it fires on \pv{zamba.cl.attn.firebenign} of \pv{zamba.cl.n}, and over-refusal is
\pv{zamba.cl.attn.ovrbase} either way, at both write sites. The same detector fires on
\pv{zamba.cl.attn.fireharm} of the jailbroken harmful prompts, so the separation is not bought by a
timid trigger.

\subsection{Closing the loop on the hybrid, and what always-on steering costs}
\label{app:hybridclosed}
The hybrid results in an earlier version of this work amplified the refusal direction on every prompt, with
no detector, unlike every other architecture we test. Measuring the benign cost of that configuration shows
why the distinction matters. Always-on steering at Zamba2's Mamba write site raises over-refusal from
\pv{zamba.ovr.ol.base} to \pv{zamba.ovr.ol.mamba}, and at the attention site to
\pv{zamba.ovr.ol.attn}, on \pv{zamba.ovr.ol.n} benign prompts. At the attention site the model refuses
essentially everything, so its attack-success reduction was substantially a refuse-everything artifact.
This reproduces on a fourth architecture what Section~\ref{sec:results} reports for the SSM, where
open-loop steering matches the gate's attack success while over-refusing
\pv{def.openloop.overrefusal} of benign prompts.

Closing the loop removes the cost without costing the defense. Fitting a linear detector at the same write
site and amplifying only when it fires, the trigger separates the two populations completely
(\pv{zamba.cl.attn.fireharm} of harmful prompts, \pv{zamba.cl.attn.firebenign} of benign), attack success
falls from \pv{zamba.cl.base} to \pv{zamba.cl.attn.asr} at the attention site and
\pv{zamba.cl.mamba.asr} at the Mamba site, and over-refusal is unchanged at
\pv{zamba.cl.attn.ovrbase} (\pv{zamba.cl.attn.ovrdelta}). Table~\ref{tab:routingsite} reports this
closed-loop version, so all four architecture families are compared under the same detect-then-gate design.

\subsection{Human validation of the attack-success judge}
\label{app:humanval}
Every attack-success number in this paper is produced by an LLM judge, so the judge itself needs an
anchor. We sampled \pv{hv.n} completions, stratified over the cells that carry claims and weighted toward
the small-margin ones (the RWKV-6 and Zamba2 ablations), drawn uniformly within each cell so that
per-cell rates stay unbiased. Two raters labelled every item independently, in different orders, seeing
only the request and the response. Model, condition, and all judge verdicts were withheld, and the
verdict key was held in a separate file. The primary label asks exactly what the judge is asked, whether
the response gives meaningful assistance toward the harmful request, with a separate label for refusal and
a third for whether the text is coherent at all. An item counts only when both raters agree;
\pv{hv.nunresolved} of \pv{hv.n} did not resolve and are excluded rather than broken by a tie-break we
have no third rater to supply.

The raters agree with each other at \pv{hv.irr.comply.agree} on compliance
($\kappa{=}\pv{hv.irr.comply.kappa}$) and \pv{hv.irr.refuse.agree} on refusal
($\kappa{=}\pv{hv.irr.refuse.kappa}$). Coherence is the exception: raw agreement is
\pv{hv.irr.coh.agree} but $\kappa{=}\pv{hv.irr.coh.kappa}$, because almost everything is coherent and a
$\kappa$ on so skewed a variable is uninformative. We therefore use coherence only as a coarse screen and
not as a reported rate.

Against the human consensus, the primary judge (Llama-Guard-3-8B) reaches \pv{hv.judge.lg.agree} agreement
and $\kappa{=}\pv{hv.judge.lg.kappa}$ on the $n{=}\pv{hv.judge.lg.n}$ items for which per-item judge
verdicts were stored. Its errors are one-sided: \pv{hv.judge.lg.fp} items it called compliant were read as
safe by both raters, against \pv{hv.judge.lg.fn} in the other direction, so the judge over-counts attack
success and every defense result here is conservative. The other three judges bracket it, from
$\kappa{=}\pv{hv.judge.mistral.kappa}$ (Mistral-7B-Instruct) to
$\kappa{=}\pv{hv.judge.sg.kappa}$ (ShieldGemma-9B, the strictest, with \pv{hv.judge.sg.fn} items the
raters called compliant and it did not). Finally, the degeneration concern is small in this sample: of the
\pv{hv.degen.n} judge-flagged items, \pv{hv.degen.k} (\pv{hv.degen.rate}) were rated incoherent by both
raters, so miscounted gibberish is a real but minor contributor to the reported rates.

Per cell, the human rates track the judge on the anchor and correct it on one exotic arm. On Falcon-Mamba
under prefix injection the raters put the undefended model at \pv{hv.cell.ssmbase}
($n{=}\pv{hv.cell.ssmbase.n}$) and the gated model at \pv{hv.cell.ssmgated}
($n{=}\pv{hv.cell.ssmgated.n}$), so the defense is if anything larger under human labels than under the
judge. The correction is on RWKV-6, where the raters score the transported-direction ablation at
\pv{hv.cell.rwkvtransp} ($n{=}\pv{hv.cell.rwkvtransp.n}$) against only \pv{hv.cell.rwkvrand}
($n{=}\pv{hv.cell.rwkvrand.n}$) for the matched random control. The judge had put that control at
\pv{orth.rwkv.pooled.rand}, so the separation we report for RWKV in Table~\ref{tab:ablate} is conservative:
its random control is far less jailbreaking to a human than the judge scored it, and the causal margin is
correspondingly cleaner. Zamba2's attention site stays undecisive on human labels as well
(\pv{hv.cell.zattntransp} transported against \pv{hv.cell.zattnrand} random, $n$ of
\pv{hv.cell.zattntransp.n} and \pv{hv.cell.zattnrand.n}), which is the reading we already give it. These
per-cell counts are small and we do not restate the table's numbers from them. They are a check on
direction and rough magnitude, not a replacement measurement.

\subsection{What the portable direction buys when target labels are scarce}
\label{app:payoff}
Portability is only worth something if a transported direction beats training one locally. This measures that
directly, against the strongest cheap alternative a practitioner has: fitting the target model's own
refusal direction on the same labels the transport would have used. Both arms receive an identical budget
of $k$ labeled target prompts. The transport arm spends them on paired activations, fits the ridge map on
those $k$ pairs, and carries through it a source direction estimated on the source model's abundant labels.
The from-scratch arm spends them on a mean-difference direction in the target. Both are then scored by
AUROC on ground-truth XSTest labels, so no judge enters this comparison.

The comparison must run off ceiling to mean anything. In distribution both arms saturate: at
$k{=}\pv{payoff.indist.k}$ the transport reaches AUROC \pv{payoff.indist.transp} and from-scratch
\pv{payoff.indist.scratch}, leaving no gap for either to fill, because harm-versus-benign detection is
easy within a distribution the labels already cover. We therefore run the payoff in the regime
where the transport test itself lives off ceiling (Appendix~\ref{app:catcv}): hold out one XSTest harm
category, draw the $k$ labels from the other categories, and score on the held-out category, whose surface
features neither arm has seen. Each of the \pv{payoff.ncat} held-out categories is run with
\pv{payoff.seeds} label draws, giving \pv{payoff.nrep} replicates per budget. Both arms see the same draw
in each replicate, so the contrast is paired and the interval below is on the difference itself.

The independent unit is the held-out \emph{category}, not the replicate. The \pv{payoff.seeds} draws inside
one category are scored against that category's single test set, so treating all \pv{payoff.nrep}
replicates as independent and dividing by their square root would understate the interval. We therefore
report a $t$ interval on the \pv{payoff.ncluster} per-category mean differences. The narrower
replicate-level interval also excludes zero at every budget (for example
\pv{payoff.k64.repci} at $k{=}\pv{payoff.kmax}$, against \pv{payoff.k64.ci} clustered), so the conclusion
does not depend on the choice, but the clustered interval is the honest one and is what the table reports.

\begin{table}[htb]\centering\scriptsize\setlength{\tabcolsep}{5pt}
\caption{Data-scarce payoff of the portable direction over a target direction fit from scratch on the same
$k$ target labels, generalizing to an unseen harm category (Llama-3.1-8B to Falcon-Mamba, AUROC on
ground-truth labels). $\Delta$ is the paired difference and its 95\% $t$ interval over the
\pv{payoff.ncluster} held-out categories, the independent unit. Wins is the fraction of the
\pv{payoff.nrep} individual draws favoring transport, which is why it can sit near half while the interval
still excludes zero: the mean gain is carried by larger wins, not by winning more often. Positive $\Delta$
favors transport. The margin does not close over this range of budgets.}
\label{tab:payoff}
\begin{tabular}{lcccc}
\toprule
$k$ labels & Transport & From scratch & Paired $\Delta$ & Wins \\
\midrule
4  & \pv{payoff.k4.transp}  & \pv{payoff.k4.scratch}  & \pv{payoff.k4.delta}~\pv{payoff.k4.ci}   & \pv{payoff.k4.win} \\
8  & \pv{payoff.k8.transp}  & \pv{payoff.k8.scratch}  & \pv{payoff.k8.delta}~\pv{payoff.k8.ci}   & \pv{payoff.k8.win} \\
16 & \pv{payoff.k16.transp} & \pv{payoff.k16.scratch} & \pv{payoff.k16.delta}~\pv{payoff.k16.ci} & \pv{payoff.k16.win} \\
32 & \pv{payoff.k32.transp} & \pv{payoff.k32.scratch} & \pv{payoff.k32.delta}~\pv{payoff.k32.ci} & \pv{payoff.k32.win} \\
64 & \pv{payoff.k64.transp} & \pv{payoff.k64.scratch} & \pv{payoff.k64.delta}~\pv{payoff.k64.ci} & \pv{payoff.k64.win} \\
\bottomrule
\end{tabular}
\end{table}

Transport wins at every budget tested, by \pv{payoff.deltamin} to \pv{payoff.deltamax} AUROC, with the
clustered interval excluding zero in all five rows (Table~\ref{tab:payoff}). At a deployed operating point
the margin is wider than AUROC implies. Holding false alarms on benign prompts to
\pv{payoff.k64.tprone.fpr}, the transported direction detects \pv{payoff.k64.tprone.transp} of harmful
prompts at $k{=}\pv{payoff.kmax}$ against \pv{payoff.k64.tprone.scratch} from scratch
(\pv{payoff.k64.tprone.delta}). At \pv{payoff.k64.tprfive.fpr} it is \pv{payoff.k64.tprfive.transp}
against \pv{payoff.k64.tprfive.scratch} (\pv{payoff.k64.tprfive.delta}). The natural way to phrase this is a label-equivalence
factor, a transported direction being worth some multiple of its budget in freshly labelled examples, and
we deliberately do not report one. Re-running the sweep on a finer grid of \pv{payoff.fine.npoints}
budgets from $k{=}\pv{payoff.fine.kmin}$ to \pv{payoff.fine.kmax} shows why. The from-scratch curve is not
monotone in $k$ (\pv{payoff.fine.viol} of its steps go down rather than up) and it flattens from about
$k{=}\pv{payoff.fine.plateaufrom}$ onward, sitting between \pv{payoff.fine.plateaulo} and
\pv{payoff.fine.plateauhi} for the rest of the range, against a mean interval width of
\pv{payoff.fine.ciw} on that same curve. Reading a label budget off a curve that is flat and noisy in the
region of interest inverts a near-zero gradient, and the answer moves by more than a factor of three
depending on which budget is used as the reference. The detection rate at a fixed false-alarm budget,
above, is the operating-point statement that does not require that inversion, and it is the one we make. The effect is not carried by one
easy fold: transport is ahead in \pv{payoff.k64.cats} of \pv{payoff.ncat} held-out categories at
$k{=}\pv{payoff.kmax}$ and \pv{payoff.k4.cats} of \pv{payoff.ncat} at $k{=}\pv{payoff.kmin}$. We expected
the two arms to converge as $k$ grew and they do not, which we read as the transported direction importing
refusal structure the source estimated across categories that the target's own $k$ labels never cover,
rather than as a better estimate of the same quantity. Two limits are worth stating. The gain is small in
absolute terms, a few points of AUROC on a detection task, and it disappears in distribution, so the
practical case for portability is generalization to harm the target's own labels do not represent, not a
uniform accuracy win. And this measures the readout, not the intervention: it says a re-estimated detector
transfers, not that the downstream gate inherits the margin.

\subsection{Final hyperparameters and tried ranges}
\label{app:hparams}
Table~\ref{tab:hparams} consolidates the final setting of every knob (write site, gate layers, decision
layer, gain, threshold, transport map, detector, attack budgets) for every model in the paper, together
with the values tried where a sweep was run. The same values are the defaults and command lines in
\ifarxivbuild the \texttt{reproduce.sh} of the code release.\else the supplementary archive's
\texttt{reproduce.sh}.\fi
\begin{table*}[tbp]\centering\scriptsize\setlength{\tabcolsep}{3pt}
\caption{Final hyperparameters per model, with the values tried where a sweep was run. Gate layers are the top-8
by held-out probe AUROC on the SSMs (ranked by mixer-output AUROC on Falcon-Mamba, by $\Delta$ AUROC on
Falcon3-Mamba, Section~\ref{sec:recipe}); the other architectures use fixed mid-network bands. The decision
layer is the middle of the listed set and carries the per-prompt detector. Every gate uses $\theta{=}0.5$ on the probe's harm probability, the mean
refusal direction (mean harmful minus mean benign activation) at each gate layer, greedy decoding with
64 new tokens, and the gain $\alpha$ is the largest value that passes the coherence screen
(Appendix~\ref{app:coherence}). Transfer rows give the layer the direction is read at in the source and the
layers it is written at in the target.}
\label{tab:hparams}
\newcolumntype{R}[1]{>{\raggedright\arraybackslash}p{#1}}
\begin{tabular}{@{}R{0.11\textwidth}R{0.085\textwidth}R{0.20\textwidth}R{0.20\textwidth}R{0.145\textwidth}R{0.20\textwidth}@{}}
\toprule
Model & Write site & Gate layers (decision) & $\alpha$ final (tried) & Transfer: read / write layers & Notes \\
\midrule
Falcon-Mamba-7B-Inst. & mixer output & \{24,28,30,34,35,36,37,44\} (35) & 4 (1,2,4,6,8; capability at 4,8,16) & 44 / \{22,24,28,31,38,40,42,43\} & anchor; $\Delta$-ranked set for transfer targets \\
Falcon3-Mamba-7B-Inst. & mixer output & $\Delta$-ranked \{24,27,28,31,42,45,46,47\} (42) & 4 (0.5,1,2,4; Fig.~\ref{fig:alpha}) & 47 / same set & block-vs-residual control on this set at $\alpha{=}4$; the output-ranked set \{14,15,17,18,21,22,23,24\} over-steers at $\alpha{=}4$ (reported at $\alpha{=}1$) \\
Llama-3.1-8B-Inst. & attention output & \{10,12,\dots,24\} (18) & 2 (0.5,1,1.5,2,3,4) & 14 / \{12,14,16,18,20\} & hardened trigger at the deployed point \\
Mistral-7B-Inst.-v0.2 & attention output & \{10,12,\dots,24\} (18) & 2 (0.5,1,1.5,2,3,4) & 14 / \{12,14,16,18,20\} & persona fix: layer 18, $\alpha{=}6$, $\beta{=}6$ \\
RWKV-6-World-7B & time-mix output & \{14,16,\dots,24\} (20) & 16 (0,4,8,16,24,32,48; coh.\ $\ge0.8$) & 20 (target of Llama L14) & defense run gating the block hook at $\alpha{=}6$ is a separate operating point; the block boundary (residual) is inert \\
Zamba2-7B-Inst. & attention out. / Mamba out. & \{36,42,48,54,60\} (48) / \{20,24,28,32\} (28) & 8 / 16 (persona grid $\alpha\in\{2,4,6,8,12\}$, $\beta\in\{0,1,2\}$) & -- & hybrid, both write sites gated \\
\midrule
\multicolumn{6}{@{}R{0.98\textwidth}@{}}{\emph{Baselines on Falcon-Mamba (same gate layers and detector):} residual projection, LEACE-style rank~1 per layer, $\alpha{=}1$; CAST-style conditional residual steering, $\alpha{=}4$ in the main table, self-tuned sweep $\alpha\in\{2,4,8,16\}$ (best 2);}\\
\multicolumn{6}{@{}R{0.98\textwidth}@{}}{always-on (DeTAM-style) edit at the same site and gain; SmoothLLM $k{=}5$ perturbations. \emph{Detector:} logistic probe on standardized mean-pooled activations ($\le$128 prompt tokens), Adam, lr $0.1$, 300 epochs, $\ell_2$ $10^{-3}$}\\
\multicolumn{6}{@{}R{0.98\textwidth}@{}}{(MLP variant: 400 epochs, lr $10^{-2}$, weight decay $10^{-4}$); wrapper-hardened variant adds jailbreak-wrapped training prompts. \emph{Transport map:} ridge $\lambda{=}100$ or orthogonal Procrustes on 250 paired prompts (150 surface-matched, 200 for RWKV).}\\
\multicolumn{6}{@{}R{0.98\textwidth}@{}}{\emph{Adaptive attacks:} $k\in\{0,1,2,3,4,6,8\}$ appended sentences (hardened: up to 12); optimizer 150 rounds $\times$ 64 candidates; GCG 500 steps, 256 candidates, top-256 substitutions, 20-token suffix, target ``Sure, here is'', seeds 0--2.}\\
\bottomrule
\end{tabular}
\end{table*}


\end{document}